\documentclass[acmsmall,nonacm]{acmart}

\AtBeginDocument{%
  }

\usepackage{longtable}
\usepackage{makecell}
\usepackage{array}
\usepackage{booktabs}
\usepackage{multirow}
\usepackage{enumitem}
\usepackage{calc}

\begin{document}
\title{Large Model Empowered Embodied AI: A Survey on Decision-Making and Embodied Learning}

\author{Wenlong Liang}
\affiliation{%
  \institution{University of Electronic Science and Technology of China}
  \city{Chengdu}
  \country{China}}
\email{202421090221@std.uestc.edu.cn}
 
\author{Rui Zhou}
\authornote{Corresponding author.}
\affiliation{%
  \institution{University of Electronic Science and Technology of China}
  \city{Chengdu}
  \country{China}}
\email{ruizhou@uestc.edu.cn}

\author{Yang Ma}
\affiliation{%
  \institution{University of Electronic Science and Technology of China}
  \city{Chengdu}
  \country{China}}
\email{202422090505@std.uestc.edu.cn}

\author{Bing Zhang}
\affiliation{%
  \institution{University of Electronic Science and Technology of China}
  \city{Chengdu}
  \country{China}}
\email{202521090224@std.uestc.edu.cn}

\author{Songlin Li}
\affiliation{%
  \institution{University of Electronic Science and Technology of China}
  \city{Chengdu}
  \country{China}}
\email{202422090530@std.uestc.edu.cn}

\author{Yijia Liao}
\affiliation{%
  \institution{University of Electronic Science and Technology of China}
  \city{Chengdu}
  \country{China}}
\email{202422090531@std.uestc.edu.cn}

\author{Ping Kuang}
\affiliation{%
  \institution{University of Electronic Science and Technology of China}
  \city{Chengdu}
  \country{China}}
\email{kuangping@uestc.edu.cn}


\renewcommand{\shortauthors}{Liang et al.}

\begin{abstract}
Embodied AI aims to develop intelligent systems with physical forms capable of perceiving, decision-making, acting, and learning in real-world environments, providing a promising way to Artificial General Intelligence (AGI). Despite decades of explorations, it remains challenging for embodied agents to achieve human-level intelligence for general-purpose tasks in open dynamic environments. Recent breakthroughs in large models have revolutionized embodied AI by enhancing perception, interaction, planning and learning. In this article, we provide a comprehensive survey on large model empowered embodied AI, focusing on autonomous decision-making and embodied learning. We investigate both hierarchical and end-to-end decision-making paradigms, detailing how large models enhance high-level planning, low-level execution, and feedback for hierarchical decision-making, and how large models enhance Vision-Language-Action (VLA) models for end-to-end decision making. For embodied learning, we introduce mainstream learning methodologies, elaborating on how large models enhance imitation learning and reinforcement learning in-depth. For the first time, we integrate world models into the survey of embodied AI, presenting their design methods and critical roles in enhancing decision-making and learning. Though solid advances have been achieved, challenges still exist, which are discussed at the end of this survey, potentially as the further research directions.
\end{abstract}

\keywords{Embodied AI, Large Model, Hierarchical Decision-Making, End-to-end, Imitation Learning, Reinforcement Learning, World Model}

\maketitle

\section{Introduction}
Embodied AI\cite{xu2024survey} aims to develop intelligent systems with physical forms capable of perceiving, decision-making, acting, and learning in real-world environments. It believes that true intelligence emerges from the interactions of an agent with its environment, providing a promising path to Artificial General Intelligence (AGI)\cite{AGI}. Although the exploration for embodied AI has spanned multiple decades, it remains a challenge to endow agents with human-level intelligence, so that they can perform general-purpose tasks in open, unstructured, and dynamic environments. Early embodied AI systems\cite{brooks2003robust,wilkins2014practical}, grounded in symbolic reasoning and behaviorism, relied on rigid preprogrammed rules, resulting in limited adaptability and superficial intelligence. Although robots are widely used in manufacturing, logistics, and specialized operations, their functionality is limited to controlled environments. Advances in machine learning\cite{murphy2012machine}, particularly deep learning\cite{krizhevsky2012imagenet}, have marked a turning point for embodied AI. Vision-guided planning and reinforcement learning-based control\cite{silver2016mastering} have significantly reduced the dependence of agents on precise environment modeling. Despite the advances, these models are usually trained on task-specific datasets and still face challenges in generalization and transferability, which restricts their adaptation to diverse scenarios for general-purpose applications. Recent breakthroughs in large models\cite{GPT,GPT-2,Llama1,Llama2}, have significantly improved the capabilities of embodied AI. With accurate perception, interaction, and planning abilities, these models have laid the foundations for general-purpose embodied agents\cite{ouyang2022training}. However, the field of embodied AI empowered by large models is still in its nascent stage, struggling with challenges in generalization, scalability, and seamless environmental interaction\cite{sun2024comprehensive}.  There is a great need to have a comprehensive and systematic review of recent advances in large model empowered embodied AI to address the gaps, challenges, and opportunities in the pursuit of AGI.

Through comprehensive investigation in these fields, we find that current studies are scattered, with complex topics but lack of systematic categorization. Existing reviews primarily focus on large models themselves, such as Large Language Models (LLM)\cite{zhao2023survey, chang2024survey, raiaan2024review} and Vision Language Models (VLM)\cite{li2024multimodal, liang2024survey, wang2024exploring}, with little attention to the synergy of large models and embodied agents. Although some reviews involve this integration, they tend to focus on the components, such as planning\cite{wan2024thinking}, learning\cite{9,8,robotlearning}, simulators\cite{wong2025survey}, and applications\cite{wong2025survey,xu2024survey,raychaudhuri2025semantic}, without systematic analysis of the overall paradigm and how these components interact to improve intelligence. Furthermore, some comprehensive surveys miss the recent advances, particularly Vision-Language-Action (VLA)\cite{18} models and end-to-end decision-making, which have gained prominence since 2024. The review \cite{15} provides a detailed introduction to VLA models, but lacks comparison with the hierarchical paradigm and detailed exploration of learning methods. In addition, due to the rapid development of this field, earlier surveys\cite{zhang2022survey,duan2022survey} cannot catch up with the latest progress. In this survey, we focus on the decision-making and learning aspects of large model empowered embodied AI, analyze and categorize the studies, figure out the latest progress, point out the remaining challenges and the future directions, providing researchers with a clear theoretical framework and practical guidance. A comparison between our survey and the related ones are listed in Table\ref{t1}.

{
\footnotesize
\begin{longtable}
{p{\dimexpr\textwidth/10-1\tabcolsep\relax}<{\centering}
                  m{\dimexpr\textwidth/10-0\tabcolsep\relax}<{\centering}
                  m{\dimexpr\textwidth/10-2\tabcolsep\relax}<{\centering}
                  m{\dimexpr\textwidth/10-2\tabcolsep\relax}<{\centering}
                  m{\dimexpr\textwidth/10-2\tabcolsep\relax}<{\centering}
                  m{\dimexpr\textwidth/10-2\tabcolsep\relax}<{\centering}
                  m{\dimexpr\textwidth/10-4\tabcolsep\relax}<{\centering}
                  m{\dimexpr\textwidth/10-4\tabcolsep\relax}<{\centering}
                  m{\dimexpr\textwidth/10-2\tabcolsep\relax}<{\centering}
                  m{\dimexpr\textwidth/10-4\tabcolsep\relax}<{\centering}}
\caption{Comparisons between our survey and the related surveys wrt. surveying scopes.} 
\label{t1} \\
\toprule
\multirow{2}{*}{\makecell[c]{Survey\\type}} & 
\multirow{2}{*}{\makecell[c]{Related\\surveys}} & 
\multirow{2}{*}{\makecell[c]{Publication\\time}} & 
\multirow{2}{*}{\makecell[c]{Large\\models}} & 
\multicolumn{2}{c}{\makecell[c]{Decision-making}} & 
\multicolumn{3}{c}{\makecell[c]{Embodied learning}} & 
\multirow{2}{*}{\makecell[c]{World\\model}} \\
\cmidrule(lr){5-6} \cmidrule(lr){7-9}
& & & & {Hierarchical} & {End2end} & {IL} & {RL} & {Other} & \\
\midrule
\endfirsthead
\caption{Comparisons between our survey and the related surveys wrt. surveying scopes (continued).} \\
\toprule
\multirow{2}{*}{\makecell[c]{Survey\\type}} & 
\multirow{2}{*}{\makecell[c]{Related\\surveys}} & 
\multirow{2}{*}{\makecell[c]{Publication\\time}} & 
\multirow{2}{*}{\makecell[c]{Large\\models}} & 
\multicolumn{2}{c}{\makecell[c]{Decision-making}} & 
\multicolumn{3}{c}{\makecell[c]{Embodied learning}} & 
\multirow{2}{*}{\makecell[c]{World\\model}} \\
\cmidrule(lr){5-6} \cmidrule(lr){7-9}
& & & & {Hierarchical} & {End2end} & {IL} & {RL} & {Other} & \\
\midrule
\endhead
\bottomrule
\endfoot
\multirow{10}{*}{\makecell[c]{Specific}} & \cite{zhao2023survey, chang2024survey, raiaan2024review,li2024multimodal, liang2024survey, wang2024exploring} & 2024 & $\surd$ & $\times$ & $\times$ & $\times$ & $\times$ & $\times$ & $\times$ \\
& \cite{7} & 2024 & $\times$ & $\surd$ & $\times$ & $\surd$ & $\surd$ & $\surd$ & $\times$ \\
& \cite{8} & 2024 & $\surd$ & $\times$ & $\times$ & $\times$ & $\times$ & $\times$ & $\times$ \\
& \cite{9,roadmap} & 2025 & $\times$ & $\times$ & $\times$ & $\surd$ & $\surd$ & $\surd$ & $\times$ \\
& \cite{wan2024thinking} & 2024 & $\times$ & $\surd$ & $\times$ & $\times$ & $\times$ & $\times$ & $\times$ \\
& \cite{robotlearning} & 2024 & $\times$ & $\surd$ & $\times$ & $\times$ & $\times$ & $\surd$ & $\times$ \\
& \cite{12} & 2025 & $\times$ & $\times$ & $\times$ & $\times$ & $\surd$ & $\times$ & $\times$ \\
& \cite{13,14} & 2024 & $\times$ & $\times$ & $\times$ & $\times$ & $\times$ & $\times$ & $\surd$ \\
\midrule
\multirow{5}{*}{\makecell[c]{Compre\\hensive}} & \cite{15} & 2024 & $\surd$ & $\surd$ & $\surd$ & $\surd$ & $\surd$ & $\times$ & $\times$ \\
& \cite{16} & 2024 & $\times$ & $\surd$ & $\surd$ & $\surd$ & $\times$ & $\times$ & $\times$ \\
& \cite{17} & 2024 & $\times$ & $\surd$ & $\surd$ & $\surd$ & $\times$ & $\times$ & $\times$ \\
& \cite{18} & 2024 & $\surd$ & $\surd$ & $\surd$ & $\surd$ & $\times$ & $\surd$ & $\times$ \\
& \textbf{Ours} & -- & ${\surd}$ & $\surd$ & $\surd$ & $\surd$ & $\surd$ & $\surd$ & $\surd$ \\
\end{longtable}
}

The main contributions of our survey are summarized as follows:
\begin{enumerate}[leftmargin=*,topsep=0pt]
  \item \textbf{Focus on the empowerment of large models from the view of embodied AI.} For hierarchical decision-making, embodied AI involves high-level planning, low-level execution, and feedback enhancement, so we review and classify the related works according to this hierarchy. For end-to-end decision making, embodied AI relies on VLA models, so we review VLA models and the enhancements. As the main embodied learning methods are imitation learning (IL) and reinforcement learning (RL), we review how large models empower policy and strategy network construction in imitation learning, and how large models empower reward function design and policy network construction in reinforcement learning. 
  \item \textbf{Comprehensive review on embodied decision-making and embodied learning.} In this survey, we provide a comprehensive review of large model empowered decision-making and learning of embodied AI. For decision-making, we review both hierarchical and end-to-end paradigms empowered by large models and compare them in detail. For embodied learning, we review not only imitation learning and reinforcement learning, but also transfer learning and meta-learning. In addition, we review the world models and how they boost decision-making and learning. 
  \item \textbf{Dual analytical approach for in-depth insights.} We adopt a dual analytical methodology that integrates both horizontal and vertical perspectives. Horizontal analysis reviews and compares diverse approaches, such as diverse large models, hierarchical versus end-to-end decision-making, imitation learning versus reinforcement learning, and diverse strategies for embodied learning. Vertical analysis traces the evolution of core models or methods, detailing their origins, subsequent advances, and open challenges. This dual methodology enables both macro-level overview and in-depth insights of mainstream methods of embodied AI.
\end{enumerate}

\begin{figure}
  \centering
  \includegraphics[width=\linewidth]{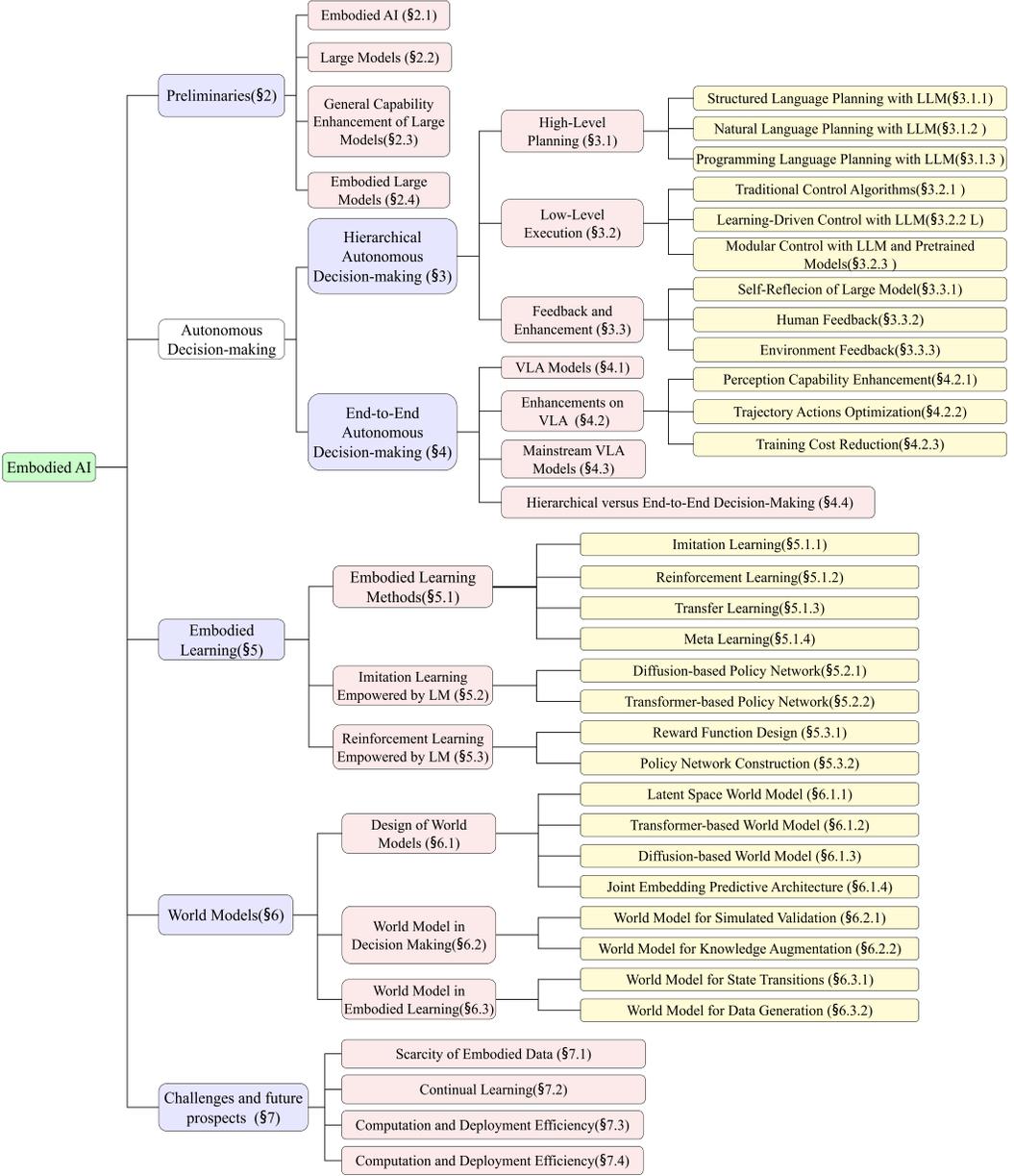}
  \caption{Organization of this survey.}
  \label{f1}
\end{figure}

The organization of our survey is shown in Fig.~\ref{f1}. Section 2 introduces the concept of embodied AI, provides an overview of large models, and discusses general capability enhancement on them. Then it presents the synergy of large models on embodied AI, setting the stage for subsequent sections. Section 3 delves into the hierarchical decision-making paradigm, elaborating on how large models empower dynamic high-level planning, low-level execution, and iterative optimization through feedback. Section 4 focuses on end-to-end decision making. It begins with an introduction and decomposition of VLA models, and then explores the latest enhancements in perception, action generation, and deployment efficiency. At the end of this section, it provides a comprehensive comparison between hierarchical and end-to-end decision-making. Section 5 presents embodied learning methodologies, in particular, large model-enhanced imitation learning and reinforcement learning. Section 6 introduces world models and discusses their role in decision-making and embodied learning of embodied intelligence. Section 7 discusses the open challenges and points out the future prospects. Section 8 concludes the survey.

\section{Preliminaries}
Large models\cite{zhao2023survey, chang2024survey, raiaan2024review} have demonstrated impressive capabilities and gained immense popularity in recent years. Researchers have started to take advantage of these models to build AI agents\cite{ouyang2022training,RT-2,Zero-shot,SayCAN}. In this section, we provide preliminary knowledge about embodied AI and large models. We first introduce the basic concepts and the overall process of embodied AI. Subsequently, we present the mainstream large models and the techniques to enhance their general capabilities. Finally, we discuss the application of large models in embodied AI systems.

\subsection{Embodied AI}

\begin{figure}
  \centering
  \includegraphics[width=\linewidth]{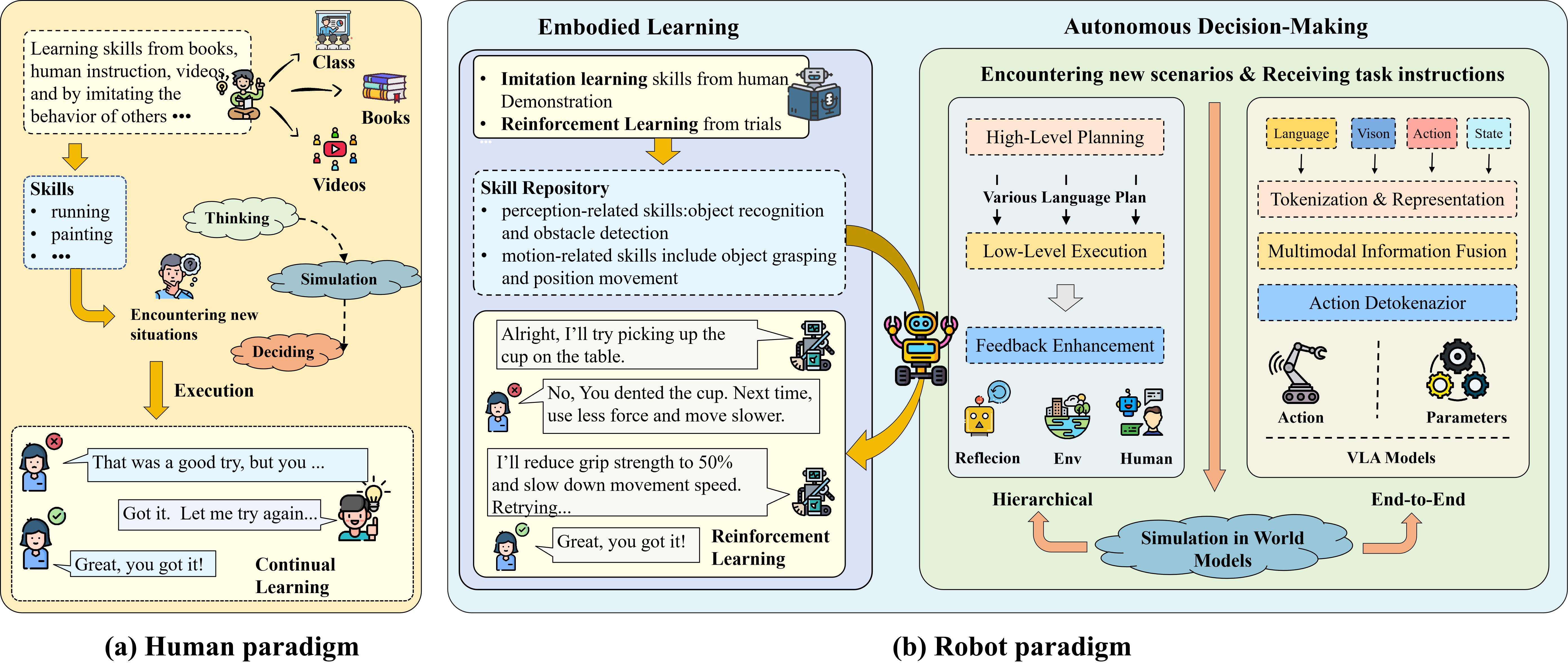}
  \caption{Embodied AI: from prospect of capabilities required during the whole process.}
  \label{f2}
  \Description{Figure:2}
\end{figure}

An embodied AI system typically comprises two primary components: physical entities and intelligent agents\cite{xu2024survey}. Physical entities, such as humanoid robots\cite{maniatopoulos2016reactive}, quadruped robots\cite{bellegarda2022robust}, and intelligent vehicles\cite{reily2022real}, execute actions and receive feedback, serving as the interface between the physical and the digital worlds. Intelligent agents form the cognitive core, enabling autonomous decision-making and learning. To perform embodied tasks, embodied AI systems interpret human intentions from language instructions, actively explore their surroundings, perceive multimodal elements from environments, and execute actions for tasks. This process imitates human learning and problem-solving paradigm. As illustrated in Fig.~\ref{f2}(a), humans learn skills from various resources, for example, books, instructional materials, and online content. When encountering unfamiliar scenarios, they assess the environment, plan necessary actions, mentally simulate strategies, and adapt themselves based on results and external feedback. Embodied agents mimic this human-like learning and problem-solving paradigm, as illustrated in Fig.~\ref{f2}(b). Through imitation learning, agents acquire skills from human demonstrations or video data. When confronting complex tasks or novel environments, they analyze the surroundings, decompose tasks according to objectives, autonomously make execution strategies, and refine plans through simulators or world models. After execution, reinforcement learning optimizes strategies and actions by integrating external feedback, improving overall performance.

The core of embodied intelligence is to enable agents to make decisions and learn new knowledge autonomously in open dynamic environments\cite{roadmap}. Autonomous decision-making can be realized by two approaches: (1) the hierarchical paradigm\cite{SayCAN}, which separates perception, planning, and execution into different modules, and (2) the end-to-end paradigm\cite{RT-2}, which integrates these functionalities into a unified framework for seamless operation. Embodied learning enables agents to refine their behavioral strategies and cognitive models autonomously through long-term environmental interactions, achieving continual improvement. It can be achieved by imitation learning\cite{RT-1} to acquire skills from demonstrations and by reinforcement learning\cite{Reinforcementlearning} to optimize skills through iterative refinement during task execution. Besides, world models\cite{worldmodel} also play a critical role in providing agents with opportunities to make trials and accumulate experiences, by simulating real-world reasoning spaces. These components work synergistically to enhance the capabilities of embodied agents, advancing toward AGI.

\subsection{Large Models}
\begin{figure}
  \centering
  \includegraphics[width=\linewidth]{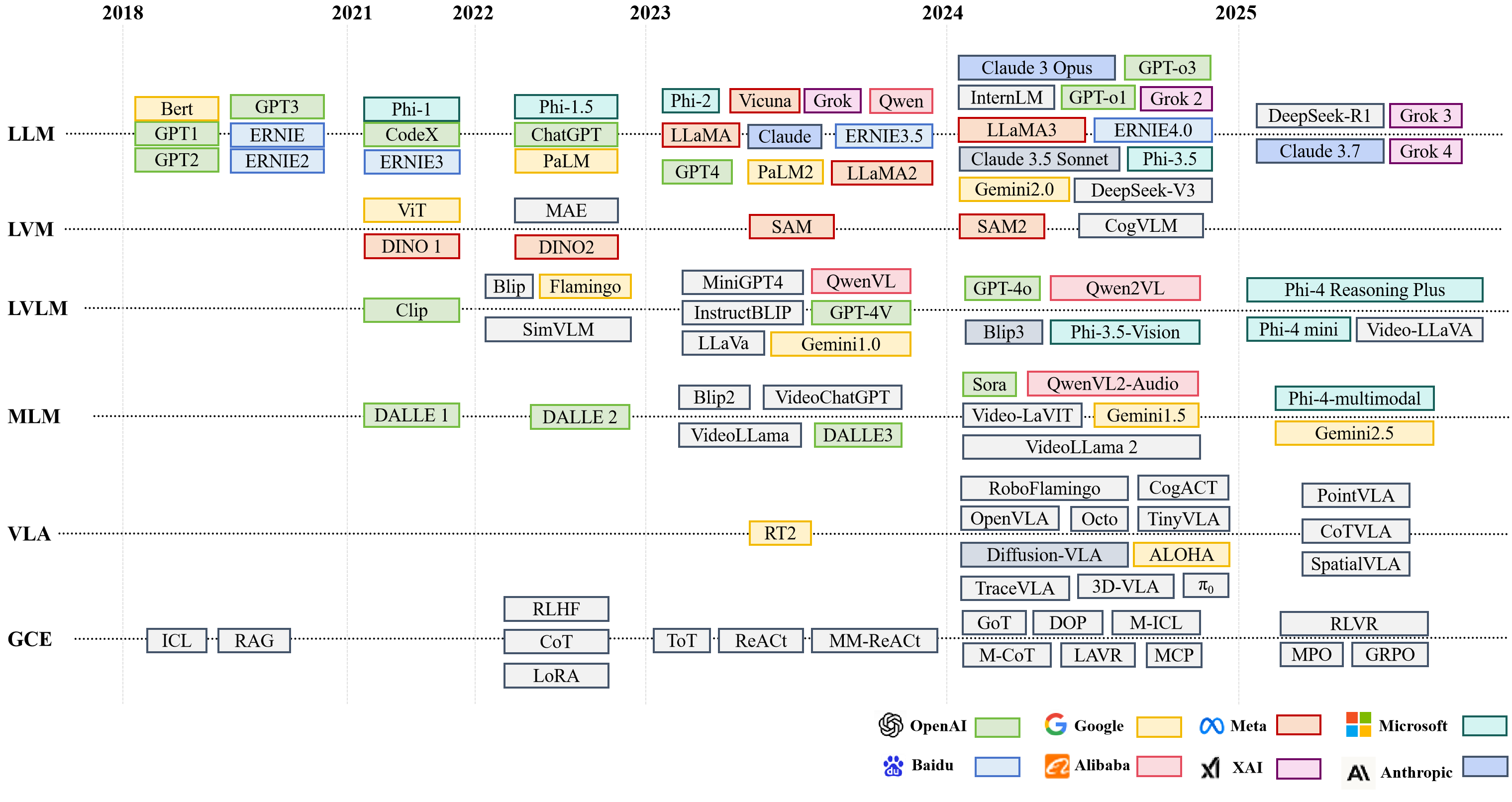}
  \caption{Timeline of major large models. }
  \label{f3}
\end{figure}

Large models, including large language model (LLM), large vision model (LVM), large vision-language model (LVLM), multimodal large model (MLM), and vision-language-action (VLA) model, have achieved remarkable breakthroughs in architecture, data scale, and task complexity, showcasing robust perception, reasoning, and interaction capabilities. Fig.~\ref{f3} shows the timeline of major large models and general capacity enhancement (GCE) to them.

\subsubsection{Large Language Model}
In 2018, Google released BERT\cite{Bert}, a bidirectional Transformer model pretrained with self-supervised tasks, markedly improving performance on natural language tasks. Subsequently, OpenAI released GPT\cite{GPT}, a generative model based on Transformer architecture, which used autoregressive training on large-scale unsupervised corpora to produce coherent text, marking a breakthrough in generative models. GPT-2\cite{GPT-2} further scaled up model size and training data, enhancing text coherence and naturalness. In 2020, GPT-3\cite{GPT3} set a milestone with its vast model capacity and diverse training data, excelling in text generation, question answering, and translation. It showcased zero- and few-shot learning capabilities for the first time, paving the way for future research. Later, Codex\cite{Codex}, pretrained on code datasets, advanced code generation and understanding. ChatGPT\cite{achiam2023gpt} (based on GPT-3.5) enabled natural and fluid interactions with users while supporting a wide range of knowledge domains. Google's PaLM\cite{chowdhery2023palm,anil2023palm} excelled in language understanding, generation, and reasoning through large-scale training and optimized computation. InstructGPT\cite{InstructGPT}, built on GPT-3\cite{GPT}, leveraged reinforcement learning with human feedback (RLHF) to align with human preferences. Meta's Vicuna\cite{Vicuna}, an open-source dialogue model, delivered high-quality interactions with low computational cost, ideal for resource-constrained systems. Meta's Llama series\cite{Llama1,Llama2} (7B, 13B, 30B, 65B parameters) significantly contributed to open-source research and development. 

\subsubsection{Large Vision Model}
LVM\cite{kirillov2023segment} is to process visual information. The Vision Transformer (ViT)\cite{ViT} adapted the Transformer architecture for computer vision, dividing images into fixed-sized patches and using self-attention to capture global dependencies. Based on it, Facebook AI released DINO\cite{DINO} and DINOv2\cite{DINOv2}, leveraging self-supervised learning with ViT. DINO employed a self-distillation approach with a student-teacher network to generate high-quality image representations, capturing semantic structures without labeled data through self-attention and contrastive learning. DINOv2 enhanced DINO with improved contrastive learning and a larger training set, boosting the representation quality. The Masked Autoencoder (MAE)\cite{MAE} exploited self-supervised learning to reconstruct masked visual inputs, enabling pretraining on vast unlabeled image datasets. The Segment Anything Model (SAM)\cite{sam,sam2}, pretrained on 11 million images, supported diverse segmentation tasks, including semantic, instance, and object segmentation, with strong adaptability through user feedback-based fine-tuning.

\subsubsection{Large Vision Language Model}
LVLM\cite{Blip-2} integrates pretrained visual encoders with vision-language fusion modules, allowing to process visual inputs and respond to vision-related queries via language prompts. CLIP\cite{CLIP}, developed by OpenAI, trained image and text encoders via contrastive learning\cite{chen2020simple} on large-scale image-text pairs, aligning paired sample features while minimizing unpaired ones to create visual representations that match textual semantics. BLIP\cite{Blip} employed bidirectional self-supervised learning to fuse visual and linguistic data, using a ``guided'' strategy to enhance pretraining efficiency and improve performance in visual question answering and image captioning. BLIP-2\cite{Blip-2} further introduced the QFormer structure, extracting visual features from a frozen image encoder and aligning them with language instructions via multimodal pretraining for efficient cross-modal fusion. Flamingo\cite{Flamingo} excelled in few-shot learning, processing multimodal data with minimal samples to support cross-modal reasoning in data-scarce scenarios. GPT-4V\cite{gpt-4v} extended traditional GPT to handle joint image-text inputs, generating image descriptions and answering visual questions with robust multimodal reasoning. DeepSeek-V3\cite{DeepSeek-V3} further expanded the boundaries of multimodal reasoning by adopting a dynamic sparse activation architecture. It introduced a hybrid routing mechanism that combined task-specific experts with dynamic parameter allocation, achieving high computational efficiency in cross-modal fusion tasks.

\subsubsection{Multimodal Large Model}
MLM can process diverse modalities, including text, vision, audio, etc. According to the input-output paradigm, MLM can be categorized to multimodal-input text-output models and multimodal-input multimodal-output models. 

Multimodal-input text-output models integrate diverse data modalities for comprehensive content understanding. For instance, Video-Chat\cite{Video-Chat} enhanced video analysis through conversational modeling, excelling in dynamic visual content understanding. Building on Llama architecture, VideoLLaMA\cite{VideoLLaMA} incorporated visual and audio inputs to enable robust video content analysis. Google's Gemini\cite{Gemini}, designed for multimodality, efficiently processed text, image, and audio for image description and multimodal question answering. PaLM-E\cite{PaLM-E} converted multimodal inputs to unified vectors and fed them to the PaLM model for end-to-end training, achieving strong multimodal understanding. 

Multimodal-input multimodal-output models generate diverse data modalities, such as text, image and video, by learning complex data distributions. For instance, DALL·E\cite{DALL·E}, extending GPT-3 with a Vector Quantized Variational Autoencoder (VQ-VAE) and a 1.2-billion-parameter Transformer, generated creative, prompt-aligned images, supporting zero-shot tasks. DALL·E2\cite{DALL·E2} enhanced DALL·E by integrating CLIP to it, employing a two-stage process: first generating low-resolution images then followed by super-resolution enhancement, which improved image quality and diversity greatly. DALL·E3\cite{DALL·E3} further refined image-prompt alignment by enhancing text encoder and training data quality. In 2024, OpenAI released Sora\cite{Sora}, a video generative model that could create high-quality coherent videos up to 60 seconds from text prompts. Sora utilized a coding network to convert inputs into discrete tokens, exploited a large-scale diffusion model to optimize sequences, and projected denoised tokens back to video space.

\subsubsection{Vision-Language-Action Model}VLA models have gained enormous attention recently. Their core objective is to map multimodal inputs to action outputs directly, instead of the intermediate steps of hierarchical decision-making, thereby improving the perception-action integration capabilities of robots. The concept of VLA was first proposed by RT-2\cite{RT-2}, which utilized pretrained vision-language models to discretize action space into action tokens and achieved generalization through joint fine-tuning of Internet data and robot data. However, its discrete action design and closed-source nature limit its flexibility and further research. To overcome the limitations, VLA models based on continuous action generation\cite{li2023vision} and open-source VLA models\cite{kim2024openvla} emerged. Recent studies on VLA models further addressed the challenges. BYO-VLA\cite{BYO-VLA}, 3D-VLA\cite{3D-VLA}, PointVLA\cite{PointVLA} coped with visual input processing. Octo\cite{Octo} and Diffusion-VLA\cite{Diffusion-VLA} addressed action generation accuracy. TinyVLA\cite{TinyVLA} and $\pi_0$\cite{PI0} improved computational efficiency.

\subsection{General Capability Enhancement of Large Models}
\begin{figure}[b]
  \centering
  \includegraphics[width=\linewidth]{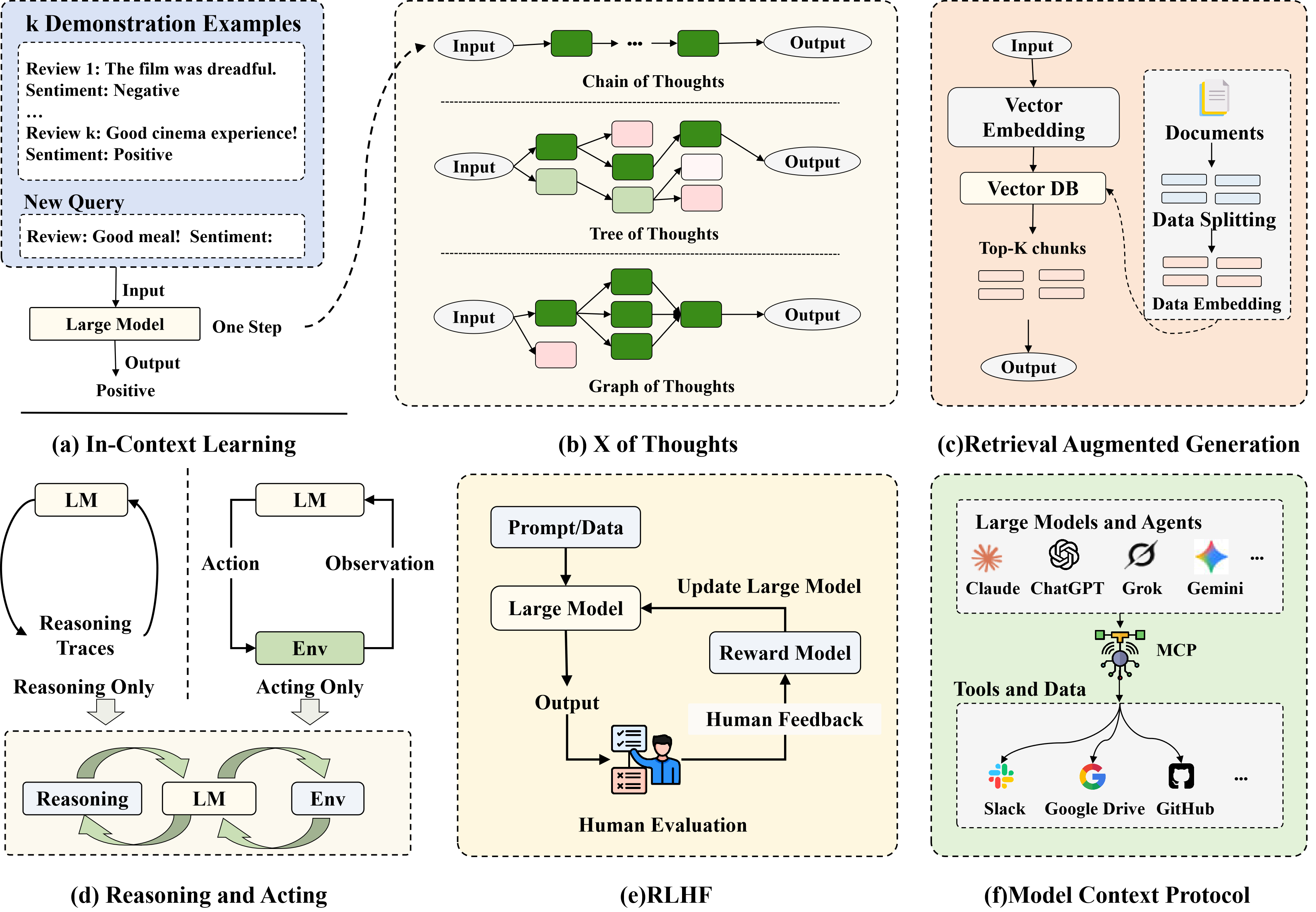}
  \caption{General capability enhancement of large models.}
  \label{f4}
\end{figure}

Large models still suffer from limitations in reasoning ability, hallucination, computational cost, and task specificity. Researchers proposed a series of techniques to enhance their general capabilities, which are illustrated in Fig.~\ref{f4}.

\textit{In-Context Learning (ICL)}\cite{ICL} enables large models to achieve zero-shot generalization through carefully designed prompts, allowing them to tackle new tasks without additional training and tuning. Leveraging the context in input prompts, large models can understand task requirements and generate relevant outputs, making themselves a versatile tool for applications ranging from natural language processing to task-specific problem-solving. Recent advances focus on optimizing prompting techniques, such as automated prompt generation and dynamic example selection, to enhance the robustness of ICL across diverse domains.

\textit{X of Thoughts (XoT)} is a family of reasoning frameworks to improve capabilities of large models in solving mathematical, logical and open-ended problems. Chain of Thoughts (CoT)\cite{CoT} incorporates intermediate reasoning steps into prompts, guiding large models to break down complex problems into manageable parts. Tree of Thoughts (ToT)\cite{ToT} extends CoT by exploring multiple reasoning paths in a tree-like structure, allowing large models to evaluate alternative solutions and backtrack when necessary. Graph of Thoughts (GoT)\cite{GoT} further advances ToT by adopting a graph structure where nodes represent intermediate states and edges capture relationships and dependencies, enabling flexible non-linear reasoning.

\textit{Retrieval Augmented Generation (RAG)}\cite{RAG} retrieves relevant information from external knowledge bases, such as databases and web sources, and feeds it to large models for accurate responses. RAG alleviates the problem of outdated or incomplete knowledge of large models, ensuring access to up-to-date and domain-specific information. Recent advances include hybrid retrieval mechanisms that combine dense and sparse retrieval methods to balance precision and efficiency, and fine-tuning strategies to align retrieved content with generative outputs effectively.

\textit{Reasoning and Acting (ReAct)}\cite{ReAct} integrates reasoning with action execution, enabling to produce explicit reasoning traces during performing tasks. By requiring large models to articulate their thinking processes before acting, ReAct enhances decision transparency and improves performance in dynamic interactive settings. 

\textit{Reinforcement Learning from Human Feedback (RLHF)}\cite{InstructGPT} integrates human preferences into training of large models, aligning large models with human values and intentions. Using human feedback as a reward signal, RLHF improves the model's ability to generate helpful, harmless, and honest outputs in dynamic, interactive settings. By prompting the model to generate multiple responses, RLHF allows humans to rank or rate them based on quality and safety, and uses this feedback to refine the model's future generation, ensuring coherence and ethical considerations.

\textit{Model Context Protocol (MCP)}\cite{MCP}, an open-source standard introduced by Anthropic, provides a standardized interface for large models to interact with external data sources, tools, and services. MCP enhances the interoperability and adaptability of large models, enabling seamless integration with diverse external systems. Recent developments in MCP focus on extending its compatibility with multimodal inputs and optimizing its performance in real-time applications.
\subsection{Embodied Large Models}
Large models empower embodied intelligence by enhancing the capabilities of intelligent agents. Through seamless integration of multiple modalities, including text, vision, audio, and tactile, Embodied Large Models (ELM), also known as Embodied Multimodal Large Models (EMLM), can empower agents to construct sophisticated systems capable of perceiving, reasoning, and acting in complex environments, playing a vital role in autonomous decision-making and embodied learning. 

Different large models empower embodied intelligent agents with varying capabilities. LLM usually serves as cognitive backbone\cite{SayCAN,LLM+P,Zero-shot}, processing natural language inputs, grasping contextual nuances, and generating actionable responses. LVM is typically used in perception tasks or as callable APIs during task execution\cite{DEPS,CLIPort}, leveraging pretrained vision encoders to predict object class, pose, and geometry. LVLM and MLM can further enhance the capabilities of intelligent agents by integrating LLMs with multiple modalities\cite{ViLA,Octopus,ViLaIn}, enabling agents to understand human instructions across text, vision, and audio, generating contextually relevant responses or actions. Recent advances in complex navigation and manipulation tasks have highlighted the advantages of MLM\cite{wong2025survey,xu2024survey}. Different from previous models that handle functionalities separately, VLA models learn end-to-end mappings from visual and linguistic inputs to executable actions\cite{RT-2,OpenVLA,PI0}. This streamlined pipeline enables agents to interpret complex instructions, perceive dynamic environments, and execute precise physical movements, resulting in more robust and versatile embodied AI systems. Beyond enhancing planning intelligence, researchers are increasingly exploring their generative capabilities to advance embodied learning\cite{robotlearning,Eureka,DiffusionPolicy} and assist in constructing world models\cite{UniSim,WKM,UniPi}, further supporting the path to AGI.
\section{Hierarchical Autonomous Decision-Making}

\begin{figure}
  \centering
  \includegraphics[width=\linewidth]{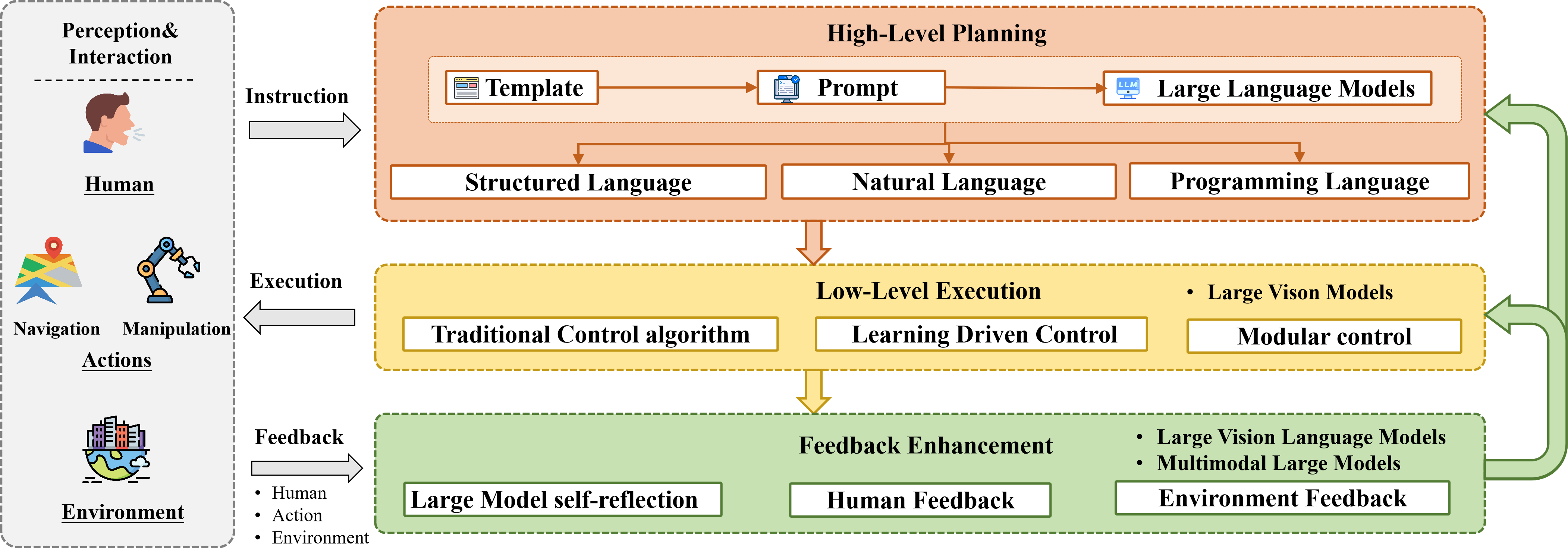}
  \caption{Hierarchical decision-making paradigm, consisting of perception and interaction, high-level planning, low-level execution, feedback and enhancement.}
  \label{f5}
\end{figure}

Autonomous decision-making of agents aims to turn environmental perception and task understanding into executable decisions and physical actions. Conventional decision-making takes a hierarchical paradigm, consisting of perception and interaction, high-level planning, low-level execution, and feedback and enhancement. The perception and interaction layer relies on vision models, the high-level planning layer relies on predefined logical rules\cite{fox2003pddl2}, and the low-level execution layer relies on classic control algorithms\cite{franklin2002feedback}. These methods excel in structured environments, but struggle in unstructured or dynamic settings, due to limited holistic optimization and high-level decision-making. 

Advances in large models, with their robust learning, reasoning, and generalization capabilities, have shown promises in complex task processing. By integrating the reasoning capabilities of large models with the execution capabilities of physical entities, it provides a new paradigm for autonomous decision-making. As shown in Fig.~\ref{f5}, environmental perception firstly interprets the agent’s surrounding environment, LLM-empowered high-level planning subsequently decomposes complex tasks into sub-tasks in consideration of perception information and task instructions, LLM-empowered low-level execution then translates sub-tasks into precise physical actions, and finally LLM-empowered feedback enhancement introduces closed-loop feedback to enhance the intelligence. 

\subsection{High-Level Planning}
\begin{figure}[h]
  \centering
  \includegraphics[width=\linewidth]{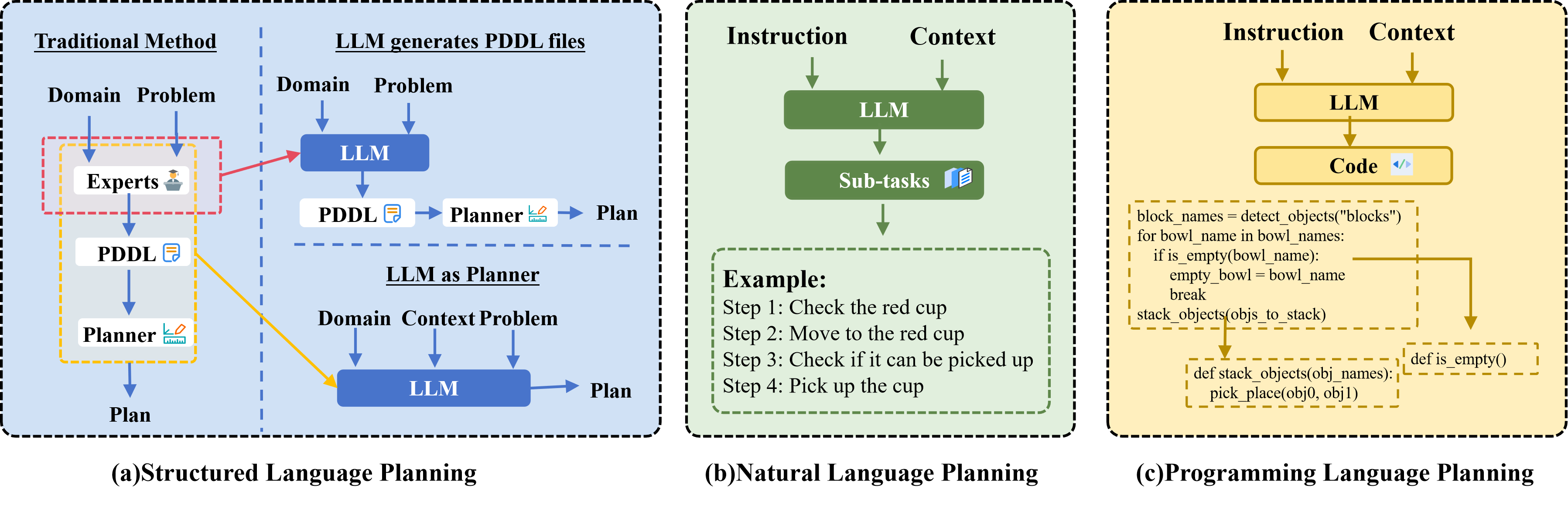}
  \caption{High-level planning empowered by large models.}
  \label{f6}
\end{figure}

High-level planning produces reasonable plans according to task instructions and perceived information. Traditional high-level planning relies on rule-based methods\cite{haslum2019introduction,McDermott1998PDDLthePD,gelfond2014knowledge}. Given initial states and goals specified in Planning Domain Definition Language (PDDL), heuristic search planner verifies action preconditions for feasibility and employs search trees to select optimal action sequences, thereby generating an efficient and cost-effective plan\cite{jiang2019multi}. Although effective in structured environments, rule-based methods struggle with adaptability in unstructured or dynamic scenarios. Large models, leveraging their zero-shot and few-shot generalization capabilities, have driven breakthroughs in addressing these challenges. According to the form of planning, LLM-empowered high-level planning can be categorized into structured language planning, natural language planning, and programming language planning, as illustrated in Fig.~\ref{f6}. 

\subsubsection{Structured Language Planning with LLM}
LLM can enhance structured language planning through two key strategies, as shown in Fig.~\ref{f6}(a). 
(1) The first strategy employs LLM as the planner, utilizing its zero/few-shot generalization capability to produce plans. However, Valmeekam et al.\cite{valmeekam2022large} demonstrated that LLM often generated infeasible plans due to strict PDDL syntax and semantics, leading to logical errors. To mitigate this problem, LLV\cite{LLV} introduced an external validator, a PDDL parser or an environment simulator, to check and iteratively refine LLM-generated plans through error feedback. FSP-LLM\cite{FSP-LLM} optimized prompt engineering to align plans with logical constraints, ensuring feasibility of tasks.
(2) The second strategy leverages LLM to automate PDDL generation, reducing manual effort in domain modeling. In LLM+P\cite{LLM+P}, LLM created PDDL domain files and problem descriptions, which were then solved by traditional planners, combining linguistic understanding with symbolic reasoning. PDDL-WM\cite{PDDL-WM} used LLM to iteratively construct and refine PDDL domain models, which were validated by parsers and user feedback to ensure correctness and executability. 
By leveraging LLM as direct planner or PDDL generator, these strategies enhance automation and reduce user involvement, thus improve planning efficiency, adaptability, and scalability significantly.

\subsubsection{Natural Language Planning with LLM}
Natural language offers greater expression flexibility than structured language, enabling full harness of LLM to decompose complex plans into sub-plans\cite{sharma2021skill,li2022pre}, as illustrated in Fig.~\ref{f6}(b). However, natural language planning often produces infeasible plans, as its outputs are often based on experiences rather than actual environments. For instance, when instructed to ``clean the room'', an LLM may propose “retrieve the vacuum cleaner”, without verifying the availability of it. Zero-shot\cite{Zero-shot} explored the feasibility of using LLM to decompose high-level tasks into a series of executable language planning steps. Their experiments showed that LLM could generate preliminary plans based on common sense reasoning, but lacked constraints on physical environments and action feasibility. 

To address this problem, SayCAN\cite{SayCAN} integrates LLM with reinforcement learning, combining LLM-generated plans with a predefined skill repository and value functions to evaluate action feasibility. By scoring actions with expected cumulative rewards, SayCAN filters out impractical steps (e.g., “jump onto the table to grab a cup”), in favor of safer high-value actions (e.g., “move to the table and reach out”). Text2Motion\cite{Text2Motion} further enhances planning for tasks involving spatial interactions by incorporating geometric feasibility. It uses LLM to propose candidate action sequences, which are then evaluated by a checker for physical viability, to avoid actions like “stacking a large box on a small ball”. However, both approaches rely on fixed skill sets, lacking adaptability to open-ended tasks. Grounded Decoding\cite{GroundedDecoding} addresses this limitation by introducing a flexible decoding strategy. It dynamically integrates LLM outputs with a real-time grounded model, which assesses action feasibility according to current environmental states and agent capabilities, guiding LLM to produce contextually viable plans. 

\subsubsection{Programming Language Planning with LLM}
Programming language planning transforms natural language instructions to executable programs, leveraging the precision of code to define spatial relationships, function calls, and control APIs for dynamic high-level planning in embodied tasks, as shown in Fig.~\ref{f6}(c). CaP\cite{CaP} converts task planning to code generation, producing Python-style programs with recursively defined functions to create a dynamic function library. For instance, in robot navigation, CaP firstly defines a “move” function, then extends it to “avoid obstacle move” or “approach target” based on task requirements. This self-extending library enhances adaptability to new tasks without predefined templates. However, CaP’s reliance on perception APIs and unconstrained code generation limit its handling of complex instructions. To overcome these limitations, Instruct2Act\cite{Instruct2Act} offers a more integrated solution by leveraging multimodal foundation models to unify perception, planning, and control. It uses vision-language models for accurate object identification and understanding of spatial relationships, providing precise environmental awareness. The perception data is then fed to LLM, which generates code-based action sequences from a predefined robot skill repository. This method substantially increases planning accuracy and allows agents to adapt to new environments effectively, especially in tasks with significant visual components. ProgPrompt\cite{ProgPrompt} employs structured prompts with environmental operations, object descriptions, and example programs to guide LLM in generating tailored, code-based plans. By incorporating predefined constraints, ProgPrompt minimizes invalid code generation and enhances cross-environment adaptability.

\subsection{Low-Level Execution}
\begin{figure}[h]
  \centering
  \includegraphics[width=\linewidth]{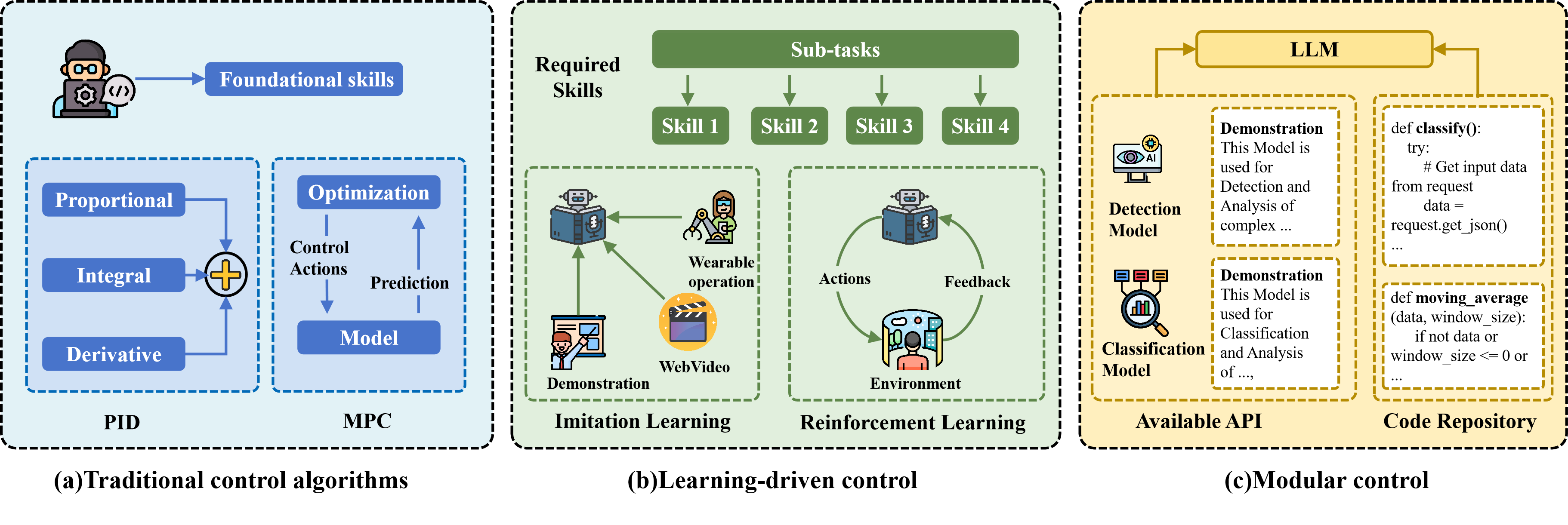}
  \caption{Low-level execution.}
  \label{f7}
\end{figure}

Following high-level task planning, low-level actions are executed using predefined skill lists\cite{Zero-shot}. Skill lists represent a series of basic capabilities or action modules required by embodied agents to perform specific tasks. They serve as a bridge between task planning and physical execution. For example, perception-related skills include object recognition and obstacle detection, while motion-related skills include object grasping and moving. Implementation of low-level skills involves control theory, machine learning, and robotics engineering. The approaches evolve from traditional control algorithms to learning-driven control and to modular control, as shown in Fig.~\ref{f7}.

\subsubsection{Traditional Control Algorithms}
Foundational skills are typically designed using traditional control algorithms, which mainly make use of classic model-based techniques with clear mathematical derivations and physical principles. Proportional-integral-derivative (PID) control\cite{MCP} adjusts parameters to minimize errors in robotic arm joint control. State feedback control\cite{sfc}, often paired with linear quadratic regulators (LQR)\cite{LQR}, optimizes performance using system state data. Model predictive control (MPC)\cite{MPC} forecasts states and generates control sequences via rolling optimization, ideal for tasks like drone path tracking. Traditional control algorithms provide mathematical interpretability, low computational complexity and real-time performance, enabling reliable task execution. However, when faced with dynamic environments, traditional control algorithms lack adaptability and struggle to handle high-dimensional uncertain system dynamics. They need to integrate with data-driven techniques, such as deep learning and reinforcement learning, to enhance generalization capabilities. For example, when a quadruped robot navigates uneven terrain, traditional PID control collaborates with learning algorithms to dynamically adjust its gait.
\subsubsection{Learning-Driven Control with LLM}
Robot learning lies at the junction of machine learning and robotics. It enables intelligent agents to develop control strategies and low-level skills from extensive data, including human demonstrations, simulations, and environmental interactions. Imitation learning and reinforcement learning represent two important learning methods for this purpose. Imitation learning trains strategies from expert demonstrations, enabling rapid policy development with reduced exploration time. Embodied-GPT\cite{Embodied-GPT} leverages a 7B language model for high-level planning and converts plans to low-level strategies via imitation learning. Reinforcement learning optimizes strategies through iterative trials and environmental rewards, suitable in high-dimensional dynamic environments. Hi-Core\cite{Hi-Core} employs a two-layer framework where LLM sets high-level strategies and sub-goals, while reinforcement learning generates specific actions at low levels. These learning-driven control methods empowered by LLM offer strong adaptability and generalization. However, their training typically requires a large amount of data and computational resources, and the convergence and stability of the strategies are difficult to guarantee. 
\subsubsection{Modular Control with LLM and Pretrained Models}
Modular control integrates LLM with pretrained strategy models, such as CLIP\cite{CLIP} for visual recognition and SAM\cite{kirillov2023segment} for segmentation. By equipping LLM with descriptions of these tools, they can be invoked dynamically during task execution. DEPS\cite{DEPS} combines multiple different modules to complete detection and actions based on task requirements and natural language descriptions of the pretrained models. PaLM-E\cite{PaLM-E} merges LLM with visual modules for segmentation and recognition. CLIPort\cite{CLIPort} leverages CLIP for open-vocabulary detection. \cite{CaP} leveraged LLM to generate codes to create libraries of callable functions for navigation and operations. This modular approach ensures scalability and reusability across diverse tasks by leveraging shared pretrained models. 

However, challenges exist. Firstly, calling external strategy models may introduce additional computational and communication delays, especially in real-time tasks (e.g., autonomous driving\cite{Autonomousdriving}), where such delays may affect the response efficiency significantly. Secondly, the overall performance of intelligent agents is highly dependent on the quality of the pretrained strategy models. If the strategy models have defects (such as insufficient generalization ability or training data bias), even with strong planning capabilities of LLM, the execution results may still be unsatisfactory. Therefore, it is important to optimize the communication efficiency between modules, improve the robustness of the strategy models, and design more intelligent invocation decision mechanisms.

\subsection{Feedback and Enhancement}
\begin{figure}[h]
  \centering
  \includegraphics[width=\linewidth]{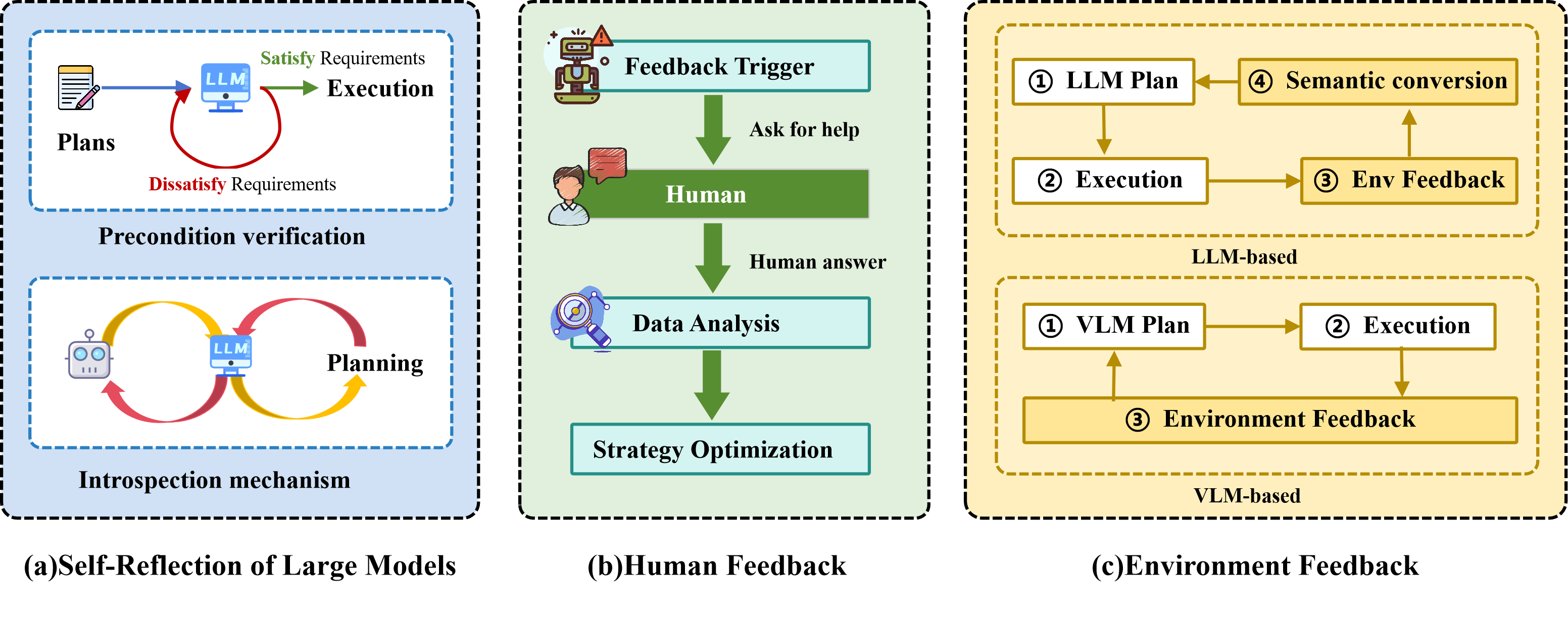}
  \caption{Feedback and enhancement.}
  \label{f8}
\end{figure}

Hierarchical decision-making architecture guides task planning through task descriptions and example prompts. To ensure the quality of task planning, a closed-loop feedback mechanism should be introduced. The feedback may come from the large model itself, humans, and external environments, as shown in Fig.~\ref{f8}.

\subsubsection{Self-Reflection of Large Models}
The large models can act as task planners, evaluators and optimizers, thus iteratively refine decision-making processes without external interventions. The agents get feedback of actions, autonomously detect and analyze failed executions, and continuously learn from previous tasks. By this self-reflection and optimization mechanism, the large models can generate robust strategies, offering advantages in long-sequence planning, multimodal tasks, and real-time scenarios. Self-reflection can implemented in two ways, as depicted in Fig.~\ref{f8}(a). 

(1) The first approach triggers plan regeneration by Re-Prompting\cite{raman2022planning} based on detected execution failures or precondition errors. Re-Prompting integrates error context (e.g., failing to unlock a door before opening it) as feedback to dynamically adjust prompts and thereby correct LLM-generated plans. DEPS\cite{raman2022planning} adopts a “describe, explain, plan, select” framework, where LLM describes execution process, explains failure causes, and re-prompts to correct plan deficiencies, enhancing interactive planning. 

(2) The second approach takes introspection mechanism, enabling LLM to evaluate and refine its output independently. Self-Refine\cite{Self-Refine} uses a single LLM for planning and optimization, iteratively improving plan rationality through multiple self-feedback cycles. Reflexion\cite{reflexion} extends it by incorporating a long-term memory to store evaluation results, combining multiple feedback mechanisms to enhance plan feasibility. ISR-LLM\cite{ISR-LLM} applies iterative self-optimization in PDDL-based planning, generating initial plans, performing rationality checks, and refining outcomes via self-feedback. Voyager\cite{Voyager}, tailored for programming language planning, builds a dynamic code skill library by extracting feedback from execution failures, enabling agents to adapt to complex tasks.

\subsubsection{Human Feedback}
Human feedback enhances planning accuracy and efficiency by establishing an interactive closed-loop mechanism with humans, as illustrated in Fig.~\ref{f8}(b). This approach enables agents to dynamically adjust behaviors based on human feedback. KNOWNO\cite{KNOWNO} introduces an uncertainty measurement framework, allowing LLM to identify knowledge gaps and seek human assistance in high-risk or uncertain scenarios. EmbodiedGPT\cite{EmbodiedGPT} takes a planning-execution-feedback loop, where agents request human inputs when low-level controls fail. This human feedback, combined with reinforcement learning and self-supervised optimization, enables the agent to iteratively refine its planning strategies, ensuring better alignment with dynamic environmental conditions. YAY Robot\cite{YAY} allows users to pause robots with commands and provide guidance, facilitating real-time language-based corrections. The feedback is recorded for strategy fine-tuning and periodic queries, enabling real-time and long-term improvements. IRAP\cite{IRAP} allows interactive question-answering with humans to acquire task-specific knowledge, enabling precise robot instructions. 

\subsubsection{Environment Feedback}
Environment feedback enhances LLM-based planning via dynamic interactions with environments, as depicted in Fig.~\ref{f8}(c). Inner Monologue\cite{InnerMonologue} transforms multimodal inputs to linguistic descriptions for an “inner monologue” reasoning, allowing LLM to adjust plans based on environmental feedback. TaPA\cite{TaPA} integrates open-vocabulary object detection and tailors plans for navigation and operations. DoReMi\cite{DoReMi} detects discrepancies between planned and actual outcomes and takes multimodal feedback to adjust tasks dynamically. In multi-agent settings, RoCo\cite{RoCo} leverages environmental feedback and inter-agent communications to correct robotic arm path plannings in real time. 

LLM-based planning often requires to convert feedback to natural language. VLM simplifies this by integrating visual inputs and language reasoning, avoiding feedback conversions. ViLaIn\cite{ViLaIn} integrates LLM with VLM to generate machine-readable PDDL from language instructions and scene observations, driving symbolic planners with high precision. ViLA\cite{ViLA} and Octopus\cite{Octopus} achieve robot vision language planning by making advantage of GPT4-V MLM to generate plans, integrating perception data for robust zero-shot reasoning. Voxposer\cite{Voxposer} further exploits MLM to extract spatial geometric information, generate 3D coordinates and constraint maps from robot observations to populate code parameters, thereby enhancing spatial accuracy in planning. 

\section{End-to-End Autonomous Decision-Making}
Hierarchical paradigm relies on separate task-planning, action-execution and feedback modules, hence it suffers from error accumulation and struggles to generalize across diverse tasks. Moreover, high-level semantic knowledge derived from large models is difficult to apply to robotic action execution directly, resulting in integration gaps. To mitigate these challenges, end-to-end autonomous decision-making gains significant attention recently, which maps multimodal inputs (i.e. visual observations and linguistic instructions) to actions directly. It is typically implemented by VLA, as illustrated in Fig.~9.

\begin{figure}[h]
  \centering
  \includegraphics[width=\linewidth]{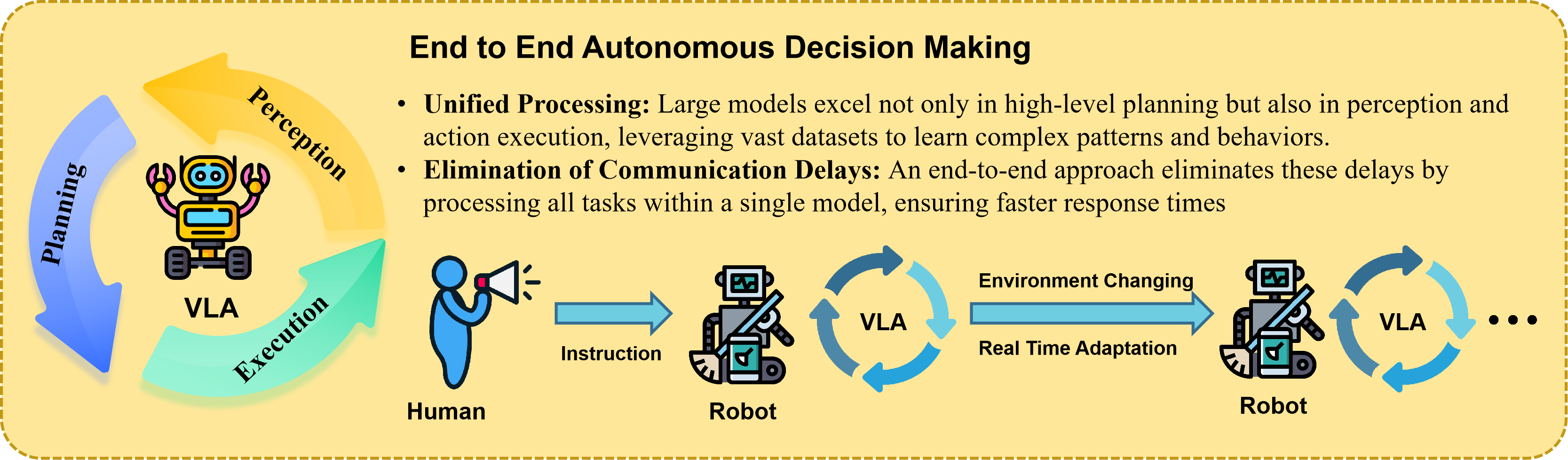}
  \caption{End-to-end decision-making by VLA.}
  \label{f9}
\end{figure}

\subsection{Vision-Language-Action Models}
\begin{figure}[h]
  \centering
  \includegraphics[width=\linewidth]{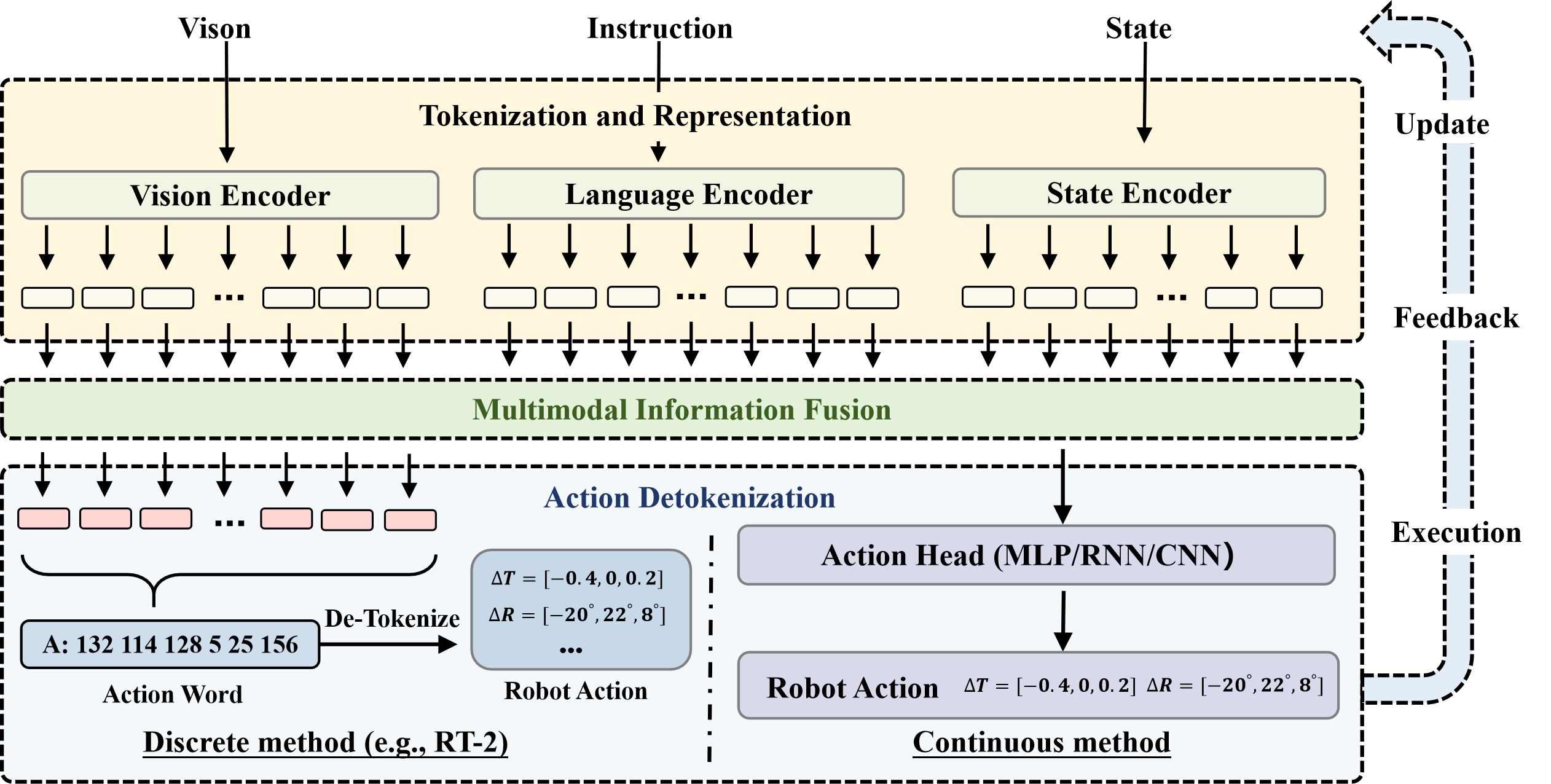}
  \caption{Vision-Language-Action Models.}
  \label{f10}
\end{figure}

VLA models represent a breakthrough in embodied AI by integrating perception, language understanding, planning, action execution and feedback optimizing into a unified framework. By leveraging rich prior knowledge of large models, VLA models can achieve precise and adaptable task execution in dynamic, open environments. A typical VLA model comprises three key components: tokenization and representation, multimodal information fusion, and action detokenization, as illustrated in the Fig.~\ref{f10}.

(1) \textit{Tokenization and Representation.} VLA models use four token types: vision, language, state, and action, to encode multimodal inputs for context-aware action generation. Vision tokens and language tokens encode environmental scenes and instructions into embeddings, forming the foundation of task and context. State tokens capture the agent’s physical configuration, including joint positions, force-torque, gripper status, end-effector pose, and object locations. Action tokens, generated autoregressively based on prior tokens, represent low-level control signals (e.g., joint angles, torque, wheel velocities) or high-level movement primitives (e.g., “move to grasp pose,” “rotate wrist”), enabling VLA models to act as language-driven policy generators.
  
(2) \textit{Multimodal Information Fusion.} Visual tokens, language tokens, and state tokens are fused into a unified embedding for decision-making through a cross-modal attention mechanism, typically implemented within a transformer architecture. This mechanism dynamically weighs the contributions of each modality, enabling the VLA model to jointly reason over object semantics, spatial layouts, and physical constraints based on the task context.
  
(3) \textit{Action Detokenization.}The fused embedding is then passed to an autoregressive decoder, typically implemented within the transformer architecture, to generate a series of action tokens corresponding to low-level control signals or high-level movement primitives. Action generation can be discrete or continuous. In discrete action generation, the model selects from a predefined set of actions, such as specific movement primitives or discretized control signals, which are mapped to executable commands. In continuous action generation, the model outputs fine-grained control signals, often sampled from a continuous distribution using a final MLP layer, enabling precise manipulation or navigation. These action tokens are detokenized by mapping them to executable control commands, which are passed to the execution loop. The loop feeds back updated state information, enabling the VLA model to dynamically adapt to perturbations, object shifts, or occlusions in real time.

Robotics Transformer 2 (RT-2)\cite{RT-2} is a renowned VLA model. It leverages Vision Transformer (ViT)\cite{ViT} for visual processing and leverages PaLM to integrates vision, language, and robot state information. Specially, RT-2 discretizes the action space into eight dimensions (including 6-DoF end-effector displacement, gripper status, and termination commands). Each dimension is divided into 256 discrete intervals except for the termination command and embedded into the VLM vocabulary as action tokens. During training, RT-2 employs a two-stage strategy: first pretraining with Internet-scale vision-language data to enhance semantic generalization; then fine-tuning to map inputs (i.e., robot camera images and text task descriptions) to outputs (i.e., action word token sequences, e.g., “1 128 91 241 5 101 127 255”). The trained VLA model can generate action word in an autoregressive manner based on vision-language inputs, and decode them into specific action sequences via a predefined mapping table. By modeling actions as “language”, RT-2 leverages large models' capabilities to enhance low-level action commands with rich semantic knowledge. 

\subsection{Enhancements on VLA}
\begin{figure}[h]
  \centering
  \includegraphics[width=\linewidth]{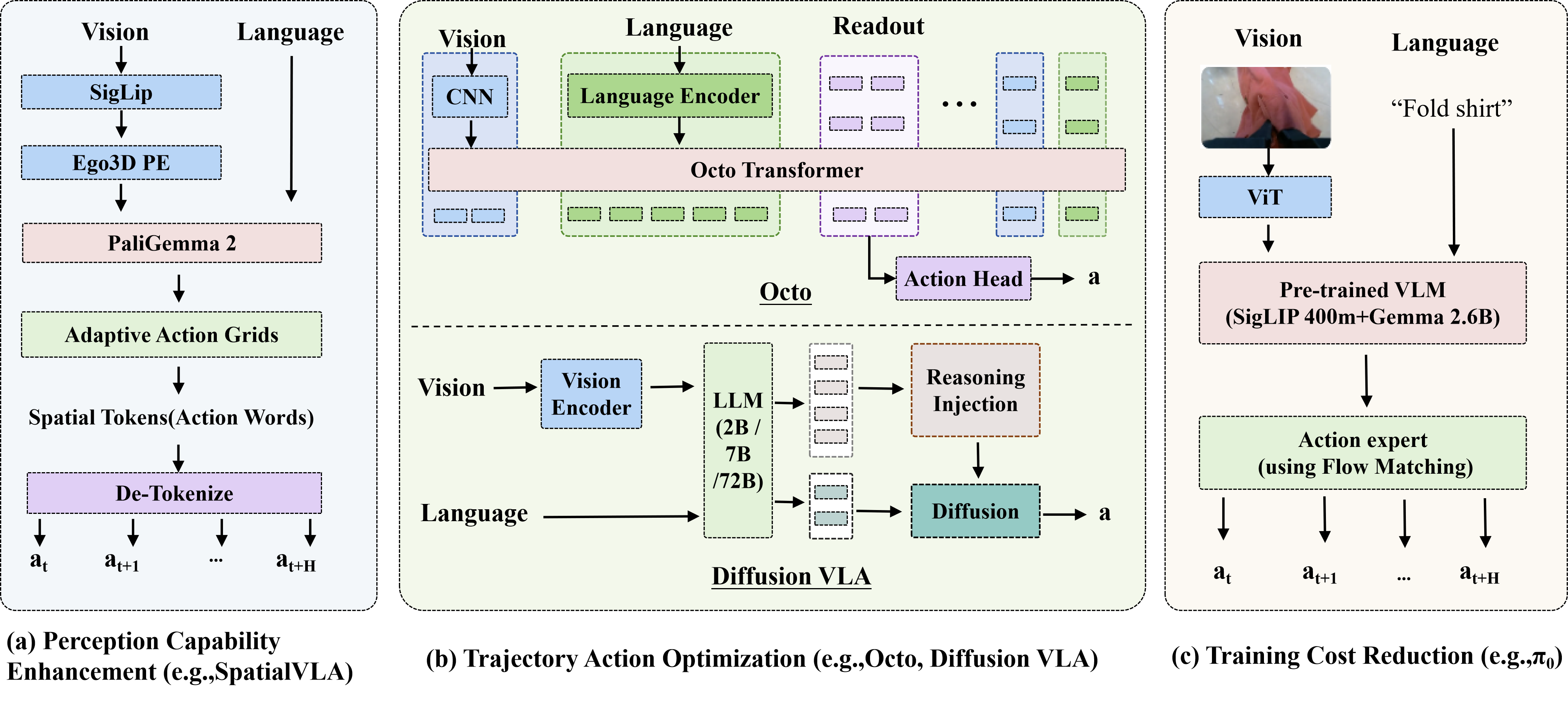}
  \caption{Enhancements on Vision-Language-Action Models.}
  \label{f11}
\end{figure}

Although the VLA end-to-end decision-making architecture is powerful, it exhibits notable limitations that constrain its performance in complex embodied tasks. Firstly, the real-time closed-loop mechanism causes VLA models highly sensitive to perturbations in visual and language inputs, where visual noise (e.g., occlusions or cluttered backgrounds) can destabilize action outputs, compromising task reliability. In addition, the reliance on 2D perception limits the model’s ability to interpret intricate 3D spatial relationships. Secondly, the action generation process often depends on simplistic policy networks at the output layer, which have difficulties in meeting the requirements of high-precision and dynamically evolving tasks, resulting in sub-optimal trajectories. Thirdly, training VLA models demands high computation resources, leading to high deployment costs and scalability challenges. To address these issues and advance the applicability of VLA in complex scenarios, researchers have proposed some enhancements. We categorize them as: perception capability enhancement (to solve the first problem), trajectory action optimization (to solve the second problem), and training cost reduction (to solve the third problem), as illustrated in the Fig.~\ref{f11}.

\subsubsection{Perception Capability Enhancement}
To improve the perception capability, BYO-VLA\cite{BYO-VLA} optimizes the tokenization and representation component by implementing a runtime observation intervention mechanism, which leverages automated image preprocessing to filter out visual noise originating from occluding objects and cluttered backgrounds. TraceVLA\cite{TraceVLA} focuses on the multimodal information fusion component, introducing visual trajectory prompts to the cross-modal attention mechanism. By incorporating trajectory-related data with vision, language, and state tokens, TraceVLA enhances spatio-temporal awareness, enabling precise action trajectory predictions. BYO-VLA improves input quality, while TraceVLA refines integration of dynamic information during fusion. For 3D perception, 3D-VLA\cite{3D-VLA} combines a 3D large model with a diffusion-based world model to process point clouds and language instructions. It generates semantic scene representations and predicts future point cloud sequences, improving 3D object relationship understanding, thereby surpassing 2D VLA models in complex 3D settings. SpatialVLA\cite{spatialvla} further emphasizes the problem of spatial understanding in robot sorting tasks. It proposes Ego3D position encoding to inject 3D information directly into input observations and adopts adaptive action schemes to improve robots' adaptability in different environments.

\subsubsection{Trajectory Action Optimization}
Discrete action space limits the expression of undefined or high-precision actions. Diffusion-enhanced methods can provide smoother and more controllable actions by modeling complex robot behaviors through diffusion models. Octo\cite{Octo} combines Transformer and diffusion models to generate robot actions. It processes multimodal inputs through Transformer, extracts visual-language features, and uses conditional diffusion decoders to iteratively optimize action sequences based on these features, so as to generate smooth and precise trajectories. By modular design and efficient fine-tuning, Octo achieves cross-task generalization with only a small amount of task-specific data. Diffusion-VLA\cite{Diffusion-VLA} combines a language model with a diffusion policy decoder into a unified framework. It uses an autoregressive language model to parse language instructions and generate preliminary task representations, which are fed into the diffusion policy decoder, optimizing the action sequence through a gradual denoising process. Diffusion-VLA employs end-to-end training across the entire framework, jointly optimizing language understanding and action generation. The diffusion process corrects the discontinuities in the autoregressive outputs at each step, ensuring the smoothness and robustness of action trajectories. Compared to Octo, Diffusion-VLA has higher computational cost but is more suitable for complex tasks requiring deep semantic-action fusion.

\subsubsection{Training Cost Reduction}
VLA models in complex tasks require high computational cost, which is limited on resource-constrained embodied platforms. To reduce the training cost, researchers have proposed optimization methods to improve inference speed, data efficiency, and real-time performance while maintaining task performance. $\pi_0$\cite{PI0} exploits flow matching to represent complex continuous action distributions. Compared to multi-step sampling used in diffusion models, flow matching optimizes action generation process through continuous flow field modeling, consequently reducing computational overhead and improving real-time performance. Compared to Diffusion-VLA\cite{Diffusion-VLA} and Octo\cite{Octo}, the improvements in computational efficiency and control accuracy make $\pi_0$ more suitable for resource-constrained embodied applications, especially tasks requiring high-precision continuous control. In addition, TinyVLA\cite{RT-2} achieves significant improvement in inference speed and data efficiency by designing a lightweight multimodal model and a diffusion strategy decoder. OpenVLA-OFT\cite{OpenVLA-OFT} uses parallel decoding instead of traditional autoregressive generation to generate complete action sequences in a single forward pass rather than one by one, thereby reducing the inference time significantly.

\subsection{Mainstream VLA Models}
A large number of VLA models have emerged recently, with various architectures and capabilities. For better understanding and deployment, we summarize and compare them in Table~\ref{t2} with respect to architecture, contributions and capability enhancements.
{
\footnotesize
\begin{longtable}{m{1.2cm}<{\centering}m{5cm}<{\centering}m{4cm}<{\centering}m{0.4cm}<{\centering}m{0.4cm}<{\centering}m{0.4cm}<{\centering}}
\caption{Mainstream VLA models (P: perception, A: trajectory action, C: training cost).} \label{t2} \\
\toprule
\multirow{2}{*}{\makecell[c]{Model}} & 
\multirow{2}{*}{\makecell[c]{Architecture}} & 
\multirow{2}{*}{\makecell[c]{Contributions}}  & 
\multicolumn{3}{c}{Enhancements} \\
\cmidrule(lr){4-6}
& & & P & A & C \\
\midrule
\endfirsthead
\caption{Mainstream VLA models (P: perception, A: trajectory action, C: training cost) (continued).} \\
\toprule
\multirow{2}{*}{\makecell[c]{Model}} & 
\multirow{2}{*}{\makecell[c]{Architecture}} & 
\multirow{2}{*}{\makecell[c]{Contributions}}  & 
\multicolumn{3}{c}{Enhancements} \\
\cmidrule(lr){4-6}
& & & P & A & C \\
\midrule
\endhead
\bottomrule
\endfoot
\footnotesize
RT-2 \cite{RT-2} (2023) & 
\begin{itemize}[leftmargin=*]
  \item Vision Encoder: ViT22B/ViT-4B
  \item Language Encoder: PaLIX/PaLM-E
  \item Action Decoder: Symbol-tuning
\end{itemize} & 
Pioneering large-scale VLA, jointly fine-tuned on web-based VQA and robotic datasets, unlocking advanced emergent functionalities. & 
$\times$ & $\surd$ & $\surd$ \\
\midrule
Seer \cite{seer} (2023) & 
\begin{itemize}[leftmargin=*]
  \item Vision Encoder: Visual backbone
  \item Language Encoder: Transformer-based
  \item Action Decoder: Autoregressive action prediction head
\end{itemize} & 
Efficiently predict future video frames from language instructions by extending a pretrained text-to-image diffusion model. & 
$\surd$ & $\times$ & $\surd$ \\
\midrule
Octo \cite{Octo} (2024)& 
\begin{itemize}[leftmargin=*]
  \item Vision Encoder: CNN
  \item Language Encoder: T5-base
  \item Action Decoder: Diffusion Transformer
\end{itemize} & 
First generalist policy trained on a massive multi-robot dataset (800k+ trajectories). A powerful open-source foundation model. & 
$\times$ & $\surd$ & $\times$ \\
\midrule
Open-VLA \cite{OpenVLA} (2024) & 
\begin{itemize}[leftmargin=*]
  \item Vision Encoder: DINOv2 + SigLIP
  \item Language Encoder: Prismatic-7B
  \item Action Decoder: Symbol-tuning
\end{itemize} & 
An open-source alternative to RT-2, superior parameter efficiency and strong generalization with efficient LoRA fine-tuning. & 
$\times$ & $\times$ & $\surd$ \\
\midrule
Mobility-VLA \cite{mobilityVLA} (2024) & 
\begin{itemize}[leftmargin=*]
  \item Vision Encoder: Long-context ViT + goal image encoder
  \item Language Encoder: T5-based instruction encoder
  \item Action Decoder: Hybrid diffusion + autoregressive ensemble
\end{itemize} & 
Leverages demonstration tour videos as an environmental prior, using a long-context VLM and topological graphs for navigating based on complex multimodal instructions. & 
$\surd$ & $\surd$ & $\times$ \\
\midrule
Tiny-VLA \cite{TinyVLA} (2025) & 
\begin{itemize}[leftmargin=*]
  \item Vision Encoder: FastViT with low-latency encoding
  \item Language Encoder: Compact language encoder (128-d)
  \item Action Decoder: Diffusion policy decoder (50M parameters)
\end{itemize} & 
Outpaces OpenVLA in speed and precision; eliminates pretraining needs; achieves 5x faster inference for real-time applications. & 
$\times$ & $\times$ & $\surd$ \\
\midrule
Diffusion-VLA \cite{Diffusion-VLA} (2024) & 
\begin{itemize}[leftmargin=*]
  \item Transformer-based visual encoder for contextual perception
  \item Language Encoder: Autoregressive reasoning module with next-token prediction
  \item Diffusion policy head for robust action sequence generation
\end{itemize} & 
Leverage diffusion-based action modeling for precise control; superior contextual awareness and reliable sequence planning. & 
$\times$ & $\surd$ & $\times$ \\
\midrule
Point-VLA \cite{PointVLA} (2025) & 
\begin{itemize}[leftmargin=*]
  \item Vision Encoder: CLIP + 3D Point Cloud
  \item Language Encoder: Llama-2
  \item Action Decoder: Transformer with spatial token fusion
\end{itemize} & 
Excel at long-horizon and spatial reasoning tasks; avoid retraining by preserving pretrained 2D knowledge & 
$\surd$ & $\times$ & $\times$ \\
\midrule
VLA-Cache \cite{casheVLA} (2025) & 
\begin{itemize}[leftmargin=*]
  \item Vision Encoder: SigLIP with token memory buffer
  \item Language Encoder: Prismatic-7B
  \item Action Decoder: Transformer with dynamic token reuse
\end{itemize} & 
Faster inference with near-zero loss; dynamically reuse static features for real-time robotics& 
$\times$ & $\times$ & $\surd$ \\
\midrule
$\pi_0$ \cite{PI0} (2024)& 
\begin{itemize}[leftmargin=*]
  \item Vision Encoder: PaliGemma VLM backbone
  \item Language Encoder: PaliGemma (multimodal)
  \item Action Decoder: Flow Matching
\end{itemize} & 
Employ flow matching to produce smooth, high-frequency (50Hz) action trajectories for real-time control. & 
$\times$ & $\surd$ & $\surd$ \\
\midrule
$\pi_0$ Fast \cite{fast} (2025)& 
\begin{itemize}[leftmargin=*]
  \item Vision Encoder: PaliGemma VLM backbone
  \item Language Encoder: PaliGemma (multimodal)
  \item Action Decoder: Autoregressive Transformer with FAST
\end{itemize} & 
Introduces an efficient action tokenization scheme based on the Discrete Cosine Transform (DCT), enabling autoregressive models to handle high-frequency tasks and significantly speeding up training. & 
$\times$ & $\surd$ & $\surd$ \\
\midrule
Edge-VLA \cite{edgevla} (2025)& 
\begin{itemize}[leftmargin=*]
  \item Vision Encoder: SigLIP + DINOv2
  \item Language Encoder: Qwen2 (0.5B parameters)
  \item Action Decoder: Joint control prediction (non-autoregressive)
\end{itemize} & 
Streamlined VLA tailored for edge devices, delivering 30–50Hz inference speed with OpenVLA-comparable performance, optimized for low-power, real-time deployment.& 
$\times$ & $\times$ & $\surd$ \\
\midrule
OpenVLA-OFT \cite{OpenVLA-OFT} (2025) & 
\begin{itemize}[leftmargin=*]
  \item Vision Encoder: SigLIP + DINOv2 (multi-view)
  \item Language Encoder: Llama-2 7B
  \item Action Decoder: Parallel decoding with action chunking and L1 regression
\end{itemize} & 
An optimized fine-tuning recipe for VLAs that integrates parallel decoding and a continuous action representation to improve inference speed and task success. & 
$\times$ & $\surd$ & $\surd$ \\
\midrule
Spatial-VLA \cite{spatialvla} (2025) & 
\begin{itemize}[leftmargin=*]
  \item Vision Encoder: SigLIP from PaLiGemma2 4B
  \item Language Encoder: PaLiGemma2
  \item Action Decoder: Adaptive Action Grids and autoregressive transformer
\end{itemize} 
 & 
Enhance spatial intelligence by injecting 3D information via `Ego3D Position Encoding' and representing actions with `Adaptive Action Grids'. & 
$\surd$ & $\surd$ & $\times$ \\
\midrule
MoLe-VLA \cite{MoleVLA} (2025) & 
\begin{itemize}[leftmargin=*]
  \item Vision Encoder: Multi-stage ViT with STAR router
  \item Language Encoder: CogKD-enhanced Transformer
  \item Action Decoder: Sparse Transformer with dynamic routing
\end{itemize} & 
A brain-inspired architecture that uses dynamic layer-skipping (Mixture-of-Layers) and knowledge distillation to improve efficiency. & 
$\times$ & $\times$ & $\surd$ \\
\midrule
DexGrasp-VLA \cite{dexgraspvla} (2025) & 
\begin{itemize}[leftmargin=*]
  \item Vision Encoder: Object-centric spatial ViT
  \item Language Encoder: Transformer with grasp sequence reasoning
  \item Action Decoder: Diffusion controller for grasp pose generation
\end{itemize} & 
A hierarchical framework for general dexterous grasping, using a VLM for high-level planning and a diffusion policy for low-level control. & 
$\times$ & $\surd$ & $\times$ \\
\midrule
Dex-VLA \cite{dexvla} (2025) & 
 & 
A large plug-in diffusion-based action expert and an embodiment curriculum learning strategy for efficient cross-robot training and adaptation. & 
$\times$ & $\surd$ & $\times$ \\
\end{longtable}
}

\subsection{Hierarchical versus End-to-End Decision-Making}
Hierarchical and end-to-end represent two distinct paradigms to achieve autonomous decision-making for embodied intelligence, each with unique design philosophies, implementation strategies, performance characteristics, and application domains. Herewith we make a comparison between them, as listed in Table~\ref{t3}, which outlines the key differences with respect to architecture, performance, interpretability, generalization, etc.

Hierarchical architectures decompose the decision-making process into multiple modules, each addressing the specific aspects of perception, planning, execution and feedback. The core idea is to break down complex tasks into manageable sub-tasks, to enhance debuggability, optimization, and maintenance. Hierarchical architectures excel in integrating domain knowledge (e.g., physical constraints, rules), offering high interpretability and reliability for embodied tasks. But their limitations are obvious. The separation of modules may lead to sub-optimal solutions due to improper coordination, particularly in dynamic complex environments. Manual task decomposition may hinder the adaptability to unseen scenarios and tasks.

End-to-end architectures employ a large-scale neural network, i.e., VLA, to map multimodal inputs to actions directly, without modular decomposition. VLAs are usually built on large multimodal models and trained on extensive datasets, achieving visual perception, language understanding, and action generation simultaneously. Due to highly integrated architecture, VLA minimizes error accumulation across modules and enables efficient learning via end-to-end optimization. As training on large-scale multimodal dataset, VLA possesses strong generalization to complex tasks in unstructured environments. However, the black-box nature of VLAs reduces interpretability, making it difficult to analyze the decision-making process. VLAs's performance relies heavily on the quality and diversity of training data. The computational cost of end-to-end training is also high.

{
\footnotesize
\begin{longtable}{m{0.15\textwidth} m{0.4\textwidth} m{0.4\textwidth}}
\caption{Comparison of hierarchical and end-to-end decision-making paradigms.} \label{t3} \\
\toprule
\makecell[c]{Aspect} & \makecell[c]{Hierarchical} & \makecell[c]{End-to-End} \\
\midrule
\endfirsthead
\caption{Comparison of hierarchical and end-to-end decision-making (continued).} \\
\toprule
\makecell[c]{Aspect} & \makecell[c]{Hierarchical} & \makecell[c]{End-to-End} \\
\midrule
\endhead
\bottomrule
\endfoot
\makecell[l]{Architecture} & 
  \begin{itemize}[leftmargin=*]
    \item Perception: dedicated modules (e.g., SLAM, CLIP)
    \item High-level planning: structured, language, program 
    \item Low-level execution: predefined skill lists 
    \item Feedback: LLM self-reflection, human, environment
  \end{itemize} & 
  \begin{itemize}[leftmargin=*]
    \item Perception: integrated in tokenization
    \item Planning: implicit via VLA pretraining
    \item Action generation: Autoregressive generation with diffusion-based decoders
    \item Feedback: inherent in closed-loop cycle
  \end{itemize} \\
\midrule
\makecell[l]{Performance} & 
  \begin{itemize}[leftmargin=*]
    \item Reliable in structured tasks
    \item Limited in dynamic settings
  \end{itemize} & 
  \begin{itemize}[leftmargin=*]
    \item Superior in complex, open-ended tasks with strong generalization
    \item Dependent on training data
  \end{itemize} \\
\midrule
\makecell[l]{Interpretability} & 
  \begin{itemize}[leftmargin=*]
    \item High, with clear modular design
  \end{itemize} & 
  \begin{itemize}[leftmargin=*]
    \item Low, due to black-box nature of neural networks
  \end{itemize} \\
\midrule
\makecell[l]{Generalization} & 
  \begin{itemize}[leftmargin=*]
    \item Limited, due to reliance on human-designed structures
  \end{itemize} & 
  \begin{itemize}[leftmargin=*]
    \item Strong, driven by large-scale pretraining
    \item Sensitive to data gaps
  \end{itemize} \\
\midrule
\makecell[l]{Real-time} & 
  \begin{itemize}[leftmargin=*]
    \item Low, inter-module communications may introduce delays in complex scenarios
  \end{itemize} & 
  \begin{itemize}[leftmargin=*]
    \item High, direct perception-to-action mapping minimizes processing overhead
  \end{itemize} \\
\midrule
\makecell[l]{Computational\\Cost} & 
  \begin{itemize}[leftmargin=*]
    \item Moderate, with independent module optimization but coordination overhead
  \end{itemize} & 
  \begin{itemize}[leftmargin=*]
    \item High, requiring significant resources for training
  \end{itemize} \\
\midrule
\makecell[l]{Application} & 
  \begin{itemize}[leftmargin=*]
    \item Suitable for industrial automation, drone navigation, autonomous driving
  \end{itemize} & 
  \begin{itemize}[leftmargin=*]
    \item Suitable for domestic robots, virtual assistants, human-robot collaboration
  \end{itemize} \\
\midrule
\makecell[l]{Advantages} & 
  \begin{itemize}[leftmargin=*]
    \item High interpretability
    \item High reliability
    \item Easy to integrate domain knowledge
  \end{itemize} & 
  \begin{itemize}[leftmargin=*]
    \item Seamless multimodal integration
    \item Efficient in complex tasks
    \item Minimal error accumulation
  \end{itemize} \\
\midrule
\makecell[l]{Limitations} & 
  \begin{itemize}[leftmargin=*]
    \item Sub-optimal, due to module coordination issues
    \item Low adaptability to unstructured settings
  \end{itemize} & 
  \begin{itemize}[leftmargin=*]
    \item Low interpretability
    \item High dependency on training data
    \item High computational costs
    \item Low generalization in out-of-distribution scenarios
  \end{itemize}
\end{longtable}
}

\section{Embodied Learning}
\begin{figure}[h]
  \centering
  \includegraphics[width=\linewidth]{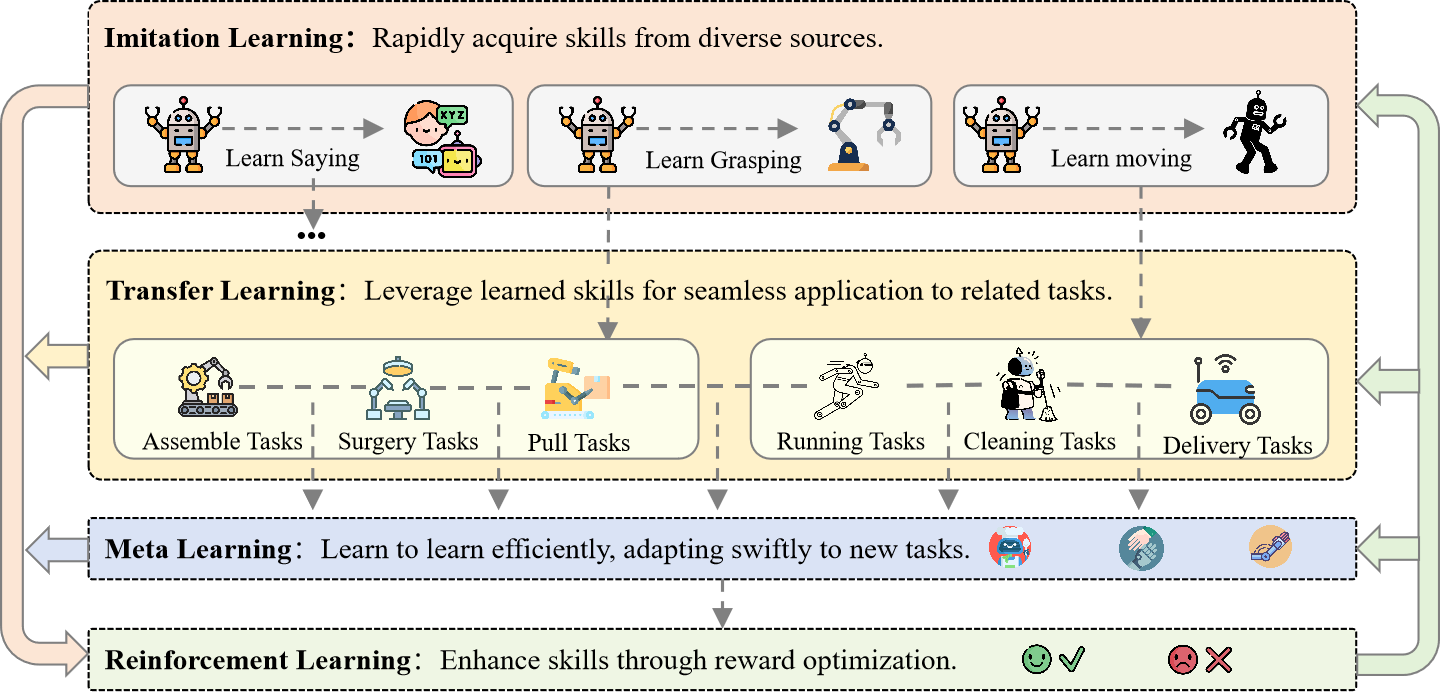}
  \caption{Embodied learning: process and methodologies.}
  \label{f12}
\end{figure}

Embodied learning aims to enable agents to acquire complex skills and refine their capabilities during interactions with environments\cite{roadmap}. Through learning and optimizing skills continuously, agents can achieve precise decision-making and real-time adaptation. This ability can be achieved through coordination of multiple learning strategies, as depicted in Fig.~\ref{f12}. Imitation learning allows agents to quickly acquire initial policies, transfer learning\cite{transferlearning} facilitates knowledge transfer across diverse tasks, meta-learning\cite{metalearning} enables agents to learn how to learn, and reinforcement learning\cite{Reinforcementlearning} optimizes policies through continuous interactions with environments. However, these learning methods still face significant technical challenges in embodied AI. Imitation learning struggles to capture complex behaviors, while reinforcement learning is often hindered by intricacies of designing effective reward functions. In recent years, the advent of Transformer and large models has motivated researchers to explore integration of large models with learning methods to overcome the limitations. In this section, we first describe the process of embodied learning and common learning methods, we then elaborate on imitation learning and reinforcement learning in more details and investigate how large models enhance these methods in embodied AI.

\subsection{Embodied Learning Methods}
Embodied agents should be able to acquire new knowledge and learn new tasks along their entire life, instead of relying on initial training datasets\cite{roadmap}. This ability is essential for the complexity and variability of the real world, where new tasks and challenges arise frequently. Embodied learning can be modeled as a goal-conditional partially observable Markov decision process, defined as an 8-tuple $(S,A,G,T,R,\Omega,O,\gamma)$, where

\begin{itemize}
  \item $S$ is the set of states of the environment. Each state encodes multimodal information, such as textual descriptions, images, or structured data.
  \item $A$ is the set of actions. Each action represents an instruction or command, often expressed in natural language.
  \item $G$ is the set of possible goals. $g\in G$ specifies a particular objective, e.g., purchase a laptop. 
  \item $T(s'|s,a)$ is the state transition probability function. For each state-action pair $(s,a)$, $T(\cdot)$ defines the probability  distribution over next states $s'\in S$. 
  \item $R:S\times A\times G\rightarrow R$ is the goal-conditional reward function, evaluating how well an action $a$ in state $s$ advances the goal . For each triplet $(s,a,g)$ , the reward can be numeric (e.g., a score) or textual (e.g., good job), providing an interactive feedback for the goal.
  \item $\Omega$ is the set of observations, which may include textual, visual, or multimodal data, representing the agent’s partial view of the state.
  \item $O(o'|s',a)$ is the observation probability function, defining the probability of observing $o'\in \Omega$ after transitioning to state $s'$ via action $a$.
  \item $\gamma\in[0,1)$ is the discount factor, balancing the immediate and the long-term rewards. It applies only when rewards are numeric.
\end{itemize}

This formulation captures the complexities of real-world scenarios, where an agent operates under partially observable stochastic dynamics. At time $t$, the agent receives an observation $o_t\in \Omega$ and a goal $g\in G$. It selects an action $a_t\in A$ according to the policy $\pi(a_t|o_t,g)$. After action execution, the environment state transitions to $s_{t+1}\sim T(s'|s_t,a_t)$, yielding an observation $o_{t+1}\sim O(o'|s_{t+1},a_t)$ and a reward $R(s_{t+1},a_t,g)$. For end-to-end decision-making, the VLA model directly encodes the policy $\pi(a|o,g)$, processing multimodal observation $o\in \Omega$ and producing action $a\in A$. For hierarchical decision-making, the high-level agent generates context-aware subgoal $g_{sub}$ via an LLM-enhanced policy $\pi_{high}(g_{sub}|o,g)$, then the low-level policy $\pi_{low}(a|o,g_{sub})$ maps the subgoal to an action $a\in A$. The low-level policy $\pi_{low}(a|o,g_{sub})$ can be learned through imitation learning or reinforcement learning. The learned policy is embedded in the model’s hierarchical architecture, and fine-tuned during training to handle specific tasks, e.g., navigation, manipulation, human-robot interaction. 

For embodied intelligence, imitation learning, reinforcement learning, transfer learning, and meta-learning all play important roles in enabling agents to act in complex real-world environments. Each learning method addresses unique challenges. Imitation learning\cite{robotlearning} allows agents to learn effective policies by emulating expert or video demonstrations, highly efficient for tasks like robotic manipulation, where high-quality data is available. But its dependence on diverse demonstrations limits adaptability to new scenarios. Reinforcement learning\cite{CRL} excels in dynamic settings through trial-and-error interactions guided by reward functions. However, it is a challenge to design proper reward functions and reinforcement learning requires high computational resources. Transfer learning\cite{transferlearning} enhances learning effect by transferring knowledge between related tasks, ideal for reusing skills. Yet it risks negative transfer when tasks differ significantly\cite{tirinzoni2018importance}. Meta-learning\cite{gupta2018meta} focuses on learning how to learn, enabling rapid adaptation to new tasks with minimal data. But it requires extensive pretraining across diverse tasks. Table \ref{t4} summaries and compares these methods in brief with respect to embodied AI. 

{
\footnotesize
\begin{longtable}{m{0.15\textwidth} m{0.25\textwidth} m{0.25\textwidth}m{0.25\textwidth}}
\caption{Comparison of learning methods wrt. embodied AI.} \label{t4}\\
\toprule
\makecell[c]{Methods} & \makecell[c]{Strengths} & \makecell[c]{Limitations}& \makecell[c]{Applications}\\
\midrule
\endfirsthead
\caption{Comparison of learning methods wrt. embodied AI (continued).}\\
\toprule
\makecell[c]{Methods} & \makecell[c]{Strengths} & \makecell[c]{Limitations}& \makecell[c]{Applications}\\
\midrule
\endhead
\bottomrule
\endfoot
\makecell[l]{Imitation\\Learning} & 
  \begin{itemize}[leftmargin=*]
    \item Rapid policy learning by mimicking expert demonstrations
    \item Efficient for tasks with high-quality data 
  \end{itemize} & 
  \begin{itemize}[leftmargin=*]
    \item Dependent on diverse, high-quality demonstrations
    \item Limited adaptability to new tasks or sparse data scenarios
  \end{itemize} &
  \begin{itemize}[leftmargin=*]
    \item Robotic manipulation
    \item Structured navigation
    \item Human-robot interaction with expert guidance
  \end{itemize}\\
\midrule
\makecell[l]{Reinforcement\\Learning} & 
  \begin{itemize}[leftmargin=*]
    \item Optimizes policies in dynamic uncertain environments via trial-and-errors
    \item Excels in tasks with clear reward signals
  \end{itemize} & 
  \begin{itemize}[leftmargin=*]
    \item Requires large samples and computational resources
    \item Sensitive to reward function and discount factor
  \end{itemize} &
  \begin{itemize}[leftmargin=*]
    \item Autonomous navigation
    \item Adaptive human-robot interaction
    \item Dynamic task optimization
  \end{itemize}\\
\midrule
\makecell[l]{Transfer Learning} & 
  \begin{itemize}[leftmargin=*]
    \item Accelerates learning by transferring knowledge between related tasks
    \item Enhances generalization in related tasks
  \end{itemize} & 
  \begin{itemize}[leftmargin=*]
    \item Risks negative transfer when tasks differ significantly
    \item Requires task similarity for effective learning
  \end{itemize} &
  \begin{itemize}[leftmargin=*]
    \item Navigation across diverse environments
    \item Manipulation with shared structures
    \item Cross-task skill reuse
  \end{itemize}\\
\midrule
\makecell[l]{Meta-Learning} & 
  \begin{itemize}[leftmargin=*]
    \item Rapid adaptation to new tasks with minimal data
    \item Ideal for diverse embodied tasks
  \end{itemize} & 
  \begin{itemize}[leftmargin=*]
    \item Demands extensive pre-training and large datasets
    \item Establishing a universal meta-policy is resource-intensive
  \end{itemize} &
  \begin{itemize}[leftmargin=*]
    \item Rapid adaptation in navigation, manipulation, or interaction across diverse tasks and environments
  \end{itemize}
\end{longtable}
}

\subsubsection{Imitation Learning}
Imitation learning is a key method in embodied learning. It enables agents to learn policies by mimicking expert demonstrations, allowing rapid acquisition of decision-making strategies for goal-oriented tasks\cite{robotlearning}. The training is supervised using datasets of expert state-action pairs $(s,a)$. The goal is to learn a policy $\pi(a|s)$ that closely replicates the expert’s behavior by minimizing the negative log-likelihood of the expert’s actions. Therefore, its objective function can be defined as:
\begin{equation}
    \mathcal{L}(\pi)=-\operatorname{E}_{\tau\sim\mathrm{PD}}[\log\pi(\mathrm{a}|s)]
\end{equation}
where $D$ is the set of expert demonstrations. Each demonstration $\tau_{i}$ consists of a sequence of state-action pairs $(s_t,a_t)$ of length $L$:
\begin{equation}
    \tau_i=[(s_1,a_1), \cdots, (s_t,a_t),\cdots,(s_L,a_L)]
\end{equation}
In continuous action spaces, the policy $\pi(\cdot)$ is often modeled as a Gaussian distribution, and the objective function is approximated using mean squared error (MSE) between the predicted and the expert actions. Imitation learning is highly sample-efficient, as it avoids extensive trial-and-error, but it is highly dependent on the quality and coverage of the demonstration data, having difficulties in unseen scenarios. A hybrid approach combining imitation learning with reinforcement learning can address this limitation by initializing the policy with imitation learning and refining it with reinforcement learning, enhancing robustness to unseen situations. 

\subsubsection{Reinforcement Learning}
Reinforcement learning is currently a dominating method in embodied learning. It enables agents to learn policies by interacting with environments through trial-and-error, making it well-suited for dynamic and uncertain settings\cite{CRL}. At each time step $t$, the agent observes a state $s$ and selects an action $a$ according to its policy $\pi(a|s)$. After action execution, the agent receives a reward $r$ from the reward function $R(s,a,g)$, the environment transitions to a new state $s'$ according to the state transition probability $T(s'|s,a)$, yielding an observation $o'\sim O(o'|s', a)$. The objective function of reinforcement learning is to maximize the expected cumulative reward:
\begin{equation}
    \mathcal{J}(\pi)=E_{\pi,T,O}\left(\sum_{t=0}^{\infty}\gamma^{t}R(s,a,g)\right)
\end{equation}
where $\gamma \in[0,1)$ is a discount factor balancing immediate and long-term rewards. Reinforcement learning excels in optimizing policies for complex tasks, but requires extensive explorations, which is computationally costly. A hybrid approach combining imitation learning and reinforcement learning can improve the issue, where imitation learning provides initial policies to reduce explorations, and reinforcement learning refines them through interactions with environments. 

\subsubsection{Transfer Learning}
In scenarios where training from scratch requires extensive samples and time, transfer learning can be applied to alleviate the effort\cite{transferlearning}. It allows agents to accelerate learning on related target tasks using the knowledge from source tasks. By transferring learned policies, features, or representations from source tasks, the agents enhance efficiency and generalization on target tasks. Given a source task with a state-action space defined by state $s\in S$, action $a \in A$, and the policy $\pi(a|s)$, transfer learning adapts the source policy $\pi_s$ to a target task with different dynamics or goals. The objective is to minimize the divergence between the source policy $\pi_s$ and the target policy $\pi_t$, by fine-tuning the policy using a small number of target task data. The process is guided by the task-specific loss of the target task, constrained by the Kullback-Leibler (KL) divergence for policy alignment:
\begin{equation}
\theta_{t}^{*}=\arg \min_{\theta_{t}} \mathbb{E}_{s \sim S_{t}}\left(D_{K L}\left(\pi_{s}\left(\cdot \mid s ; \theta_{s}\right) \| \pi_{t}\left(\cdot \mid s ; \theta_{t}\right)\right)\right)+\lambda \mathcal{L}_{t}\left(\theta_{t}\right)
\end{equation}
where $\theta_t^*$ represents the optimal policy parameters for the target task, $\theta_s$ and $\theta_t$ are the parameters of the source and the target policies, respectively, $D_{KL}$ measures the divergence between the source policy $\pi_s$ and the target policy $\pi_t$, $\mathcal{L}_t$ is the task-specific loss of the target task, and $ \lambda$ is a regularization parameter balancing policy alignment and task performance. This process ensures that the transferred knowledge aligns with the target task’s state transition probability $T(s'|s,a)$ and reward function $R(s,a,g)$. In embodied settings, transfer learning enables agents to reuse learned behaviors across different environments and goals, reducing training time. However, significant disparities between source and target tasks may lead to negative transfer, where performance degrades due to mismatched knowledge.

\subsubsection{Meta-Learning}
Meta-learning can also be used in embodied AI, enabling agents to learn how to learn\cite{gupta2018meta,metalearning}, so that they can swiftly infer optimal policies for new tasks from a small number of samples. At each time step $t$, the agent receives an observation $o\in \Omega$ and a goal $g$, and selects an action $a$ according to the meta-policy that adapts to task-specific dynamics defined by the state transition probability $T(s'|s,a)$ and the reward function $R(s,a,g)$. The objective is to optimize the expected performance across tasks by minimizing the loss on task-specific data. In the context of Model-Agnostic Meta-Learning (MAML)\cite{MAML}, this is achieved by learning an initial set of model parameters $\theta$ that can be adapted quickly to new tasks with minimal updates. Specifically, for a set of tasks $\mathcal{T}_i$, MAML optimizes the meta-objective as:
\begin{equation}
\begin{split}
& \theta^{*}=\arg \min_{\theta} \sum_{\mathcal{T_i}} \mathcal{L}_{\mathcal{T}_{i}}\left(f_{\theta_{i}}\right)   \\
& \theta_{i}=\theta-\alpha \nabla_{\theta}\mathcal{L}_{\mathcal{T}_{i}}\left(f_{\theta}\right)
\end{split}
\end{equation}
where $\theta^*$ represents the optimal meta-policy parameters, $\mathcal{L}_{\mathcal{T}_i}$ is the task-specific loss, $f_\theta$ is the model parameterized by $\theta$, $\theta_i$ is the task-specific parameters after gradient updates with the learning rate $\alpha$, and the outer optimization minimizes the expected loss across tasks after adaptation. Meta-learning can enable agents to quickly adapt to new tasks by fine-tuning a pretrained model using a small number of demonstrations or interactions. The meta-policy can be embedded in a large model and refined during training to handle diverse tasks. Despite the strengths, meta-learning requires substantial pretraining and large samples across tasks, posing challenges in establishing a universal learning policy, particularly when tasks vary significantly in state spaces or dynamics. 

\subsection{Imitation Learning Empowered by Large Models}
The primary objective of imitation learning is to enable agents to achieve expert-level performance by mimicking their actions from demonstrators. Imitation learning can be realized by different approaches, including behavior cloning\cite{florence2022implicit}, inverse reinforcement learning\cite{ng2000algorithms}, generative adversarial imitation learning\cite{ho2016generative}, and hierarchical imitation learning\cite{belkhale2023hydra}, each contributing to the construction of policy networks. Among these approaches, behavior cloning stands out as the most important one, which formulates imitation learning as a supervised regression task. Given observations $o\in \Omega$ and goals $g\in G$, the policy network $\pi$ predicts the expected actions $a\in A$. 

The policy network $\pi$ needs to accurately map observations $o$ and goals $g$ to actions $a$, so as to ensure high imitation fidelity, even in complex, dynamic, partially observable environments. Beyond replication, imitation learning seeks to endow agents with the ability to generalize to unseen states, goals, or environments. The generalization ability is essential for real-world applications, such as robotic manipulation, autonomous navigation, and human-robot interaction, where environmental dynamics and task requirements often deviate from training scenarios. Additionally, imitation learning aims to ensure robustness against distribution shifts, where small errors in action predictions do not accumulate to cause significant deviations from the expert’s trajectories, particularly in stochastic or dynamic settings. Finally, imitation learning strives for sample efficiency, trying to enable agents to learn effective policies from a limited number of expert demonstrations, reducing the dependency on extensive, high-quality datasets. 

\begin{figure}
  \centering
  \includegraphics[width=\linewidth]{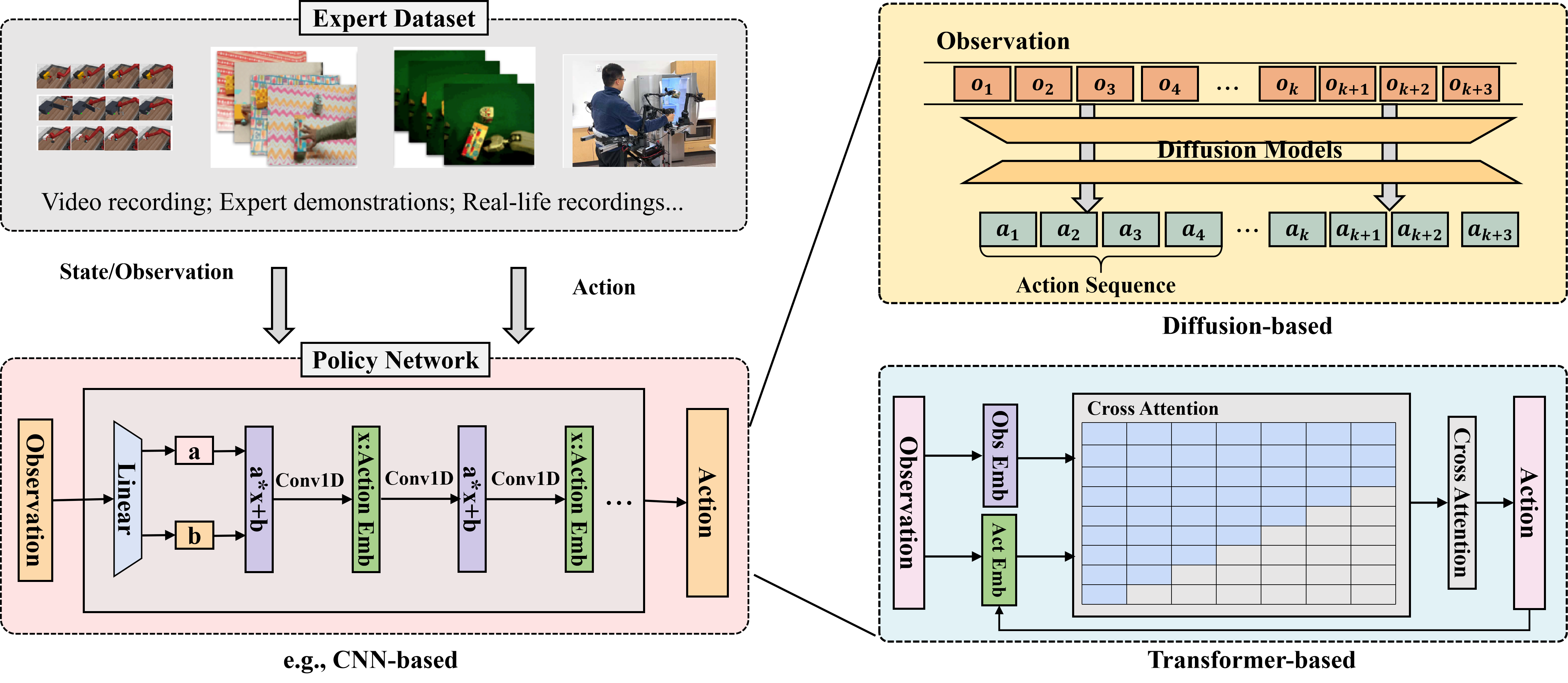}
  \caption{Imitation learning empowered by diffusion models or Transformers.}
  \label{f13}
\end{figure}

Behavior cloning still have difficulties in meeting these requirements to build a robust policy\cite{robotlearning}. Its dependence on high-quality expert demonstrations hampers generalization to unseen states or goals. Expert demonstrations often exhibit randomness, multimodality and complexity, which are difficult for policy networks to capture, leading to compromised imitation fidelity and degraded performance. Recent advances in large models have significantly enhanced behavior cloning, addressing its inherent limitations. As illustrated in Fig.~\ref{f13}, \textbf{large models empower imitation learning in the following aspects: (1) use diffusion models to construct policy networks; (2) use Transformer to construct policy networks.}

\subsubsection{Diffusion-based Policy Network}
Diffusion models are excellent in handling complex multimodal distributions\cite{diffusion}. They can be used to generate diverse action trajectories, so as to enhance robustness and expressiveness of policies. Recent researches have started to integrate diffusion models into policy networks to overcome the limitations of traditional imitation learning. Pearce\cite{Pearce} proposes a diffusion model-based imitation learning framework, which incorporates diffusion models into policy networks. By optimizing expert demonstrations iteratively through noise addition and removal, the framework can capture the diversity of action distributions and generate diverse action sequences. DABC\cite{DABC} adopts a two-stage process to train the policy network with the empowerment of a diffusion model. It first pretrains a base policy network through behavior cloning, then refines the modeling of action distributions via a diffusion model. Diffusion Policy\cite{DiffusionPolicy} proposes a policy network that takes the diffusion model as the decision model for vision-driven robot tasks. It uses visual input and robot’s current state as conditions, employs U-Net as denoising network, predicts denoising steps based on visual features and state vectors, thereby generating continuous action sequences. To enhance the spatial perception capabilities of the policy network, 3D-Diffusion\cite{3D-Diffusion} proposes a diffusion policy framework based on 3D inputs. It uses simple 3D representations as inputs, leverages a diffusion model to generate action sequences, thus improving the generalization of visual motion policies by capturing spatial information. Compared to 2D policy networks, 3D-Diffusion can better understand geometric relationships and spatial constraints in 3D environments.

\subsubsection{Transformer-based Policy Network}
Transformer architectures can empower imitation learning by treating expert trajectories as sequential data and utilizing self-attention mechanisms to model dependencies between actions, states and goals. This end-to-end approach minimizes error accumulation in intermediate steps, enhancing consistency and accuracy of policies. Google’s RT-1\cite{RT-1} is the first to demonstrate the potential of Transformers in robot control. By combining a large-scale, diverse dataset (130k+ trajectories, 700+ tasks) with a pretrained vision-language model, it significantly improves task generalization for unseen tasks and scenes. Subsequent work RT-Trajectory\cite{RT-Trajectory} introduces the “trajectory sketch” method, incorporating low-level visual cues to enhance the task generalization capabilities of end-to-end Transformers. Stanford University’s ALOHA\cite{ALOHA} utilizes the encoding-decoding structure of Transformers to generate robotic arm action sequences from multi-view images, achieving precise double-arm operations with low-cost hardware. Its follow-up research uses an action chunking strategy to predict multi-step action sequences, significantly improving the stability and consistency of long-term tasks. Mobile ALOHA\cite{MobileALOHA} extends the original task to whole-body coordinated mobile operation tasks, introducing a mobile platform and a teleoperation interface to handle more complex dual-arm tasks. For 3D spatial operations, HiveFormer\cite{ALOHA} and RVT\cite{RVT} utilize multi-view data and CLIP for visual-language feature fusion and directly predict 6D grasp poses, achieving state-of-the-art performance on RLBench and real-world robotic arm tasks, highlighting the advantages of Transformers in complex spatial modeling. To grasp deformable objects (e.g., fabric or soft materials), Man proposes a Transformer framework combining visual and tactile feedback, optimizing grasp parameters through exploratory actions. Google’s RoboCat\cite{RoboCat} employs cross-task, cross-entity embodied imitation learning, integrating VQ-GAN\cite{VQ-GAN} to tokenize visual inputs, utilizing Decision Transformer to predict actions and observations, achieving rapid policy generalization with only a few samples. RoboAgent\cite{RoboAgent} adopts a similar encoding-decoding structure, fusing vision, task descriptions, and robot states to minimize action sequence prediction errors. CrossFormer\cite{CrossFormer} proposes a Transformer-based imitation learning architecture for cross-embodied tasks, trained on large-scale expert data to unify the processing of manipulation, navigation, mobility, and aerial tasks, demonstrating the potential of multi-task learning.

\subsection{Reinforcement Learning Empowered by Large Models}
Through interactions with environments, reinforcement learning\cite{Reinforcementlearning} allows agents to develop optimal control strategies, adapt to diverse unseen scenarios, maintain robustness in dynamic environments, and learn from limited data, thereby enabling sophisticated tasks in the real world. Initially reinforcement learning is based on fundamental techniques, such as policy search and value function optimization, exemplified by Q-learning\cite{Q-Learning} and State Action Reward State Action (SARSA)\cite{SARSA}. With the dominance of deep learning, reinforcement learning is integrated with deep neural networks, known as Deep Reinforcement Learning (DRL). DRL allows agents to learn intricate policies from high-dimensional inputs, leading to significant accomplishments, e.g., AlphaGo\cite{AlphaGo} and Deep Q-Network (DQN)\cite{DQN}. DRL enables agents to learn autonomously in new environments, without explicit human interventions, thus allowing for wide applications ranging from gaming to robot control and beyond. Subsequent advances further improve the learning effects. Proximal Policy Optimization (PPO)\cite{PPO} improves stability and efficiency of policy optimization through a clipped probability ratio. Soft Actor-Critic (SAC)\cite{SAC} improves exploration and robustness by incorporating a maximum entropy framework.  

\begin{figure}[h]
  \centering
  \includegraphics[width=0.9\linewidth]{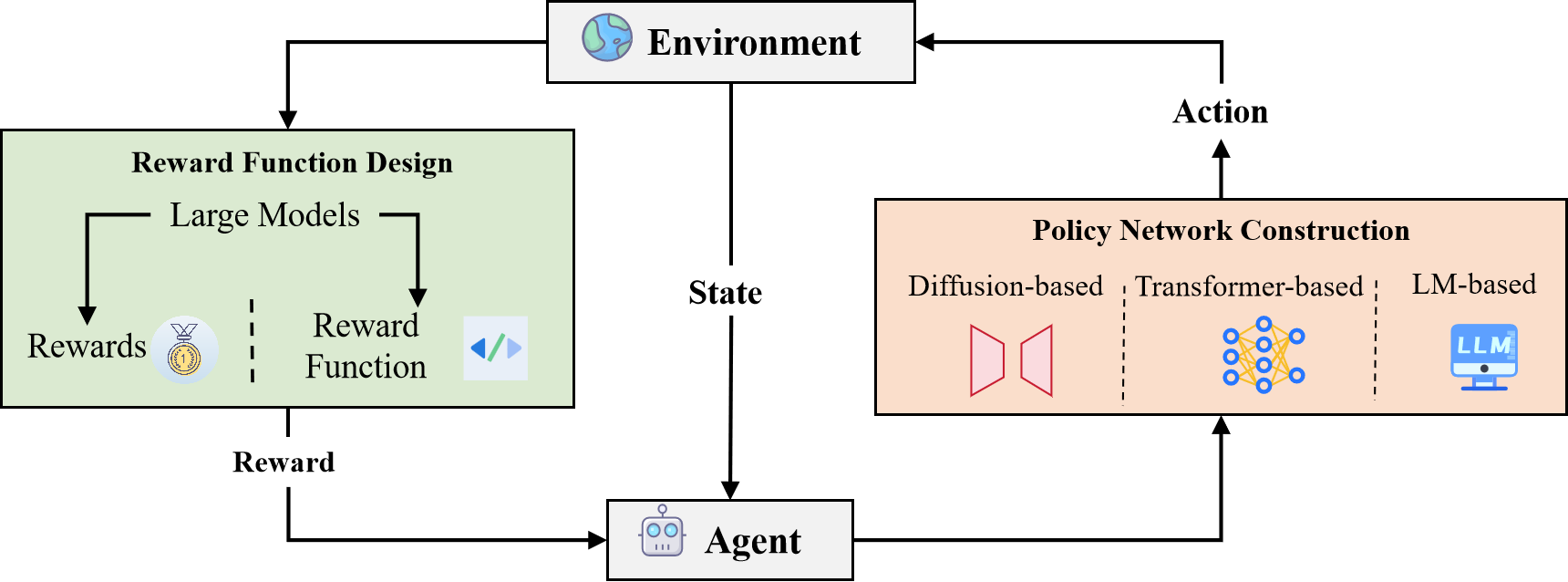}
  \caption{Reinforcement learning empowered by large models. }
  \label{f14}
\end{figure}

Despite these progresses, reinforcement learning still faces limitations in constructing policy networks $\pi$ and designing reward functions $R(s,a,g)$. \textbf{Recent advances in large models empower reinforcement learning in the following aspects: (1) improve reward function design; (2) optimize policy network construction by modeling complex action distributions.} They are as illustrated in Fig.~\ref{f14}.

\subsubsection{Reward Function Design}
Designing reward function has been a challenge for reinforcement learning\cite{eschmann2021reward}, due to its complexity and task-specific nature. Traditional reward functions are designed manually by domain experts, requiring comprehensive consideration of factors and expertise, such as task completion, energy consumption, safety, and weigh of each factor, which is quite difficult. Manual design often leads to sparse or poorly scaled rewards, causing issues like reward hacking, where the agent exploits unintended signals to maximize the reward without achieving the intended goal. 

Large models offer a promising solution by generating \textbf{(1) reward signals $r$} or \textbf{(2) reward function $R(s,a,g)$}, reducing reliance on manual design and capturing complex multimodal feedback. Kwon et al. and Language to Rewards (L2R)\cite{L2R} introduce zero-shot and few-shot approaches respectively, utilizing GPT-3 to produce reward signals directly from textual behavior prompts, translating high-level goals to hardware-specific control policies. However, their sparse rewards restrict the usage in complex tasks and successful generation depends heavily on precise prompts or specific templates. Text2Reward\cite{Text2Reward} improves this by generating dense interpretable Python reward functions from environment descriptions and examples, iteratively refining them via human feedback, achieving high success rates across robotic manipulation and locomotion tasks. Eureka\cite{Eureka} leverages GPT-4 to create dense rewards from task and environment prompts. It mitigates Text2Reward’s reliance on human feedback by employing an automated iterative strategy for reward function optimization, surpassing human-crafted rewards. Further, Auto MC-Reward\cite{automc-Reward} implements full automation with a multi-stage pipeline for Minecraft, where a reward designer generates reward signals, a validator ensures quality, and a trajectory analyzer refines rewards through failure-driven iterations. Auto MC-Reward boosts the efficiency significantly, but its domain-specific focus constrains its generalization compared to Eureka and Text2Reward. 

\subsubsection{Policy Network Construction}
\begin{figure}[h]
  \centering
  \includegraphics[width=\linewidth]{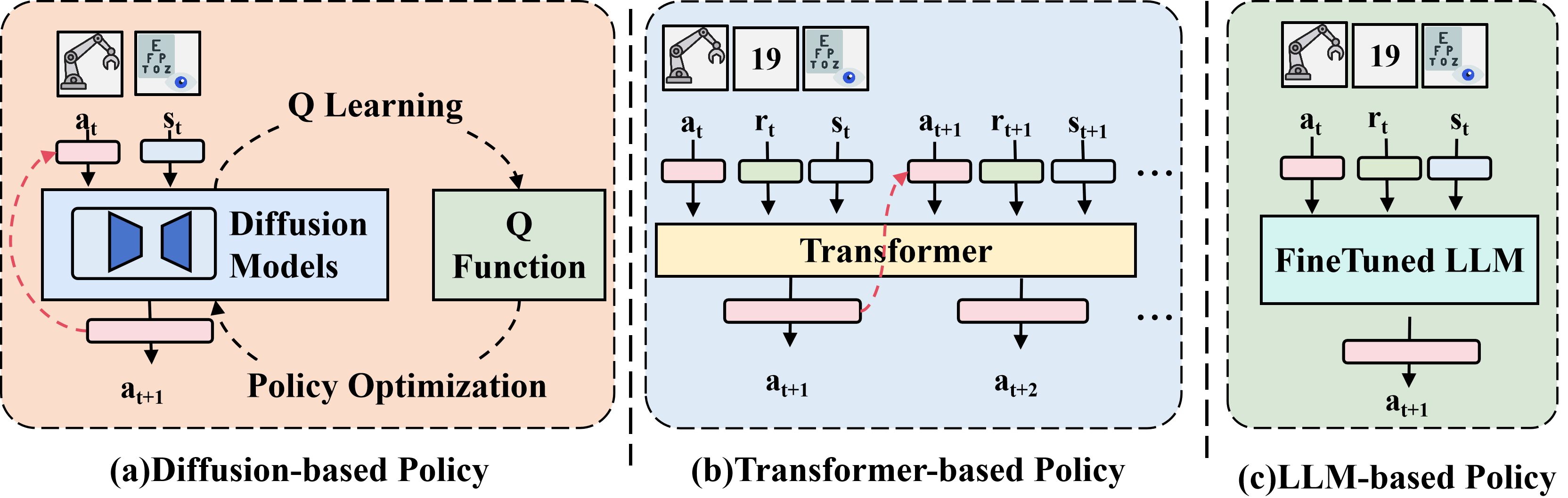}
  \caption{Policy network construction empowered by large models. }
  \label{f15}
\end{figure}

Offline reinforcement learning learns\cite{lange2012batch} optimal policies from pre-collected datasets without online interactions. But the reliance on static datasets may result in errors for actions absent from the datasets. Policy regularization may mitigate this problem by constraining deviations from behavior policy. But limitations in policy expressiveness and sub-optimal regularization methods may cause under-performance. To enhance expressiveness and adaptability of offline reinforcement learning, researchers propose to \textbf{exploit (1) diffusion models, (2) Transformer-based architectures, and (3) LLMs to empower the construction of policy networks}, as illustrated in Fig.~\ref{f15}. 

\paragraph{Policy network construction with diffusion models} Diffusion models\cite{diffusion} enhance policy expressiveness by modeling complex action distributions through iterative noising and denoising. Diffusion-QL\cite{Diffusion-QL} employs diffusion models as foundation policy, modeling action distributions and training to maximize value function objectives within the Q-learning framework. This approach generates high-reward policies that fit multimodal or non-standard action distributions in offline datasets. However, diffusion models require a large number of denoising steps to produce actions from fully noised states. To alleviate the effort, EDP\cite{EDP} introduces an efficient sampling method that reconstructs actions from intermediate noised states in a single step, significantly reducing the computational overhead. EDP can integrate with various offline reinforcement learning frameworks, enhancing sampling efficiency while maintaining policy expressiveness.

\paragraph{Policy network construction with Transformer-based architectures} Transformer-based architectures leverage the self-attention mechanism to capture long-term dependencies in trajectories, thereby improving policy flexibility and accuracy. Decision Transformer\cite{DecisionTransformer} re-frames offline reinforcement learning as a conditional sequence modeling problem, treating state-action-reward trajectories as sequential inputs, and applies supervised learning to generate optimal actions from offline datasets. Based on it, Prompt-DT\cite{Prompt-DT} enhances generalization in few-shot scenarios by incorporating prompt engineering, using trajectory prompts with task-specific encoding to guide action generation for new tasks. To improve adaptability in dynamic environments, Online Decision Transformer (ODT)\cite{ODT} pretrains Transformer via offline reinforcement learning to learn sequence generation, then fine-tunes it through online reinforcement learning interactions. Q-Transformer\cite{Q-Transformer} integrates Transformer’s sequence modeling with Q-function estimation, learning Q-values autoregressively to generate optimal actions. In multi-task offline reinforcement learning, Gato\cite{Gato} adopts a Transformer-based sequence modeling approach, but it relies heavily on dataset optimality and incurs high training cost due to large parameters. 

\paragraph{Policy network construction with LLM} Built upon the sequence modeling capabilities of Transformers, LLM introduces a new paradigm by leveraging pretrained knowledge to streamline offline reinforcement learning tasks. GLAM\cite{GLAM} uses LLM as policy agents, generating executable action sequences for language-defined tasks, which are optimized online via PPO with contextual memory for improved sequence consistency in long-horizon planning. LaMo\cite{LaMo} employs GPT-2 as base policy, which is fine-tuned with LoRA to preserve prior knowledge, converting state-action-reward sequences into language prompts for task-aligned policy generation. Reid\cite{Reid} explores LLM’s transferability using pretrained BERT, which is fine-tuned for specific tasks and augmented by external knowledge bases. Evaluations on D4RL benchmarks\cite{D4RL} show that Reid outperforms Decision Transformer meanwhile reducing training time, demonstrating LLM’s efficiency in offline reinforcement learning.

\section{World Models}
World models serve as internal simulations or representations of environments. With world models, intelligent systems can anticipate future states, comprehend cause-and-effect relationships, and make reasonable decisions, without solely relying on real-world interactions, which is expensive and often infeasible. Providing a rich cognitive framework, world models facilitate more efficient learning, decision-making, and adaptation in complex dynamic environments, thereby enhancing agents’ capacities to execute sophisticated tasks. In this section, we investigate the design of world models and examine how they contribute to decision-making and embodied learning.

\begin{figure}
  \centering
  \includegraphics[width=\linewidth]{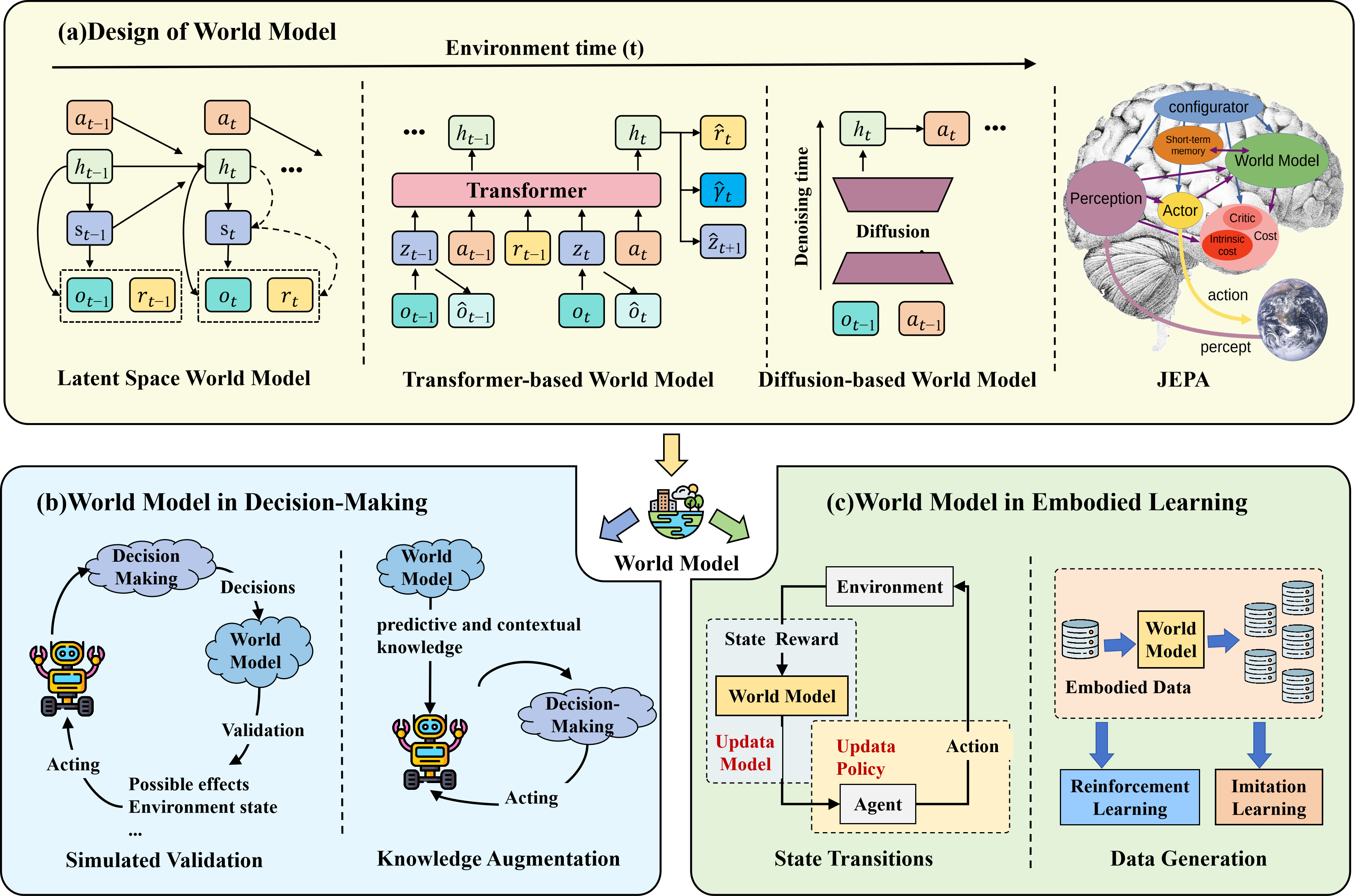}
  \caption{World models and applications in decision-making and embodied learning. }
  \label{f16}
\end{figure}

\subsection{Design of World Models}
The concept of world model can be traced back to reinforcement learning\cite{worldmodel}. Traditional RL relies on repeated agent-environment interactions, incurring high computational costs, hence impractical in data-scarce or complex scenarios. Instead of learning behaviors solely through repeated interactions, world models enable agents to learn in a simulated environment. This approach is especially valuable in data-scarce or complex scenarios. In terms of design, current world models can be categorized into four types: latent space world model, Transformer-based world model, diffusion-based world model, and joint embedding predictive architecture, as shown in the upper part of Fig.~\ref{f16}.

\subsubsection{Latent Space World Model}
The latent space world model is represented by Recurrent State Space Model (RSSM)\cite{ha2018recurrent,hafner2019dream}, which facilitates predictions in latent spaces. RSSM learns dynamic environment models from pixel observations and plans actions in encoded latent spaces. By decomposing the latent state into stochastic and deterministic parts, RSSM considers both deterministic and stochastic factors of the environment. Due to RSSM's exceptional performance in continuous control tasks of robots, many works based on RSSM emerged. PlaNet\cite{PlaNet} employs RSSM with a Gated Recurrent Unit (GRU) and a Convolutional Variational AutoEncoder (CVAE), leveraging CNN for latent dynamics and model predictive control. Dreamer\cite{DreamerV1} advances it by learning the actor and the value networks from latent representations. Dreamer V2\cite{DreamerV2} further uses the actor-critic algorithm to learn the behaviors purely from the imagined sequences generated by the world model, achieving performance comparable to human players on the Atari 200M benchmark. Dreamer V3\cite{DreamerV3} enhances stability with symlog predictions, layer normalization, and normalized returns via exponential moving average, outperforming specialized algorithms in continuous control tasks.

\subsubsection{Transformer-based World Model}
Latent space world models usually rely on CNN or Recurrent Neural Networks (RNN), hence they face challenges when operating in high-dimensional, continuous, or multimodal environments. Transformer-based world models offer a powerful alternative. They leverage the attention mechanism to model multimodal inputs, overcoming the limitations of CNN and RNN, exhibiting superior performance, particularly in complex memory-interaction tasks. IRIS\cite{IRIS} is one of the first to apply Transformer in world models, where agents learn skills within an autoregressive Transformer-based world model. IRIS tokenizes images using a Vector Quantized Variational Autoencoder (VQ-VAE) and employs an autoregressive Transformer to predict future tokens, showing exceptional performance in low-data Atari 100k settings. Google’s Genie\cite{Genie}, built on spatial-temporal Transformer\cite{ST-Transformer} and trained on vast unlabeled Internet video datasets via self-supervised learning, outperforms traditional RSSM. Genie provides a new paradigm for manipulable, generative, interactive environments, highlighting the transformative potential of Transformer. TWM\cite{TWM} proposes a Transformer-XL-based world model. It migrates Transformer-XL’s segment-level recurrence mechanism to world model, enabling the capture of long-term dependencies between the states of the environment. To further enhance efficiency, TWM trains a model-free agent within the latent imagination, avoiding full inference during runtime. STORM\cite{STORM} utilizes a stochastic Transformer, thus not relying on look-ahead search on Atari 100k benchmark. It fuses state and action into a single token, which improves training efficiency and matches the performance of Dreamer V3 on Atari 100k benchmark. These Transformer-based world models discretize states, actions, and observations into sequences, leveraging self-attention to capture long-term dependencies, significantly improving prediction accuracy, sample efficiency, and adaptability across tasks.

\subsubsection{Diffusion-based World Model}
Diffusion-based world models, exemplified by OpenAI’s Sora\cite{Sora}, represent a significant advance in generating predictive video sequences in the original image space. Distinct from latent space world models and Transformer-based world models, Sora leverages an encoding network to convert videos and images into tokens, followed by a large-scale diffusion model that applies noising and denoising processes to these tokens, subsequently mapping them back to the original image space to produce multi-step image predictions based on language descriptions. This capability makes Sora highly applicable to embodied tasks. For example, Sora can generate trajectory videos for agents across future time steps using robot task descriptions and trajectory priors, enhancing model-based reinforcement learning. UniPi\cite{UniPi} employs diffusion models to model agent trajectories in image space, generating future key video frames from language inputs and initial images, followed by super-resolution in time series to create consistent, high-quality image sequences. UniSim\cite{UniSim} further improves trajectory prediction by jointly training diffusion models on Internet data and robot interaction videos, enabling prediction of long-sequence video trajectories for both high-level and low-level task instructions. 

\subsubsection{Joint Embedding Predictive Architecture}
The above data-driven world models excel in natural language processing tasks, but lack real-world common sense due to their reliance on training data. The Joint Embedding Predictive Architecture (JEPA)\cite{JEPA}, proposed by Yann LeCun at Meta, is a groundbreaking approach to overcome the limitations regarding common sense. Inspired by human brains' efficient learning, JEPA introduces hierarchical planning and self-supervised learning in a high-level representation space. Hierarchical planning breaks complex tasks into multiple abstraction levels, each addressing specific sub-tasks to simplify decision-making and control, focusing on semantic features rather than pixel-level outputs as traditional generative models do. Through self-supervised learning, JEPA trains networks to predict missing or hidden input data, enabling pretraining on large unlabeled datasets and fine-tuning for diverse tasks. JEPA’s architecture comprises a perception module and a cognitive module, forming a world model using latent variables to capture essential information while filtering redundancies, supporting efficient decision-making and future scenario planning. By incorporating a dual-system concept, JEPA balances “fast” intuitive reactions with “slow” deliberate reasoning. This combination of hierarchical planning, self-supervised learning, and a robust world model makes JEPA a scalable, cognition-inspired framework for complex, real-world environments.

\subsection{World Model in Decision-Making}
World models can provide agents with powerful internal representations, enabling them to predict environmental dynamics and potential outcomes before taking actual actions. For decision-making, it plays two main roles: \textbf{(1) simulated validation} and \textbf{(2) knowledge augmentation}, as shown in the left of Fig.~\ref{f16}. Through these mechanisms, world models can improve agents’ ability to plan and execute tasks in complex dynamic environments significantly.

\subsubsection{World Model for Simulated Validation}
In robotics, testing decisions can be incredibly expensive and time-consuming, especially the sequential and long-term tasks, where current decisions profoundly affect future performances. World models can mitigate this issue by enabling simulated validation, allowing agents to “try out” actions and observe the likely consequences without real-world commitments. This simulated validation dramatically shortens iteration time and facilitates safe testing of corner cases or high-risk scenarios that would be impractical otherwise. The ability to predict how actions affect future environmental states helps agents to identify and avoid potential mistakes, ultimately optimizing performance. NeBula\cite{NeBula} constructs probabilistic belief spaces using Bayesian filtering, which allows robots to reason effectively across diverse structural configurations, including unknown environments, offering a sophisticated way to predict outcomes under uncertainty. UniSim\cite{UniSim} is a generative simulator for real-world interactions, which can simulate the visual outcomes of both high-level instructions and low-level controls. It contains a unified generative framework that takes actions as input, integrating diverse datasets across different modulations. 

\subsubsection{World Model for Knowledge Augmentation}
To successfully accomplish real-world tasks, agents often require rich knowledge and environmental common sense. World models can augment agents with predictive and contextual knowledge essential for strategy planning. By predicting future environmental states or enriching agents’ understanding of the world, world models enable agents to anticipate outcomes, avoid mistakes, and optimize performance over time. The World Knowledge Model (WKM)\cite{WKM} imitates humans’ mental world knowledge by providing global prior knowledge before a task and maintaining local dynamic knowledge during the task. It synthesizes global task knowledge and local state knowledge from experts and sampled trajectories, achieving superior planning performance when integrated with LLM. Agent-Pro\cite{Agent-Pro} transforms an agent’s interactions with its environment (especially with other agents in interactive tasks) into “beliefs”. These beliefs represent the agent’s social understanding of the environment and inform subsequent decisions and behavioral strategy updates. GovSim\cite{GovSim} explores the emergence of cooperative behaviors within societies of LLM agents. These agents gather information about the external world and other agents’ strategies through multi-agent conversations, implicitly forming their own high-level insights and representations of the world model.

\subsection{World Model in Embodied Learning}
Apart from decision making, world models can enable agents to learn new skills and new behaviors efficiently. Different from model-free reinforcement learning, which often incurs high computational costs and data inefficiency from direct agent-environment interactions, model-based reinforcement learning leverages world models to streamline the learning process by \textbf{(1) simulating state transitions} and \textbf{(2) generating data}, as shown in the right of Fig.~\ref{f16}.

\subsubsection{World Model for State Transitions}
Traditional reinforcement learning is model-free and learns directly from agent-environment interactions, which is computationally intensive and impractical in data-scarce or complex scenarios. Model-based reinforcement learning alleviates these limitations by leveraging a world model that explicitly captures state transitions and dynamics, allowing agents to enhance their learning process from simulated environments for safe, cost-effective and data-efficient training. The world model creates virtual representations of the real world, so that the agents can explore hypothetical actions and refine policies without inherent risks or costs associated with real-world interactions. RobotDreamPolicy\cite{RobotDreamPolicy} learns a world model and develops the policy within it, drastically reducing interactions with the real environment. DayDreamer\cite{DayDreamer} leverages Dreamer V2, an RSSM-based world model, to encode observations into latent states and predict future states, achieving rapid skill learning in real robots with high sample efficiency. SWIM\cite{SWIM} takes a step further by utilizing Internet-scale human video data to understand rich human interactions and gain meaningful affordances. It is initially trained on a large dataset of egocentric videos and then fine-tuned with robot data to adapt to robot domains. Subsequently, behaviors for specified tasks can be learned efficiently with this world model. 

\subsubsection{World Model for Data Generation}
Beyond empowering learning and optimizing policies, world models, especially diffusion-based world models, can be used to synthesize data, which is particularly valuable for embodied AI as collecting diverse and extensive real-world data is challenging. Diffusion-based World models can synthesize realistic trajectory data, state representations, and dynamics, thereby augmenting existing datasets or creating entirely new datasets to enhance learning processes. SynthER\cite{SynthER} utilizes diffusion-based world models to generate low-dimensional offline RL trajectory data to augment the original datasets. Their evaluations demonstrate that the diffusion models can effectively learn state representations and dynamical equations from trajectory data. MTDiff\cite{MTDiff} applies diffusion-based world models to generate multi-task trajectories, using expert trajectories as prompts to guide the generation of agent trajectories that align with specific task objectives and dynamics. VPDD\cite{VPDD} trains the trajectory prediction world models using a large-scale human operation dataset, and then fine-tunes the action generation module with only a small amount of labeled action data, significantly reducing the need of extensive robot interaction data for policy learning.

\section{Challenge and Future Prospects}
Embodied intelligence presents unprecedented opportunities far beyond virtual confines. However, challenges still exist. In this section, we discuss the major open issues of scarcity of high-quality embodied data, continual learning for long-term adaptability, computational and deployment efficiency, and sim-to-real gap. By investigating the core challenges, we point out the potential research directions towards robust, adaptable, and truly intelligent embodied systems.

\subsection{Scarcity of Embodied Data}
Training embodied intelligent agents requires vast and diverse datasets. RT-X\cite{openx} collected robot arm data from over 60 laboratories and built the open X-Embodiment dataset. AutoRT\cite{autort} proposed systems for automatic data collection in new environments. However, real-world robotic data is still insufficient. The reasons lie in the immense diversity in robot designs, the intricate nature of real-world interactions, and the specific requirements of various tasks, etc. The state-of-the-art embodied datasets, e.g., VIMA\cite{VIMA} (with 650,000 demonstrations) and RT-1\cite{GLAM} (with 130,000 demonstrations), are still dwarfed by their vision-language counterparts, such as LAION-5B (with 5.75 billion text-image pairs). To address the issue of embodied data scarcity, researchers have tried various solutions.
\begin{enumerate}[leftmargin=*]
\item Leverage world models, particularly diffusion-based world models, to synthesize new data from existing agent experiences. SynthER\cite{SynthER} utilizes a diffusion-based world model to synthesize data and augment offline RL trajectory datasets, significantly enhancing performance in both offline and online settings. 
\item Integrate large human datasets. Ego4D\cite{Ego4D} provides rich real-world dynamics and observations gleaned from Internet-sourced videos. This approach helps improve the contextual understanding for robotic tasks by leveraging common human behaviors and interactions. However, due to the morphological differences between humans and robots, directly transferring human actions to robots often results in misalignment and reduced transferability. 
\end{enumerate}
Data generation via world models emphasizes data quality and diversity, while human data integration leverages real-world context. They still face the challenges of reality gap, computational costs, and alignment issues. 

\subsection{Continual Learning}
Embodied intelligent systems should be able to autonomously update knowledge and optimize strategies through ongoing interactions with open dynamic environments, meanwhile maintaining previously acquired capabilities across evolving tasks and conditions. This ability can be achieved by continual learning\cite{mehta2023empirical}. Without continual learning, an intelligent agent needs to be retrained for every new scenario or slight environmental change, severely limiting its real-world utility. However, significant obstacles exist for continual learning. 
\begin{enumerate}[leftmargin=*]
\item Catastrophic forgetting\cite{parisi2019continual} remains a core issue, where learning new tasks degrades prior knowledge. For instance, a robot perfectly trained for navigation on flat surfaces may lose its proficiency when subsequently tasked with rugged terrain, limiting the robot’s accumulation of experiences. 
\item Efficient autonomous exploration is difficult, as current methods still struggle to balance exploring new experiences and exploiting existing knowledge effectively, especially in high-dimensional state spaces or scenarios with sparse rewards. 
\item Inherent unpredictability of the real world, including sensor degradation or mechanical wear, further complicates continual learning, necessitating robust self-diagnostic and self-repair capabilities of agents. 
\end{enumerate}
To address these challenges, researchers are exploring various approaches. \textbf{Experience replay}\cite{bagus2021investigation} can mitigate catastrophic forgetting by periodically revisiting historical data. \textbf{Regularization techniques}\cite{kirkpatrick2017overcoming} preserve prior task knowledge by constraining weight updates during new task learning. \textbf{Data mixing strategies}\cite{kumar2022fine} integrate varying proportions of prior data distributions with new data to reduce feature distortion. Frameworks such as CycleResearcher\cite{CycleResearcher} facilitate embodied learning in complex processes by \textbf{optimizing policy and reward models}. Future advances may include \textbf{enhance self-supervised learning} to drive active exploration via intrinsic motivation, and \textbf{incorporate multi-agent collaboration mechanisms} to accelerate individual learning through collective interactions.

\subsection{Computation and Deployment Efficiency}
The increasingly sophisticated models of embodied intelligence demand substantial computational resources for training and deployment. For instance, DiffusionVLA\cite{Sora} requires hundreds of high-end GPUs and weeks of training on million-scale trajectory datasets, with computations reaching petaflops (PFLOPs). Its iterative sampling during inference results in a latency of several seconds, which is an impediment for real-time control applications in robotics. The Transformer-based VLA RT-2\cite{RT-2} maintains a complex architecture that demands approximately 20GB of video memory. Although RT-2 has reduced training costs through pretraining, this high memory requirement complicates the deployment on resource-constrained edge devices, such as actual robots. Cloud-based deployment as an alternative is often impractical due to the concerns regarding data privacy, security, and real-time operational constraints inherent in physical robot interactions. To mitigate these challenges, several strategies are being explored. 
\begin{enumerate}[leftmargin=*]
\item Parameter-Efficient Fine-Tuning (PEFT) methods, such as LoRA\cite{LoRA}, significantly reduce fine-tuning costs. By updating only low-rank matrices, they can achieve the training costs approximately 1/10 of full fine-tuning. However, this efficiency may compromise the performance on highly complex tasks. 
\item Model compression techniques, including knowledge distillation ad quantization, can be leveraged to deploy large models on limited hardware. TinyVLA employs knowledge distillation to compress the large model to approximately 10 million parameters. Combined with fast sampling algorithms and 4-bit quantization, TinyVLA\cite{RT-2} achieves an impressive 30ms inference latency and a 2GB memory footprint, making it viable for edge devices. 
\item Hardware acceleration offers another solution for efficiency, as MiniGPT-4\cite{MiniGPT-4} does. Although providing immediate performance gains, hardware-specific optimizations often lack generalization across different platforms. 
\end{enumerate}
Optimizing large-scale models through compression techniques and designing inherently lightweight architectures are the most feasible approaches to achieve effective and widespread deployment of embodied AI on edge devices. 

\subsection{Sim-to-Real Gap}
Embodied AI requires a huge amount of data to train the agents. However, it is prohibitively expensive or impractical to collect such real-world data for diverse embodiments. Simulators alleviate this problem by enabling the agents to train on large and diverse simulated datasets\cite{rudin2022learning}, which proves to be a cost-effective and scalable solution. After training in simulators, the agents are deployed in real-world settings through sim-to-real transfer. 

However, sim-to-real transfer suffers from the “sim-to-real gap”\cite{wagenmaker2024overcoming}, due to fundamental discrepancies between simulated and real-world environments. These discrepancies manifest in various forms, such as inaccurate physical dynamics\cite{chen2023interact} and differences in visual rendering\cite{bertsch2023unlimiformer}. For example, friction, collision, and fluid behaviors are difficult to model accurately; lighting, camera exposure, and material properties are difficult to simulate. An agent trained in a simulated environment often fails unexpectedly when confronted with the subtle imperfections and complexities of the real world, as simulation cannot fully replicate reality. Therefore, well-trained policies may fail in out-of-distribution scenarios in the real world. Furthermore, modeling real-world environments accurately is inherently challenging\cite{dambekodi2020playing}. Slight differences between the simulated and the real world tend to accumulate in long-term decision-making, resulting in policies that are not robust or adaptable to environmental changes. 

Advanced simulators, such as the differentiable and highly realistic Genesis\cite{newbury2024review}, are working actively to narrow this gap through precise physics modeling and photorealistic rendering, thereby increasing the transferability of agents from simulators to the real world. Nonetheless, bridging the sim-to-real gap remains a major challenge for robust embodied AI.

\section{Conclusions}
Advent of large models endows embodied agents with great intelligent capabilities. In this article, we conduct a comprehensive survey on techniques and recent advances in embodied AI, focusing on autonomous decision-making and embodied learning empowered by large models. We begin the survey by introducing the preliminary knowledge of embodied AI and various major large models, reviewing their recent developments and applications on embodied intelligence. We then elaborate on the decision-making methodologies of embodied AI, detailing both hierarchical and end-to-end paradigms, their underlying mechanisms, and the latest advances. Afterwards, we review the embodied learning mechanisms, focusing on imitation learning and reinforcement learning, particularly how the large models empower them. Subsequently, we introduce the world models, presenting their design methods and their vital roles in decision-making and embodied learning. Finally, we discuss the open challenges in embodied intelligence, including embodied data scarcity, continual learning, computation and deployment efficiency, and sim-to-real gap, along with the potential solutions. Through this systematic investigation, we provide researchers and engineers with an in-depth summarization and analysis of status-quo and open challenges in the field of embodied AI, meanwhile figuring out the potential directions to the path towards artificial general intelligence.

\bibliographystyle{ACM-Reference-Format}
\bibliography{EmbodiedAI}


\begin{thebibliography}{234}


\ifx \showCODEN    \undefined \def \showCODEN     #1{\unskip}     \fi
\ifx \showDOI      \undefined \def \showDOI       #1{#1}\fi
\ifx \showISBNx    \undefined \def \showISBNx     #1{\unskip}     \fi
\ifx \showISBNxiii \undefined \def \showISBNxiii  #1{\unskip}     \fi
\ifx \showISSN     \undefined \def \showISSN      #1{\unskip}     \fi
\ifx \showLCCN     \undefined \def \showLCCN      #1{\unskip}     \fi
\ifx \shownote     \undefined \def \shownote      #1{#1}          \fi
\ifx \showarticletitle \undefined \def \showarticletitle #1{#1}   \fi
\ifx \showURL      \undefined \def \showURL       {\relax}        \fi
\providecommand\bibfield[2]{#2}
\providecommand\bibinfo[2]{#2}
\providecommand\natexlab[1]{#1}
\providecommand\showeprint[2][]{arXiv:#2}

\bibitem[Achiam et~al\mbox{.}(2023)]%
        {achiam2023gpt}
\bibfield{author}{\bibinfo{person}{Josh Achiam}, \bibinfo{person}{Steven Adler}, \bibinfo{person}{Sandhini Agarwal}, \bibinfo{person}{Lama Ahmad}, \bibinfo{person}{Ilge Akkaya}, \bibinfo{person}{Florencia~Leoni Aleman}, \bibinfo{person}{Diogo Almeida}, \bibinfo{person}{Janko Altenschmidt}, \bibinfo{person}{Sam Altman}, \bibinfo{person}{Shyamal Anadkat}, {et~al\mbox{.}}} \bibinfo{year}{2023}\natexlab{}.
\newblock \showarticletitle{Gpt-4 technical report}.
\newblock \bibinfo{journal}{\emph{arXiv preprint arXiv:2303.08774}} (\bibinfo{year}{2023}).
\newblock


\bibitem[Afram and Janabi-Sharifi(2014)]%
        {MPC}
\bibfield{author}{\bibinfo{person}{Abdul Afram} {and} \bibinfo{person}{Farrokh Janabi-Sharifi}.} \bibinfo{year}{2014}\natexlab{}.
\newblock \showarticletitle{Theory and applications of HVAC control systems--A review of model predictive control (MPC)}.
\newblock \bibinfo{journal}{\emph{Building and environment}}  \bibinfo{volume}{72} (\bibinfo{year}{2014}), \bibinfo{pages}{343--355}.
\newblock


\bibitem[Agha et~al\mbox{.}(2021)]%
        {NeBula}
\bibfield{author}{\bibinfo{person}{Ali Agha}, \bibinfo{person}{Kyohei Otsu}, \bibinfo{person}{Benjamin Morrell}, \bibinfo{person}{David~D Fan}, \bibinfo{person}{Rohan Thakker}, \bibinfo{person}{Angel Santamaria-Navarro}, \bibinfo{person}{Sung-Kyun Kim}, \bibinfo{person}{Amanda Bouman}, \bibinfo{person}{Xianmei Lei}, \bibinfo{person}{Jeffrey Edlund}, {et~al\mbox{.}}} \bibinfo{year}{2021}\natexlab{}.
\newblock \showarticletitle{Nebula: Quest for robotic autonomy in challenging environments; team costar at the darpa subterranean challenge}.
\newblock \bibinfo{journal}{\emph{arXiv preprint arXiv:2103.11470}} (\bibinfo{year}{2021}).
\newblock


\bibitem[Ahn et~al\mbox{.}(2022)]%
        {SayCAN}
\bibfield{author}{\bibinfo{person}{Michael Ahn}, \bibinfo{person}{Anthony Brohan}, \bibinfo{person}{Noah Brown}, \bibinfo{person}{Yevgen Chebotar}, \bibinfo{person}{Omar Cortes}, \bibinfo{person}{Byron David}, \bibinfo{person}{Chelsea Finn}, \bibinfo{person}{Chuyuan Fu}, \bibinfo{person}{Keerthana Gopalakrishnan}, \bibinfo{person}{Karol Hausman}, {et~al\mbox{.}}} \bibinfo{year}{2022}\natexlab{}.
\newblock \showarticletitle{Do as i can, not as i say: Grounding language in robotic affordances}.
\newblock \bibinfo{journal}{\emph{arXiv preprint arXiv:2204.01691}} (\bibinfo{year}{2022}).
\newblock


\bibitem[Ahn et~al\mbox{.}(2024)]%
        {autort}
\bibfield{author}{\bibinfo{person}{Michael Ahn}, \bibinfo{person}{Debidatta Dwibedi}, \bibinfo{person}{Chelsea Finn}, \bibinfo{person}{Montse~Gonzalez Arenas}, \bibinfo{person}{Keerthana Gopalakrishnan}, \bibinfo{person}{Karol Hausman}, \bibinfo{person}{Brian Ichter}, \bibinfo{person}{Alex Irpan}, \bibinfo{person}{Nikhil Joshi}, \bibinfo{person}{Ryan Julian}, {et~al\mbox{.}}} \bibinfo{year}{2024}\natexlab{}.
\newblock \showarticletitle{Autort: Embodied foundation models for large scale orchestration of robotic agents}.
\newblock \bibinfo{journal}{\emph{arXiv preprint arXiv:2401.12963}} (\bibinfo{year}{2024}).
\newblock


\bibitem[Alayrac et~al\mbox{.}(2022)]%
        {Flamingo}
\bibfield{author}{\bibinfo{person}{Jean-Baptiste Alayrac}, \bibinfo{person}{Jeff Donahue}, \bibinfo{person}{Pauline Luc}, \bibinfo{person}{Antoine Miech}, \bibinfo{person}{Iain Barr}, \bibinfo{person}{Yana Hasson}, \bibinfo{person}{Karel Lenc}, \bibinfo{person}{Arthur Mensch}, \bibinfo{person}{Katherine Millican}, \bibinfo{person}{Malcolm Reynolds}, {et~al\mbox{.}}} \bibinfo{year}{2022}\natexlab{}.
\newblock \showarticletitle{Flamingo: a visual language model for few-shot learning}.
\newblock \bibinfo{journal}{\emph{Advances in neural information processing systems}}  \bibinfo{volume}{35} (\bibinfo{year}{2022}), \bibinfo{pages}{23716--23736}.
\newblock


\bibitem[An et~al\mbox{.}(2025)]%
        {9}
\bibfield{author}{\bibinfo{person}{Shan An}, \bibinfo{person}{Ziyu Meng}, \bibinfo{person}{Chao Tang}, \bibinfo{person}{Yuning Zhou}, \bibinfo{person}{Tengyu Liu}, \bibinfo{person}{Fangqiang Ding}, \bibinfo{person}{Shufang Zhang}, \bibinfo{person}{Yao Mu}, \bibinfo{person}{Ran Song}, \bibinfo{person}{Wei Zhang}, {et~al\mbox{.}}} \bibinfo{year}{2025}\natexlab{}.
\newblock \showarticletitle{Dexterous manipulation through imitation learning: A survey}.
\newblock \bibinfo{journal}{\emph{arXiv preprint arXiv:2504.03515}} (\bibinfo{year}{2025}).
\newblock


\bibitem[Anil et~al\mbox{.}(2023)]%
        {anil2023palm}
\bibfield{author}{\bibinfo{person}{Rohan Anil}, \bibinfo{person}{Andrew~M Dai}, \bibinfo{person}{Orhan Firat}, \bibinfo{person}{Melvin Johnson}, \bibinfo{person}{Dmitry Lepikhin}, \bibinfo{person}{Alexandre Passos}, \bibinfo{person}{Siamak Shakeri}, \bibinfo{person}{Emanuel Taropa}, \bibinfo{person}{Paige Bailey}, \bibinfo{person}{Zhifeng Chen}, {et~al\mbox{.}}} \bibinfo{year}{2023}\natexlab{}.
\newblock \showarticletitle{Palm 2 technical report}.
\newblock \bibinfo{journal}{\emph{arXiv preprint arXiv:2305.10403}} (\bibinfo{year}{2023}).
\newblock


\bibitem[Arora and Kambhampati(2023)]%
        {LLV}
\bibfield{author}{\bibinfo{person}{Daman Arora} {and} \bibinfo{person}{Subbarao Kambhampati}.} \bibinfo{year}{2023}\natexlab{}.
\newblock \showarticletitle{Learning and leveraging verifiers to improve planning capabilities of pre-trained language models}.
\newblock \bibinfo{journal}{\emph{arXiv preprint arXiv:2305.17077}} (\bibinfo{year}{2023}).
\newblock


\bibitem[Bagus and Gepperth(2021)]%
        {bagus2021investigation}
\bibfield{author}{\bibinfo{person}{Benedikt Bagus} {and} \bibinfo{person}{Alexander Gepperth}.} \bibinfo{year}{2021}\natexlab{}.
\newblock \showarticletitle{An investigation of replay-based approaches for continual learning}. In \bibinfo{booktitle}{\emph{2021 International Joint Conference on Neural Networks (IJCNN)}}. IEEE, \bibinfo{pages}{1--9}.
\newblock


\bibitem[Barto(2021)]%
        {Reinforcementlearning}
\bibfield{author}{\bibinfo{person}{Andrew~G Barto}.} \bibinfo{year}{2021}\natexlab{}.
\newblock \showarticletitle{Reinforcement learning: An introduction. by richard’s sutton}.
\newblock \bibinfo{journal}{\emph{SIAM Rev}} \bibinfo{volume}{6}, \bibinfo{number}{2} (\bibinfo{year}{2021}), \bibinfo{pages}{423}.
\newblock


\bibitem[Belkhale et~al\mbox{.}(2023)]%
        {belkhale2023hydra}
\bibfield{author}{\bibinfo{person}{Suneel Belkhale}, \bibinfo{person}{Yuchen Cui}, {and} \bibinfo{person}{Dorsa Sadigh}.} \bibinfo{year}{2023}\natexlab{}.
\newblock \showarticletitle{Hydra: Hybrid robot actions for imitation learning}. In \bibinfo{booktitle}{\emph{Conference on Robot Learning}}. PMLR, \bibinfo{pages}{2113--2133}.
\newblock


\bibitem[Bellegarda et~al\mbox{.}(2022)]%
        {bellegarda2022robust}
\bibfield{author}{\bibinfo{person}{Guillaume Bellegarda}, \bibinfo{person}{Yiyu Chen}, \bibinfo{person}{Zhuochen Liu}, {and} \bibinfo{person}{Quan Nguyen}.} \bibinfo{year}{2022}\natexlab{}.
\newblock \showarticletitle{Robust high-speed running for quadruped robots via deep reinforcement learning}. In \bibinfo{booktitle}{\emph{2022 IEEE/RSJ International Conference on Intelligent Robots and Systems (IROS)}}. IEEE, \bibinfo{pages}{10364--10370}.
\newblock


\bibitem[Bertsch et~al\mbox{.}(2023)]%
        {bertsch2023unlimiformer}
\bibfield{author}{\bibinfo{person}{Amanda Bertsch}, \bibinfo{person}{Uri Alon}, \bibinfo{person}{Graham Neubig}, {and} \bibinfo{person}{Matthew Gormley}.} \bibinfo{year}{2023}\natexlab{}.
\newblock \showarticletitle{Unlimiformer: Long-range transformers with unlimited length input}.
\newblock \bibinfo{journal}{\emph{Advances in Neural Information Processing Systems}}  \bibinfo{volume}{36} (\bibinfo{year}{2023}), \bibinfo{pages}{35522--35543}.
\newblock


\bibitem[Besta et~al\mbox{.}(2024)]%
        {GoT}
\bibfield{author}{\bibinfo{person}{Maciej Besta}, \bibinfo{person}{Nils Blach}, \bibinfo{person}{Ales Kubicek}, \bibinfo{person}{Robert Gerstenberger}, \bibinfo{person}{Michal Podstawski}, \bibinfo{person}{Lukas Gianinazzi}, \bibinfo{person}{Joanna Gajda}, \bibinfo{person}{Tomasz Lehmann}, \bibinfo{person}{Hubert Niewiadomski}, \bibinfo{person}{Piotr Nyczyk}, {et~al\mbox{.}}} \bibinfo{year}{2024}\natexlab{}.
\newblock \showarticletitle{Graph of thoughts: Solving elaborate problems with large language models}. In \bibinfo{booktitle}{\emph{Proceedings of the AAAI conference on artificial intelligence}}, Vol.~\bibinfo{volume}{38}. \bibinfo{pages}{17682--17690}.
\newblock


\bibitem[Betker et~al\mbox{.}(2023)]%
        {DALL·E3}
\bibfield{author}{\bibinfo{person}{James Betker}, \bibinfo{person}{Gabriel Goh}, \bibinfo{person}{Li Jing}, \bibinfo{person}{Tim Brooks}, \bibinfo{person}{Jianfeng Wang}, \bibinfo{person}{Linjie Li}, \bibinfo{person}{Long Ouyang}, \bibinfo{person}{Juntang Zhuang}, \bibinfo{person}{Joyce Lee}, \bibinfo{person}{Yufei Guo}, {et~al\mbox{.}}} \bibinfo{year}{2023}\natexlab{}.
\newblock \showarticletitle{Improving image generation with better captions}.
\newblock \bibinfo{journal}{\emph{Computer Science. https://cdn. openai. com/papers/dall-e-3. pdf}} \bibinfo{volume}{2}, \bibinfo{number}{3} (\bibinfo{year}{2023}), \bibinfo{pages}{8}.
\newblock


\bibitem[Bharadhwaj et~al\mbox{.}(2024)]%
        {RoboAgent}
\bibfield{author}{\bibinfo{person}{Homanga Bharadhwaj}, \bibinfo{person}{Jay Vakil}, \bibinfo{person}{Mohit Sharma}, \bibinfo{person}{Abhinav Gupta}, \bibinfo{person}{Shubham Tulsiani}, {and} \bibinfo{person}{Vikash Kumar}.} \bibinfo{year}{2024}\natexlab{}.
\newblock \showarticletitle{Roboagent: Generalization and efficiency in robot manipulation via semantic augmentations and action chunking}. In \bibinfo{booktitle}{\emph{2024 IEEE International Conference on Robotics and Automation (ICRA)}}. IEEE, \bibinfo{pages}{4788--4795}.
\newblock


\bibitem[Black et~al\mbox{.}(2024)]%
        {PI0}
\bibfield{author}{\bibinfo{person}{Kevin Black}, \bibinfo{person}{Noah Brown}, \bibinfo{person}{Danny Driess}, \bibinfo{person}{Adnan Esmail}, \bibinfo{person}{Michael Equi}, \bibinfo{person}{Chelsea Finn}, \bibinfo{person}{Niccolo Fusai}, \bibinfo{person}{Lachy Groom}, \bibinfo{person}{Karol Hausman}, \bibinfo{person}{Brian Ichter}, {et~al\mbox{.}}} \bibinfo{year}{2024}\natexlab{}.
\newblock \showarticletitle{$\pi_0$: A Vision-Language-Action Flow Model for General Robot Control}.
\newblock \bibinfo{journal}{\emph{arXiv preprint arXiv:2410.24164}} (\bibinfo{year}{2024}).
\newblock


\bibitem[Bousmalis et~al\mbox{.}(2023)]%
        {RoboCat}
\bibfield{author}{\bibinfo{person}{Konstantinos Bousmalis}, \bibinfo{person}{Giulia Vezzani}, \bibinfo{person}{Dushyant Rao}, \bibinfo{person}{Coline Devin}, \bibinfo{person}{Alex~X Lee}, \bibinfo{person}{Maria Bauz{\'a}}, \bibinfo{person}{Todor Davchev}, \bibinfo{person}{Yuxiang Zhou}, \bibinfo{person}{Agrim Gupta}, \bibinfo{person}{Akhil Raju}, {et~al\mbox{.}}} \bibinfo{year}{2023}\natexlab{}.
\newblock \showarticletitle{Robocat: A self-improving generalist agent for robotic manipulation}.
\newblock \bibinfo{journal}{\emph{arXiv preprint arXiv:2306.11706}} (\bibinfo{year}{2023}).
\newblock


\bibitem[Brohan et~al\mbox{.}(2022)]%
        {RT-1}
\bibfield{author}{\bibinfo{person}{Anthony Brohan}, \bibinfo{person}{Noah Brown}, \bibinfo{person}{Justice Carbajal}, \bibinfo{person}{Yevgen Chebotar}, \bibinfo{person}{Joseph Dabis}, \bibinfo{person}{Chelsea Finn}, \bibinfo{person}{Keerthana Gopalakrishnan}, \bibinfo{person}{Karol Hausman}, \bibinfo{person}{Alex Herzog}, \bibinfo{person}{Jasmine Hsu}, {et~al\mbox{.}}} \bibinfo{year}{2022}\natexlab{}.
\newblock \showarticletitle{Rt-1: Robotics transformer for real-world control at scale}.
\newblock \bibinfo{journal}{\emph{arXiv preprint arXiv:2212.06817}} (\bibinfo{year}{2022}).
\newblock


\bibitem[Brooks(2003)]%
        {brooks2003robust}
\bibfield{author}{\bibinfo{person}{Rodney Brooks}.} \bibinfo{year}{2003}\natexlab{}.
\newblock \showarticletitle{A robust layered control system for a mobile robot}.
\newblock \bibinfo{journal}{\emph{IEEE journal on robotics and automation}} \bibinfo{volume}{2}, \bibinfo{number}{1} (\bibinfo{year}{2003}), \bibinfo{pages}{14--23}.
\newblock


\bibitem[Brooks et~al\mbox{.}(2024)]%
        {Sora}
\bibfield{author}{\bibinfo{person}{Tim Brooks}, \bibinfo{person}{Bill Peebles}, \bibinfo{person}{Connor Holmes}, \bibinfo{person}{Will DePue}, \bibinfo{person}{Yufei Guo}, \bibinfo{person}{Li Jing}, \bibinfo{person}{David Schnurr}, \bibinfo{person}{Joe Taylor}, \bibinfo{person}{Troy Luhman}, \bibinfo{person}{Eric Luhman}, {et~al\mbox{.}}} \bibinfo{year}{2024}\natexlab{}.
\newblock \showarticletitle{Video generation models as world simulators}.
\newblock \bibinfo{journal}{\emph{OpenAI Blog}} \bibinfo{volume}{1}, \bibinfo{number}{8} (\bibinfo{year}{2024}), \bibinfo{pages}{1}.
\newblock


\bibitem[Brown et~al\mbox{.}(2020)]%
        {ICL}
\bibfield{author}{\bibinfo{person}{Tom Brown}, \bibinfo{person}{Benjamin Mann}, \bibinfo{person}{Nick Ryder}, \bibinfo{person}{Melanie Subbiah}, \bibinfo{person}{Jared~D Kaplan}, \bibinfo{person}{Prafulla Dhariwal}, \bibinfo{person}{Arvind Neelakantan}, \bibinfo{person}{Pranav Shyam}, \bibinfo{person}{Girish Sastry}, \bibinfo{person}{Amanda Askell}, {et~al\mbox{.}}} \bibinfo{year}{2020}\natexlab{}.
\newblock \showarticletitle{Language models are few-shot learners}.
\newblock \bibinfo{journal}{\emph{Advances in neural information processing systems}}  \bibinfo{volume}{33} (\bibinfo{year}{2020}), \bibinfo{pages}{1877--1901}.
\newblock


\bibitem[Bruce et~al\mbox{.}(2024)]%
        {Genie}
\bibfield{author}{\bibinfo{person}{Jake Bruce}, \bibinfo{person}{Michael~D Dennis}, \bibinfo{person}{Ashley Edwards}, \bibinfo{person}{Jack Parker-Holder}, \bibinfo{person}{Yuge Shi}, \bibinfo{person}{Edward Hughes}, \bibinfo{person}{Matthew Lai}, \bibinfo{person}{Aditi Mavalankar}, \bibinfo{person}{Richie Steigerwald}, \bibinfo{person}{Chris Apps}, {et~al\mbox{.}}} \bibinfo{year}{2024}\natexlab{}.
\newblock \showarticletitle{Genie: Generative interactive environments}. In \bibinfo{booktitle}{\emph{Forty-first International Conference on Machine Learning}}.
\newblock


\bibitem[Budzianowski et~al\mbox{.}(2025)]%
        {edgevla}
\bibfield{author}{\bibinfo{person}{Pawe{\l} Budzianowski}, \bibinfo{person}{Wesley Maa}, \bibinfo{person}{Matthew Freed}, \bibinfo{person}{Jingxiang Mo}, \bibinfo{person}{Winston Hsiao}, \bibinfo{person}{Aaron Xie}, \bibinfo{person}{Tomasz M{\l}oduchowski}, \bibinfo{person}{Viraj Tipnis}, {and} \bibinfo{person}{Benjamin Bolte}.} \bibinfo{year}{2025}\natexlab{}.
\newblock \showarticletitle{EdgeVLA: Efficient Vision-Language-Action Models}.
\newblock \bibinfo{journal}{\emph{arXiv preprint arXiv:2507.14049}} (\bibinfo{year}{2025}).
\newblock


\bibitem[Cao et~al\mbox{.}(2024)]%
        {8}
\bibfield{author}{\bibinfo{person}{Yuji Cao}, \bibinfo{person}{Huan Zhao}, \bibinfo{person}{Yuheng Cheng}, \bibinfo{person}{Ting Shu}, \bibinfo{person}{Yue Chen}, \bibinfo{person}{Guolong Liu}, \bibinfo{person}{Gaoqi Liang}, \bibinfo{person}{Junhua Zhao}, \bibinfo{person}{Jinyue Yan}, {and} \bibinfo{person}{Yun Li}.} \bibinfo{year}{2024}\natexlab{}.
\newblock \showarticletitle{Survey on large language model-enhanced reinforcement learning: Concept, taxonomy, and methods}.
\newblock \bibinfo{journal}{\emph{IEEE Transactions on Neural Networks and Learning Systems}} (\bibinfo{year}{2024}).
\newblock


\bibitem[Caron et~al\mbox{.}(2021)]%
        {DINO}
\bibfield{author}{\bibinfo{person}{Mathilde Caron}, \bibinfo{person}{Hugo Touvron}, \bibinfo{person}{Ishan Misra}, \bibinfo{person}{Herv{\'e} J{\'e}gou}, \bibinfo{person}{Julien Mairal}, \bibinfo{person}{Piotr Bojanowski}, {and} \bibinfo{person}{Armand Joulin}.} \bibinfo{year}{2021}\natexlab{}.
\newblock \showarticletitle{Emerging properties in self-supervised vision transformers}. In \bibinfo{booktitle}{\emph{Proceedings of the IEEE/CVF international conference on computer vision}}. \bibinfo{pages}{9650--9660}.
\newblock


\bibitem[Carta et~al\mbox{.}(2023)]%
        {GLAM}
\bibfield{author}{\bibinfo{person}{Thomas Carta}, \bibinfo{person}{Cl{\'e}ment Romac}, \bibinfo{person}{Thomas Wolf}, \bibinfo{person}{Sylvain Lamprier}, \bibinfo{person}{Olivier Sigaud}, {and} \bibinfo{person}{Pierre-Yves Oudeyer}.} \bibinfo{year}{2023}\natexlab{}.
\newblock \showarticletitle{Grounding large language models in interactive environments with online reinforcement learning}. In \bibinfo{booktitle}{\emph{International Conference on Machine Learning}}. PMLR, \bibinfo{pages}{3676--3713}.
\newblock


\bibitem[Chang et~al\mbox{.}(2024)]%
        {chang2024survey}
\bibfield{author}{\bibinfo{person}{Yupeng Chang}, \bibinfo{person}{Xu Wang}, \bibinfo{person}{Jindong Wang}, \bibinfo{person}{Yuan Wu}, \bibinfo{person}{Linyi Yang}, \bibinfo{person}{Kaijie Zhu}, \bibinfo{person}{Hao Chen}, \bibinfo{person}{Xiaoyuan Yi}, \bibinfo{person}{Cunxiang Wang}, \bibinfo{person}{Yidong Wang}, {et~al\mbox{.}}} \bibinfo{year}{2024}\natexlab{}.
\newblock \showarticletitle{A survey on evaluation of large language models}.
\newblock \bibinfo{journal}{\emph{ACM transactions on intelligent systems and technology}} \bibinfo{volume}{15}, \bibinfo{number}{3} (\bibinfo{year}{2024}), \bibinfo{pages}{1--45}.
\newblock


\bibitem[Chebotar et~al\mbox{.}(2023)]%
        {Q-Transformer}
\bibfield{author}{\bibinfo{person}{Yevgen Chebotar}, \bibinfo{person}{Quan Vuong}, \bibinfo{person}{Karol Hausman}, \bibinfo{person}{Fei Xia}, \bibinfo{person}{Yao Lu}, \bibinfo{person}{Alex Irpan}, \bibinfo{person}{Aviral Kumar}, \bibinfo{person}{Tianhe Yu}, \bibinfo{person}{Alexander Herzog}, \bibinfo{person}{Karl Pertsch}, {et~al\mbox{.}}} \bibinfo{year}{2023}\natexlab{}.
\newblock \showarticletitle{Q-transformer: Scalable offline reinforcement learning via autoregressive q-functions}. In \bibinfo{booktitle}{\emph{Conference on Robot Learning}}. PMLR, \bibinfo{pages}{3909--3928}.
\newblock


\bibitem[Chen et~al\mbox{.}(2021a)]%
        {DecisionTransformer}
\bibfield{author}{\bibinfo{person}{Lili Chen}, \bibinfo{person}{Kevin Lu}, \bibinfo{person}{Aravind Rajeswaran}, \bibinfo{person}{Kimin Lee}, \bibinfo{person}{Aditya Grover}, \bibinfo{person}{Misha Laskin}, \bibinfo{person}{Pieter Abbeel}, \bibinfo{person}{Aravind Srinivas}, {and} \bibinfo{person}{Igor Mordatch}.} \bibinfo{year}{2021}\natexlab{a}.
\newblock \showarticletitle{Decision transformer: Reinforcement learning via sequence modeling}.
\newblock \bibinfo{journal}{\emph{Advances in neural information processing systems}}  \bibinfo{volume}{34} (\bibinfo{year}{2021}), \bibinfo{pages}{15084--15097}.
\newblock


\bibitem[Chen et~al\mbox{.}(2021b)]%
        {Codex}
\bibfield{author}{\bibinfo{person}{Mark Chen}, \bibinfo{person}{Jerry Tworek}, \bibinfo{person}{Heewoo Jun}, \bibinfo{person}{Qiming Yuan}, \bibinfo{person}{Henrique Ponde De~Oliveira Pinto}, \bibinfo{person}{Jared Kaplan}, \bibinfo{person}{Harri Edwards}, \bibinfo{person}{Yuri Burda}, \bibinfo{person}{Nicholas Joseph}, \bibinfo{person}{Greg Brockman}, {et~al\mbox{.}}} \bibinfo{year}{2021}\natexlab{b}.
\newblock \showarticletitle{Evaluating large language models trained on code}.
\newblock \bibinfo{journal}{\emph{arXiv preprint arXiv:2107.03374}} (\bibinfo{year}{2021}).
\newblock


\bibitem[Chen and Chang(2023)]%
        {chen2023interact}
\bibfield{author}{\bibinfo{person}{Po-Lin Chen} {and} \bibinfo{person}{Cheng-Shang Chang}.} \bibinfo{year}{2023}\natexlab{}.
\newblock \showarticletitle{Interact: Exploring the potentials of chatgpt as a cooperative agent}.
\newblock \bibinfo{journal}{\emph{arXiv preprint arXiv:2308.01552}} (\bibinfo{year}{2023}).
\newblock


\bibitem[Chen et~al\mbox{.}(2023)]%
        {DABC}
\bibfield{author}{\bibinfo{person}{Shang-Fu Chen}, \bibinfo{person}{Hsiang-Chun Wang}, \bibinfo{person}{Ming-Hao Hsu}, \bibinfo{person}{Chun-Mao Lai}, {and} \bibinfo{person}{Shao-Hua Sun}.} \bibinfo{year}{2023}\natexlab{}.
\newblock \showarticletitle{Diffusion model-augmented behavioral cloning}.
\newblock \bibinfo{journal}{\emph{arXiv preprint arXiv:2302.13335}} (\bibinfo{year}{2023}).
\newblock


\bibitem[Chen et~al\mbox{.}(2020)]%
        {chen2020simple}
\bibfield{author}{\bibinfo{person}{Ting Chen}, \bibinfo{person}{Simon Kornblith}, \bibinfo{person}{Mohammad Norouzi}, {and} \bibinfo{person}{Geoffrey Hinton}.} \bibinfo{year}{2020}\natexlab{}.
\newblock \showarticletitle{A simple framework for contrastive learning of visual representations}. In \bibinfo{booktitle}{\emph{International conference on machine learning}}. PmLR, \bibinfo{pages}{1597--1607}.
\newblock


\bibitem[Chi et~al\mbox{.}(2023)]%
        {DiffusionPolicy}
\bibfield{author}{\bibinfo{person}{Cheng Chi}, \bibinfo{person}{Zhenjia Xu}, \bibinfo{person}{Siyuan Feng}, \bibinfo{person}{Eric Cousineau}, \bibinfo{person}{Yilun Du}, \bibinfo{person}{Benjamin Burchfiel}, \bibinfo{person}{Russ Tedrake}, {and} \bibinfo{person}{Shuran Song}.} \bibinfo{year}{2023}\natexlab{}.
\newblock \showarticletitle{Diffusion policy: Visuomotor policy learning via action diffusion}.
\newblock \bibinfo{journal}{\emph{The International Journal of Robotics Research}} (\bibinfo{year}{2023}), \bibinfo{pages}{02783649241273668}.
\newblock


\bibitem[Chiang et~al\mbox{.}(2024)]%
        {mobilityVLA}
\bibfield{author}{\bibinfo{person}{Hao-Tien~Lewis Chiang}, \bibinfo{person}{Zhuo Xu}, \bibinfo{person}{Zipeng Fu}, \bibinfo{person}{Mithun~George Jacob}, \bibinfo{person}{Tingnan Zhang}, \bibinfo{person}{Tsang-Wei~Edward Lee}, \bibinfo{person}{Wenhao Yu}, \bibinfo{person}{Connor Schenck}, \bibinfo{person}{David Rendleman}, \bibinfo{person}{Dhruv Shah}, {et~al\mbox{.}}} \bibinfo{year}{2024}\natexlab{}.
\newblock \showarticletitle{Mobility vla: Multimodal instruction navigation with long-context vlms and topological graphs}.
\newblock \bibinfo{journal}{\emph{arXiv preprint arXiv:2407.07775}} (\bibinfo{year}{2024}).
\newblock


\bibitem[Chiang et~al\mbox{.}(2023)]%
        {Vicuna}
\bibfield{author}{\bibinfo{person}{WL Chiang} {et~al\mbox{.}}} \bibinfo{year}{2023}\natexlab{}.
\newblock \bibinfo{title}{Vicuna: An Open-Source Chatbot Impressing GPT-4 With 90\% ChatGPT Quality. Accessed: Apr. 14, 2023}.
\newblock
\newblock


\bibitem[Chowdhery et~al\mbox{.}(2023)]%
        {chowdhery2023palm}
\bibfield{author}{\bibinfo{person}{Aakanksha Chowdhery}, \bibinfo{person}{Sharan Narang}, \bibinfo{person}{Jacob Devlin}, \bibinfo{person}{Maarten Bosma}, \bibinfo{person}{Gaurav Mishra}, \bibinfo{person}{Adam Roberts}, \bibinfo{person}{Paul Barham}, \bibinfo{person}{Hyung~Won Chung}, \bibinfo{person}{Charles Sutton}, \bibinfo{person}{Sebastian Gehrmann}, {et~al\mbox{.}}} \bibinfo{year}{2023}\natexlab{}.
\newblock \showarticletitle{Palm: Scaling language modeling with pathways}.
\newblock \bibinfo{journal}{\emph{Journal of Machine Learning Research}} \bibinfo{volume}{24}, \bibinfo{number}{240} (\bibinfo{year}{2023}), \bibinfo{pages}{1--113}.
\newblock


\bibitem[Croitoru et~al\mbox{.}(2023)]%
        {diffusion}
\bibfield{author}{\bibinfo{person}{Florinel-Alin Croitoru}, \bibinfo{person}{Vlad Hondru}, \bibinfo{person}{Radu~Tudor Ionescu}, {and} \bibinfo{person}{Mubarak Shah}.} \bibinfo{year}{2023}\natexlab{}.
\newblock \showarticletitle{Diffusion models in vision: A survey}.
\newblock \bibinfo{journal}{\emph{IEEE transactions on pattern analysis and machine intelligence}} \bibinfo{volume}{45}, \bibinfo{number}{9} (\bibinfo{year}{2023}), \bibinfo{pages}{10850--10869}.
\newblock


\bibitem[Dambekodi et~al\mbox{.}(2020)]%
        {dambekodi2020playing}
\bibfield{author}{\bibinfo{person}{Sahith Dambekodi}, \bibinfo{person}{Spencer Frazier}, \bibinfo{person}{Prithviraj Ammanabrolu}, {and} \bibinfo{person}{Mark~O Riedl}.} \bibinfo{year}{2020}\natexlab{}.
\newblock \showarticletitle{Playing text-based games with common sense}.
\newblock \bibinfo{journal}{\emph{arXiv preprint arXiv:2012.02757}} (\bibinfo{year}{2020}).
\newblock


\bibitem[Devlin et~al\mbox{.}(2019)]%
        {Bert}
\bibfield{author}{\bibinfo{person}{Jacob Devlin}, \bibinfo{person}{Ming-Wei Chang}, \bibinfo{person}{Kenton Lee}, {and} \bibinfo{person}{Kristina Toutanova}.} \bibinfo{year}{2019}\natexlab{}.
\newblock \showarticletitle{Bert: Pre-training of deep bidirectional transformers for language understanding}. In \bibinfo{booktitle}{\emph{Proceedings of the 2019 conference of the North American chapter of the association for computational linguistics: human language technologies, volume 1 (long and short papers)}}. \bibinfo{pages}{4171--4186}.
\newblock


\bibitem[Ding et~al\mbox{.}(2024)]%
        {13}
\bibfield{author}{\bibinfo{person}{Jingtao Ding}, \bibinfo{person}{Yunke Zhang}, \bibinfo{person}{Yu Shang}, \bibinfo{person}{Yuheng Zhang}, \bibinfo{person}{Zefang Zong}, \bibinfo{person}{Jie Feng}, \bibinfo{person}{Yuan Yuan}, \bibinfo{person}{Hongyuan Su}, \bibinfo{person}{Nian Li}, \bibinfo{person}{Nicholas Sukiennik}, {et~al\mbox{.}}} \bibinfo{year}{2024}\natexlab{}.
\newblock \showarticletitle{Understanding world or predicting future? a comprehensive survey of world models}.
\newblock \bibinfo{journal}{\emph{Comput. Surveys}} (\bibinfo{year}{2024}).
\newblock


\bibitem[Doshi et~al\mbox{.}(2024)]%
        {CrossFormer}
\bibfield{author}{\bibinfo{person}{Ria Doshi}, \bibinfo{person}{Homer Walke}, \bibinfo{person}{Oier Mees}, \bibinfo{person}{Sudeep Dasari}, {and} \bibinfo{person}{Sergey Levine}.} \bibinfo{year}{2024}\natexlab{}.
\newblock \showarticletitle{Scaling cross-embodied learning: One policy for manipulation, navigation, locomotion and aviation}.
\newblock \bibinfo{journal}{\emph{arXiv preprint arXiv:2408.11812}} (\bibinfo{year}{2024}).
\newblock


\bibitem[Dosovitskiy et~al\mbox{.}(2020)]%
        {ViT}
\bibfield{author}{\bibinfo{person}{Alexey Dosovitskiy}, \bibinfo{person}{Lucas Beyer}, \bibinfo{person}{Alexander Kolesnikov}, \bibinfo{person}{Dirk Weissenborn}, \bibinfo{person}{Xiaohua Zhai}, \bibinfo{person}{Thomas Unterthiner}, \bibinfo{person}{Mostafa Dehghani}, \bibinfo{person}{Matthias Minderer}, \bibinfo{person}{Georg Heigold}, \bibinfo{person}{Sylvain Gelly}, {et~al\mbox{.}}} \bibinfo{year}{2020}\natexlab{}.
\newblock \showarticletitle{An image is worth 16x16 words: Transformers for image recognition at scale}.
\newblock \bibinfo{journal}{\emph{arXiv preprint arXiv:2010.11929}} (\bibinfo{year}{2020}).
\newblock


\bibitem[Driess et~al\mbox{.}(2023)]%
        {PaLM-E}
\bibfield{author}{\bibinfo{person}{Danny Driess}, \bibinfo{person}{Fei Xia}, \bibinfo{person}{Mehdi~SM Sajjadi}, \bibinfo{person}{Corey Lynch}, \bibinfo{person}{Aakanksha Chowdhery}, \bibinfo{person}{Ayzaan Wahid}, \bibinfo{person}{Jonathan Tompson}, \bibinfo{person}{Quan Vuong}, \bibinfo{person}{Tianhe Yu}, \bibinfo{person}{Wenlong Huang}, {et~al\mbox{.}}} \bibinfo{year}{2023}\natexlab{}.
\newblock \showarticletitle{Palm-e: An embodied multimodal language model}.
\newblock  (\bibinfo{year}{2023}).
\newblock


\bibitem[Du et~al\mbox{.}(2023)]%
        {UniPi}
\bibfield{author}{\bibinfo{person}{Yilun Du}, \bibinfo{person}{Sherry Yang}, \bibinfo{person}{Bo Dai}, \bibinfo{person}{Hanjun Dai}, \bibinfo{person}{Ofir Nachum}, \bibinfo{person}{Josh Tenenbaum}, \bibinfo{person}{Dale Schuurmans}, {and} \bibinfo{person}{Pieter Abbeel}.} \bibinfo{year}{2023}\natexlab{}.
\newblock \showarticletitle{Learning universal policies via text-guided video generation}.
\newblock \bibinfo{journal}{\emph{Advances in neural information processing systems}}  \bibinfo{volume}{36} (\bibinfo{year}{2023}), \bibinfo{pages}{9156--9172}.
\newblock


\bibitem[Duan et~al\mbox{.}(2022)]%
        {duan2022survey}
\bibfield{author}{\bibinfo{person}{Jiafei Duan}, \bibinfo{person}{Samson Yu}, \bibinfo{person}{Hui~Li Tan}, \bibinfo{person}{Hongyuan Zhu}, {and} \bibinfo{person}{Cheston Tan}.} \bibinfo{year}{2022}\natexlab{}.
\newblock \showarticletitle{A survey of embodied ai: From simulators to research tasks}.
\newblock \bibinfo{journal}{\emph{IEEE Transactions on Emerging Topics in Computational Intelligence}} \bibinfo{volume}{6}, \bibinfo{number}{2} (\bibinfo{year}{2022}), \bibinfo{pages}{230--244}.
\newblock


\bibitem[Eschmann(2021)]%
        {eschmann2021reward}
\bibfield{author}{\bibinfo{person}{Jonas Eschmann}.} \bibinfo{year}{2021}\natexlab{}.
\newblock \showarticletitle{Reward function design in reinforcement learning}.
\newblock \bibinfo{journal}{\emph{Reinforcement learning algorithms: Analysis and Applications}} (\bibinfo{year}{2021}), \bibinfo{pages}{25--33}.
\newblock


\bibitem[Esser et~al\mbox{.}(2021)]%
        {VQ-GAN}
\bibfield{author}{\bibinfo{person}{Patrick Esser}, \bibinfo{person}{Robin Rombach}, {and} \bibinfo{person}{Bjorn Ommer}.} \bibinfo{year}{2021}\natexlab{}.
\newblock \showarticletitle{Taming transformers for high-resolution image synthesis}. In \bibinfo{booktitle}{\emph{Proceedings of the IEEE/CVF conference on computer vision and pattern recognition}}. \bibinfo{pages}{12873--12883}.
\newblock


\bibitem[Fakoor et~al\mbox{.}(2019)]%
        {metalearning}
\bibfield{author}{\bibinfo{person}{Rasool Fakoor}, \bibinfo{person}{Pratik Chaudhari}, \bibinfo{person}{Stefano Soatto}, {and} \bibinfo{person}{Alexander~J Smola}.} \bibinfo{year}{2019}\natexlab{}.
\newblock \showarticletitle{Meta-q-learning}.
\newblock \bibinfo{journal}{\emph{arXiv preprint arXiv:1910.00125}} (\bibinfo{year}{2019}).
\newblock


\bibitem[Finn et~al\mbox{.}(2017)]%
        {MAML}
\bibfield{author}{\bibinfo{person}{Chelsea Finn}, \bibinfo{person}{Pieter Abbeel}, {and} \bibinfo{person}{Sergey Levine}.} \bibinfo{year}{2017}\natexlab{}.
\newblock \showarticletitle{Model-agnostic meta-learning for fast adaptation of deep networks}. In \bibinfo{booktitle}{\emph{International conference on machine learning}}. PMLR, \bibinfo{pages}{1126--1135}.
\newblock


\bibitem[Florence et~al\mbox{.}(2022)]%
        {florence2022implicit}
\bibfield{author}{\bibinfo{person}{Pete Florence}, \bibinfo{person}{Corey Lynch}, \bibinfo{person}{Andy Zeng}, \bibinfo{person}{Oscar~A Ramirez}, \bibinfo{person}{Ayzaan Wahid}, \bibinfo{person}{Laura Downs}, \bibinfo{person}{Adrian Wong}, \bibinfo{person}{Johnny Lee}, \bibinfo{person}{Igor Mordatch}, {and} \bibinfo{person}{Jonathan Tompson}.} \bibinfo{year}{2022}\natexlab{}.
\newblock \showarticletitle{Implicit behavioral cloning}. In \bibinfo{booktitle}{\emph{Conference on robot learning}}. PMLR, \bibinfo{pages}{158--168}.
\newblock


\bibitem[Floridi and Chiriatti(2020)]%
        {GPT3}
\bibfield{author}{\bibinfo{person}{Luciano Floridi} {and} \bibinfo{person}{Massimo Chiriatti}.} \bibinfo{year}{2020}\natexlab{}.
\newblock \showarticletitle{GPT-3: Its nature, scope, limits, and consequences}.
\newblock \bibinfo{journal}{\emph{Minds and machines}} \bibinfo{volume}{30}, \bibinfo{number}{4} (\bibinfo{year}{2020}), \bibinfo{pages}{681--694}.
\newblock


\bibitem[Fox and Long(2003)]%
        {fox2003pddl2}
\bibfield{author}{\bibinfo{person}{Maria Fox} {and} \bibinfo{person}{Derek Long}.} \bibinfo{year}{2003}\natexlab{}.
\newblock \showarticletitle{PDDL2. 1: An extension to PDDL for expressing temporal planning domains}.
\newblock \bibinfo{journal}{\emph{Journal of artificial intelligence research}}  \bibinfo{volume}{20} (\bibinfo{year}{2003}), \bibinfo{pages}{61--124}.
\newblock


\bibitem[Franklin et~al\mbox{.}(2002)]%
        {franklin2002feedback}
\bibfield{author}{\bibinfo{person}{Gene~F Franklin}, \bibinfo{person}{J~David Powell}, \bibinfo{person}{Abbas Emami-Naeini}, {and} \bibinfo{person}{J~David Powell}.} \bibinfo{year}{2002}\natexlab{}.
\newblock \bibinfo{booktitle}{\emph{Feedback control of dynamic systems}}. Vol.~\bibinfo{volume}{4}.
\newblock \bibinfo{publisher}{Prentice hall Upper Saddle River}.
\newblock


\bibitem[Fu et~al\mbox{.}(2020)]%
        {D4RL}
\bibfield{author}{\bibinfo{person}{Justin Fu}, \bibinfo{person}{Aviral Kumar}, \bibinfo{person}{Ofir Nachum}, \bibinfo{person}{George Tucker}, {and} \bibinfo{person}{Sergey Levine}.} \bibinfo{year}{2020}\natexlab{}.
\newblock \showarticletitle{D4rl: Datasets for deep data-driven reinforcement learning}.
\newblock \bibinfo{journal}{\emph{arXiv preprint arXiv:2004.07219}} (\bibinfo{year}{2020}).
\newblock


\bibitem[Fu et~al\mbox{.}(2024)]%
        {MobileALOHA}
\bibfield{author}{\bibinfo{person}{Zipeng Fu}, \bibinfo{person}{Tony~Z Zhao}, {and} \bibinfo{person}{Chelsea Finn}.} \bibinfo{year}{2024}\natexlab{}.
\newblock \showarticletitle{Mobile aloha: Learning bimanual mobile manipulation with low-cost whole-body teleoperation}.
\newblock \bibinfo{journal}{\emph{arXiv preprint arXiv:2401.02117}} (\bibinfo{year}{2024}).
\newblock


\bibitem[Gelfond and Kahl(2014)]%
        {gelfond2014knowledge}
\bibfield{author}{\bibinfo{person}{Michael Gelfond} {and} \bibinfo{person}{Yulia Kahl}.} \bibinfo{year}{2014}\natexlab{}.
\newblock \bibinfo{booktitle}{\emph{Knowledge representation, reasoning, and the design of intelligent agents: The answer-set programming approach}}.
\newblock \bibinfo{publisher}{Cambridge University Press}.
\newblock


\bibitem[Goyal et~al\mbox{.}(2023)]%
        {RVT}
\bibfield{author}{\bibinfo{person}{Ankit Goyal}, \bibinfo{person}{Jie Xu}, \bibinfo{person}{Yijie Guo}, \bibinfo{person}{Valts Blukis}, \bibinfo{person}{Yu-Wei Chao}, {and} \bibinfo{person}{Dieter Fox}.} \bibinfo{year}{2023}\natexlab{}.
\newblock \showarticletitle{Rvt: Robotic view transformer for 3d object manipulation}. In \bibinfo{booktitle}{\emph{Conference on Robot Learning}}. PMLR, \bibinfo{pages}{694--710}.
\newblock


\bibitem[Grauman et~al\mbox{.}(2022)]%
        {Ego4D}
\bibfield{author}{\bibinfo{person}{Kristen Grauman}, \bibinfo{person}{Andrew Westbury}, \bibinfo{person}{Eugene Byrne}, \bibinfo{person}{Zachary Chavis}, \bibinfo{person}{Antonino Furnari}, \bibinfo{person}{Rohit Girdhar}, \bibinfo{person}{Jackson Hamburger}, \bibinfo{person}{Hao Jiang}, \bibinfo{person}{Miao Liu}, \bibinfo{person}{Xingyu Liu}, {et~al\mbox{.}}} \bibinfo{year}{2022}\natexlab{}.
\newblock \showarticletitle{Ego4d: Around the world in 3,000 hours of egocentric video}. In \bibinfo{booktitle}{\emph{Proceedings of the IEEE/CVF conference on computer vision and pattern recognition}}. \bibinfo{pages}{18995--19012}.
\newblock


\bibitem[Gu et~al\mbox{.}(2023a)]%
        {RT-Trajectory}
\bibfield{author}{\bibinfo{person}{Jiayuan Gu}, \bibinfo{person}{Sean Kirmani}, \bibinfo{person}{Paul Wohlhart}, \bibinfo{person}{Yao Lu}, \bibinfo{person}{Montserrat~Gonzalez Arenas}, \bibinfo{person}{Kanishka Rao}, \bibinfo{person}{Wenhao Yu}, \bibinfo{person}{Chuyuan Fu}, \bibinfo{person}{Keerthana Gopalakrishnan}, \bibinfo{person}{Zhuo Xu}, {et~al\mbox{.}}} \bibinfo{year}{2023}\natexlab{a}.
\newblock \showarticletitle{Rt-trajectory: Robotic task generalization via hindsight trajectory sketches}.
\newblock \bibinfo{journal}{\emph{arXiv preprint arXiv:2311.01977}} (\bibinfo{year}{2023}).
\newblock


\bibitem[Gu et~al\mbox{.}(2023b)]%
        {seer}
\bibfield{author}{\bibinfo{person}{Xianfan Gu}, \bibinfo{person}{Chuan Wen}, \bibinfo{person}{Weirui Ye}, \bibinfo{person}{Jiaming Song}, {and} \bibinfo{person}{Yang Gao}.} \bibinfo{year}{2023}\natexlab{b}.
\newblock \showarticletitle{Seer: Language instructed video prediction with latent diffusion models}.
\newblock \bibinfo{journal}{\emph{arXiv preprint arXiv:2303.14897}} (\bibinfo{year}{2023}).
\newblock


\bibitem[Guan et~al\mbox{.}(2023)]%
        {PDDL-WM}
\bibfield{author}{\bibinfo{person}{Lin Guan}, \bibinfo{person}{Karthik Valmeekam}, \bibinfo{person}{Sarath Sreedharan}, {and} \bibinfo{person}{Subbarao Kambhampati}.} \bibinfo{year}{2023}\natexlab{}.
\newblock \showarticletitle{Leveraging pre-trained large language models to construct and utilize world models for model-based task planning}.
\newblock \bibinfo{journal}{\emph{Advances in Neural Information Processing Systems}}  \bibinfo{volume}{36} (\bibinfo{year}{2023}), \bibinfo{pages}{79081--79094}.
\newblock


\bibitem[Guo et~al\mbox{.}(2024)]%
        {DoReMi}
\bibfield{author}{\bibinfo{person}{Yanjiang Guo}, \bibinfo{person}{Yen-Jen Wang}, \bibinfo{person}{Lihan Zha}, {and} \bibinfo{person}{Jianyu Chen}.} \bibinfo{year}{2024}\natexlab{}.
\newblock \showarticletitle{Doremi: Grounding language model by detecting and recovering from plan-execution misalignment}. In \bibinfo{booktitle}{\emph{2024 IEEE/RSJ International Conference on Intelligent Robots and Systems (IROS)}}. IEEE, \bibinfo{pages}{12124--12131}.
\newblock


\bibitem[Gupta et~al\mbox{.}(2018)]%
        {gupta2018meta}
\bibfield{author}{\bibinfo{person}{Abhishek Gupta}, \bibinfo{person}{Russell Mendonca}, \bibinfo{person}{YuXuan Liu}, \bibinfo{person}{Pieter Abbeel}, {and} \bibinfo{person}{Sergey Levine}.} \bibinfo{year}{2018}\natexlab{}.
\newblock \showarticletitle{Meta-reinforcement learning of structured exploration strategies}.
\newblock \bibinfo{journal}{\emph{Advances in neural information processing systems}}  \bibinfo{volume}{31} (\bibinfo{year}{2018}).
\newblock


\bibitem[Ha and Schmidhuber(2018)]%
        {ha2018recurrent}
\bibfield{author}{\bibinfo{person}{David Ha} {and} \bibinfo{person}{J{\"u}rgen Schmidhuber}.} \bibinfo{year}{2018}\natexlab{}.
\newblock \showarticletitle{Recurrent world models facilitate policy evolution}.
\newblock \bibinfo{journal}{\emph{Advances in neural information processing systems}}  \bibinfo{volume}{31} (\bibinfo{year}{2018}).
\newblock


\bibitem[Haarnoja et~al\mbox{.}(2018)]%
        {SAC}
\bibfield{author}{\bibinfo{person}{Tuomas Haarnoja}, \bibinfo{person}{Aurick Zhou}, \bibinfo{person}{Pieter Abbeel}, {and} \bibinfo{person}{Sergey Levine}.} \bibinfo{year}{2018}\natexlab{}.
\newblock \showarticletitle{Soft actor-critic: Off-policy maximum entropy deep reinforcement learning with a stochastic actor}. In \bibinfo{booktitle}{\emph{International conference on machine learning}}. Pmlr, \bibinfo{pages}{1861--1870}.
\newblock


\bibitem[Hafner et~al\mbox{.}(2019a)]%
        {hafner2019dream}
\bibfield{author}{\bibinfo{person}{Danijar Hafner}, \bibinfo{person}{Timothy Lillicrap}, \bibinfo{person}{Jimmy Ba}, {and} \bibinfo{person}{Mohammad Norouzi}.} \bibinfo{year}{2019}\natexlab{a}.
\newblock \showarticletitle{Dream to control: Learning behaviors by latent imagination}.
\newblock \bibinfo{journal}{\emph{arXiv preprint arXiv:1912.01603}} (\bibinfo{year}{2019}).
\newblock


\bibitem[Hafner et~al\mbox{.}(2019b)]%
        {DreamerV1}
\bibfield{author}{\bibinfo{person}{Danijar Hafner}, \bibinfo{person}{Timothy Lillicrap}, \bibinfo{person}{Jimmy Ba}, {and} \bibinfo{person}{Mohammad Norouzi}.} \bibinfo{year}{2019}\natexlab{b}.
\newblock \showarticletitle{Dream to control: Learning behaviors by latent imagination}.
\newblock \bibinfo{journal}{\emph{arXiv preprint arXiv:1912.01603}} (\bibinfo{year}{2019}).
\newblock


\bibitem[Hafner et~al\mbox{.}(2019c)]%
        {PlaNet}
\bibfield{author}{\bibinfo{person}{Danijar Hafner}, \bibinfo{person}{Timothy Lillicrap}, \bibinfo{person}{Ian Fischer}, \bibinfo{person}{Ruben Villegas}, \bibinfo{person}{David Ha}, \bibinfo{person}{Honglak Lee}, {and} \bibinfo{person}{James Davidson}.} \bibinfo{year}{2019}\natexlab{c}.
\newblock \showarticletitle{Learning latent dynamics for planning from pixels}. In \bibinfo{booktitle}{\emph{International conference on machine learning}}. PMLR, \bibinfo{pages}{2555--2565}.
\newblock


\bibitem[Hafner et~al\mbox{.}(2020)]%
        {DreamerV2}
\bibfield{author}{\bibinfo{person}{Danijar Hafner}, \bibinfo{person}{Timothy Lillicrap}, \bibinfo{person}{Mohammad Norouzi}, {and} \bibinfo{person}{Jimmy Ba}.} \bibinfo{year}{2020}\natexlab{}.
\newblock \showarticletitle{Mastering atari with discrete world models}.
\newblock \bibinfo{journal}{\emph{arXiv preprint arXiv:2010.02193}} (\bibinfo{year}{2020}).
\newblock


\bibitem[Hafner et~al\mbox{.}(2023)]%
        {DreamerV3}
\bibfield{author}{\bibinfo{person}{Danijar Hafner}, \bibinfo{person}{Jurgis Pasukonis}, \bibinfo{person}{Jimmy Ba}, {and} \bibinfo{person}{Timothy Lillicrap}.} \bibinfo{year}{2023}\natexlab{}.
\newblock \showarticletitle{Mastering diverse domains through world models}.
\newblock \bibinfo{journal}{\emph{arXiv preprint arXiv:2301.04104}} (\bibinfo{year}{2023}).
\newblock


\bibitem[Hancock et~al\mbox{.}(2024)]%
        {BYO-VLA}
\bibfield{author}{\bibinfo{person}{Asher~J Hancock}, \bibinfo{person}{Allen~Z Ren}, {and} \bibinfo{person}{Anirudha Majumdar}.} \bibinfo{year}{2024}\natexlab{}.
\newblock \showarticletitle{Run-time observation interventions make vision-language-action models more visually robust}.
\newblock \bibinfo{journal}{\emph{arXiv preprint arXiv:2410.01971}} (\bibinfo{year}{2024}).
\newblock


\bibitem[Haslum et~al\mbox{.}(2019)]%
        {haslum2019introduction}
\bibfield{author}{\bibinfo{person}{Patrik Haslum}, \bibinfo{person}{Nir Lipovetzky}, \bibinfo{person}{Daniele Magazzeni}, \bibinfo{person}{Christian Muise}, \bibinfo{person}{Ronald Brachman}, \bibinfo{person}{Francesca Rossi}, {and} \bibinfo{person}{Peter Stone}.} \bibinfo{year}{2019}\natexlab{}.
\newblock \bibinfo{booktitle}{\emph{An introduction to the planning domain definition language}}. Vol.~\bibinfo{volume}{13}.
\newblock \bibinfo{publisher}{Springer}.
\newblock


\bibitem[He et~al\mbox{.}(2024)]%
        {VPDD}
\bibfield{author}{\bibinfo{person}{Haoran He}, \bibinfo{person}{Chenjia Bai}, \bibinfo{person}{Ling Pan}, \bibinfo{person}{Weinan Zhang}, \bibinfo{person}{Bin Zhao}, {and} \bibinfo{person}{Xuelong Li}.} \bibinfo{year}{2024}\natexlab{}.
\newblock \showarticletitle{Large-scale actionless video pre-training via discrete diffusion for efficient policy learning}.
\newblock \bibinfo{journal}{\emph{CoRR}} (\bibinfo{year}{2024}).
\newblock


\bibitem[He et~al\mbox{.}(2023)]%
        {MTDiff}
\bibfield{author}{\bibinfo{person}{Haoran He}, \bibinfo{person}{Chenjia Bai}, \bibinfo{person}{Kang Xu}, \bibinfo{person}{Zhuoran Yang}, \bibinfo{person}{Weinan Zhang}, \bibinfo{person}{Dong Wang}, \bibinfo{person}{Bin Zhao}, {and} \bibinfo{person}{Xuelong Li}.} \bibinfo{year}{2023}\natexlab{}.
\newblock \showarticletitle{Diffusion model is an effective planner and data synthesizer for multi-task reinforcement learning}.
\newblock \bibinfo{journal}{\emph{Advances in neural information processing systems}}  \bibinfo{volume}{36} (\bibinfo{year}{2023}), \bibinfo{pages}{64896--64917}.
\newblock


\bibitem[He et~al\mbox{.}(2022)]%
        {MAE}
\bibfield{author}{\bibinfo{person}{Kaiming He}, \bibinfo{person}{Xinlei Chen}, \bibinfo{person}{Saining Xie}, \bibinfo{person}{Yanghao Li}, \bibinfo{person}{Piotr Doll{\'a}r}, {and} \bibinfo{person}{Ross Girshick}.} \bibinfo{year}{2022}\natexlab{}.
\newblock \showarticletitle{Masked autoencoders are scalable vision learners}. In \bibinfo{booktitle}{\emph{Proceedings of the IEEE/CVF conference on computer vision and pattern recognition}}. \bibinfo{pages}{16000--16009}.
\newblock


\bibitem[Ho and Ermon(2016)]%
        {ho2016generative}
\bibfield{author}{\bibinfo{person}{Jonathan Ho} {and} \bibinfo{person}{Stefano Ermon}.} \bibinfo{year}{2016}\natexlab{}.
\newblock \showarticletitle{Generative adversarial imitation learning}.
\newblock \bibinfo{journal}{\emph{Advances in neural information processing systems}}  \bibinfo{volume}{29} (\bibinfo{year}{2016}).
\newblock


\bibitem[Hori et~al\mbox{.}(2024)]%
        {IRAP}
\bibfield{author}{\bibinfo{person}{Kazuki Hori}, \bibinfo{person}{Kanata Suzuki}, {and} \bibinfo{person}{Tetsuya Ogata}.} \bibinfo{year}{2024}\natexlab{}.
\newblock \showarticletitle{Interactively robot action planning with uncertainty analysis and active questioning by large language model}. In \bibinfo{booktitle}{\emph{2024 IEEE/SICE International Symposium on System Integration (SII)}}. IEEE, \bibinfo{pages}{85--91}.
\newblock


\bibitem[Hou et~al\mbox{.}(2025)]%
        {MCP}
\bibfield{author}{\bibinfo{person}{Xinyi Hou}, \bibinfo{person}{Yanjie Zhao}, \bibinfo{person}{Shenao Wang}, {and} \bibinfo{person}{Haoyu Wang}.} \bibinfo{year}{2025}\natexlab{}.
\newblock \showarticletitle{Model context protocol (mcp): Landscape, security threats, and future research directions}.
\newblock \bibinfo{journal}{\emph{arXiv preprint arXiv:2503.23278}} (\bibinfo{year}{2025}).
\newblock


\bibitem[Hu et~al\mbox{.}(2022)]%
        {LoRA}
\bibfield{author}{\bibinfo{person}{Edward~J Hu}, \bibinfo{person}{Yelong Shen}, \bibinfo{person}{Phillip Wallis}, \bibinfo{person}{Zeyuan Allen-Zhu}, \bibinfo{person}{Yuanzhi Li}, \bibinfo{person}{Shean Wang}, \bibinfo{person}{Lu Wang}, \bibinfo{person}{Weizhu Chen}, {et~al\mbox{.}}} \bibinfo{year}{2022}\natexlab{}.
\newblock \showarticletitle{Lora: Low-rank adaptation of large language models.}
\newblock \bibinfo{journal}{\emph{ICLR}} \bibinfo{volume}{1}, \bibinfo{number}{2} (\bibinfo{year}{2022}), \bibinfo{pages}{3}.
\newblock


\bibitem[Hu et~al\mbox{.}(2023)]%
        {ViLA}
\bibfield{author}{\bibinfo{person}{Yingdong Hu}, \bibinfo{person}{Fanqi Lin}, \bibinfo{person}{Tong Zhang}, \bibinfo{person}{Li Yi}, {and} \bibinfo{person}{Yang Gao}.} \bibinfo{year}{2023}\natexlab{}.
\newblock \showarticletitle{Look before you leap: Unveiling the power of gpt-4v in robotic vision-language planning}.
\newblock \bibinfo{journal}{\emph{arXiv preprint arXiv:2311.17842}} (\bibinfo{year}{2023}).
\newblock


\bibitem[Huang et~al\mbox{.}(2023a)]%
        {Instruct2Act}
\bibfield{author}{\bibinfo{person}{Siyuan Huang}, \bibinfo{person}{Zhengkai Jiang}, \bibinfo{person}{Hao Dong}, \bibinfo{person}{Yu Qiao}, \bibinfo{person}{Peng Gao}, {and} \bibinfo{person}{Hongsheng Li}.} \bibinfo{year}{2023}\natexlab{a}.
\newblock \showarticletitle{Instruct2act: Mapping multi-modality instructions to robotic actions with large language model}.
\newblock \bibinfo{journal}{\emph{arXiv preprint arXiv:2305.11176}} (\bibinfo{year}{2023}).
\newblock


\bibitem[Huang et~al\mbox{.}(2022a)]%
        {Zero-shot}
\bibfield{author}{\bibinfo{person}{Wenlong Huang}, \bibinfo{person}{Pieter Abbeel}, \bibinfo{person}{Deepak Pathak}, {and} \bibinfo{person}{Igor Mordatch}.} \bibinfo{year}{2022}\natexlab{a}.
\newblock \showarticletitle{Language models as zero-shot planners: Extracting actionable knowledge for embodied agents}. In \bibinfo{booktitle}{\emph{International conference on machine learning}}. PMLR, \bibinfo{pages}{9118--9147}.
\newblock


\bibitem[Huang et~al\mbox{.}(2023b)]%
        {Voxposer}
\bibfield{author}{\bibinfo{person}{Wenlong Huang}, \bibinfo{person}{Chen Wang}, \bibinfo{person}{Ruohan Zhang}, \bibinfo{person}{Yunzhu Li}, \bibinfo{person}{Jiajun Wu}, {and} \bibinfo{person}{Li Fei-Fei}.} \bibinfo{year}{2023}\natexlab{b}.
\newblock \showarticletitle{Voxposer: Composable 3d value maps for robotic manipulation with language models}.
\newblock \bibinfo{journal}{\emph{arXiv preprint arXiv:2307.05973}} (\bibinfo{year}{2023}).
\newblock


\bibitem[Huang et~al\mbox{.}(2023c)]%
        {GroundedDecoding}
\bibfield{author}{\bibinfo{person}{Wenlong Huang}, \bibinfo{person}{Fei Xia}, \bibinfo{person}{Dhruv Shah}, \bibinfo{person}{Danny Driess}, \bibinfo{person}{Andy Zeng}, \bibinfo{person}{Yao Lu}, \bibinfo{person}{Pete Florence}, \bibinfo{person}{Igor Mordatch}, \bibinfo{person}{Sergey Levine}, \bibinfo{person}{Karol Hausman}, {et~al\mbox{.}}} \bibinfo{year}{2023}\natexlab{c}.
\newblock \showarticletitle{Grounded decoding: Guiding text generation with grounded models for embodied agents}.
\newblock \bibinfo{journal}{\emph{Advances in Neural Information Processing Systems}}  \bibinfo{volume}{36} (\bibinfo{year}{2023}), \bibinfo{pages}{59636--59661}.
\newblock


\bibitem[Huang et~al\mbox{.}(2022b)]%
        {InnerMonologue}
\bibfield{author}{\bibinfo{person}{Wenlong Huang}, \bibinfo{person}{Fei Xia}, \bibinfo{person}{Ted Xiao}, \bibinfo{person}{Harris Chan}, \bibinfo{person}{Jacky Liang}, \bibinfo{person}{Pete Florence}, \bibinfo{person}{Andy Zeng}, \bibinfo{person}{Jonathan Tompson}, \bibinfo{person}{Igor Mordatch}, \bibinfo{person}{Yevgen Chebotar}, {et~al\mbox{.}}} \bibinfo{year}{2022}\natexlab{b}.
\newblock \showarticletitle{Inner monologue: Embodied reasoning through planning with language models}.
\newblock \bibinfo{journal}{\emph{arXiv preprint arXiv:2207.05608}} (\bibinfo{year}{2022}).
\newblock


\bibitem[Jiang et~al\mbox{.}(2022)]%
        {VIMA}
\bibfield{author}{\bibinfo{person}{Yunfan Jiang}, \bibinfo{person}{Agrim Gupta}, \bibinfo{person}{Zichen Zhang}, \bibinfo{person}{Guanzhi Wang}, \bibinfo{person}{Yongqiang Dou}, \bibinfo{person}{Yanjun Chen}, \bibinfo{person}{Li Fei-Fei}, \bibinfo{person}{Anima Anandkumar}, \bibinfo{person}{Yuke Zhu}, {and} \bibinfo{person}{Linxi Fan}.} \bibinfo{year}{2022}\natexlab{}.
\newblock \showarticletitle{Vima: General robot manipulation with multimodal prompts}.
\newblock \bibinfo{journal}{\emph{arXiv preprint arXiv:2210.03094}} \bibinfo{volume}{2}, \bibinfo{number}{3} (\bibinfo{year}{2022}), \bibinfo{pages}{6}.
\newblock


\bibitem[Jiang et~al\mbox{.}(2019)]%
        {jiang2019multi}
\bibfield{author}{\bibinfo{person}{Yuqian Jiang}, \bibinfo{person}{Harel Yedidsion}, \bibinfo{person}{Shiqi Zhang}, \bibinfo{person}{Guni Sharon}, {and} \bibinfo{person}{Peter Stone}.} \bibinfo{year}{2019}\natexlab{}.
\newblock \showarticletitle{Multi-robot planning with conflicts and synergies}.
\newblock \bibinfo{journal}{\emph{Autonomous Robots}} \bibinfo{volume}{43}, \bibinfo{number}{8} (\bibinfo{year}{2019}), \bibinfo{pages}{2011--2032}.
\newblock


\bibitem[Kang et~al\mbox{.}(2023)]%
        {EDP}
\bibfield{author}{\bibinfo{person}{Bingyi Kang}, \bibinfo{person}{Xiao Ma}, \bibinfo{person}{Chao Du}, \bibinfo{person}{Tianyu Pang}, {and} \bibinfo{person}{Shuicheng Yan}.} \bibinfo{year}{2023}\natexlab{}.
\newblock \showarticletitle{Efficient diffusion policies for offline reinforcement learning}.
\newblock \bibinfo{journal}{\emph{Advances in Neural Information Processing Systems}}  \bibinfo{volume}{36} (\bibinfo{year}{2023}), \bibinfo{pages}{67195--67212}.
\newblock


\bibitem[Kim et~al\mbox{.}(2025)]%
        {OpenVLA-OFT}
\bibfield{author}{\bibinfo{person}{Moo~Jin Kim}, \bibinfo{person}{Chelsea Finn}, {and} \bibinfo{person}{Percy Liang}.} \bibinfo{year}{2025}\natexlab{}.
\newblock \showarticletitle{Fine-tuning vision-language-action models: Optimizing speed and success}.
\newblock \bibinfo{journal}{\emph{arXiv preprint arXiv:2502.19645}} (\bibinfo{year}{2025}).
\newblock


\bibitem[Kim et~al\mbox{.}(2024b)]%
        {kim2024openvla}
\bibfield{author}{\bibinfo{person}{Moo~Jin Kim}, \bibinfo{person}{Karl Pertsch}, \bibinfo{person}{Siddharth Karamcheti}, \bibinfo{person}{Ted Xiao}, \bibinfo{person}{Ashwin Balakrishna}, \bibinfo{person}{Suraj Nair}, \bibinfo{person}{Rafael Rafailov}, \bibinfo{person}{Ethan Foster}, \bibinfo{person}{Grace Lam}, \bibinfo{person}{Pannag Sanketi}, {et~al\mbox{.}}} \bibinfo{year}{2024}\natexlab{b}.
\newblock \showarticletitle{Openvla: An open-source vision-language-action model}.
\newblock \bibinfo{journal}{\emph{arXiv preprint arXiv:2406.09246}} (\bibinfo{year}{2024}).
\newblock


\bibitem[Kim et~al\mbox{.}(2024c)]%
        {OpenVLA}
\bibfield{author}{\bibinfo{person}{Moo~Jin Kim}, \bibinfo{person}{Karl Pertsch}, \bibinfo{person}{Siddharth Karamcheti}, \bibinfo{person}{Ted Xiao}, \bibinfo{person}{Ashwin Balakrishna}, \bibinfo{person}{Suraj Nair}, \bibinfo{person}{Rafael Rafailov}, \bibinfo{person}{Ethan Foster}, \bibinfo{person}{Grace Lam}, \bibinfo{person}{Pannag Sanketi}, {et~al\mbox{.}}} \bibinfo{year}{2024}\natexlab{c}.
\newblock \showarticletitle{Openvla: An open-source vision-language-action model}.
\newblock \bibinfo{journal}{\emph{arXiv preprint arXiv:2406.09246}} (\bibinfo{year}{2024}).
\newblock


\bibitem[Kim et~al\mbox{.}(2024a)]%
        {17}
\bibfield{author}{\bibinfo{person}{Yeseung Kim}, \bibinfo{person}{Dohyun Kim}, \bibinfo{person}{Jieun Choi}, \bibinfo{person}{Jisang Park}, \bibinfo{person}{Nayoung Oh}, {and} \bibinfo{person}{Daehyung Park}.} \bibinfo{year}{2024}\natexlab{a}.
\newblock \showarticletitle{A survey on integration of large language models with intelligent robots}.
\newblock \bibinfo{journal}{\emph{Intelligent Service Robotics}} \bibinfo{volume}{17}, \bibinfo{number}{5} (\bibinfo{year}{2024}), \bibinfo{pages}{1091--1107}.
\newblock


\bibitem[Kirillov et~al\mbox{.}(2023a)]%
        {kirillov2023segment}
\bibfield{author}{\bibinfo{person}{Alexander Kirillov}, \bibinfo{person}{Eric Mintun}, \bibinfo{person}{Nikhila Ravi}, \bibinfo{person}{Hanzi Mao}, \bibinfo{person}{Chloe Rolland}, \bibinfo{person}{Laura Gustafson}, \bibinfo{person}{Tete Xiao}, \bibinfo{person}{Spencer Whitehead}, \bibinfo{person}{Alexander~C Berg}, \bibinfo{person}{Wan-Yen Lo}, {et~al\mbox{.}}} \bibinfo{year}{2023}\natexlab{a}.
\newblock \showarticletitle{Segment anything}. In \bibinfo{booktitle}{\emph{Proceedings of the IEEE/CVF international conference on computer vision}}. \bibinfo{pages}{4015--4026}.
\newblock


\bibitem[Kirillov et~al\mbox{.}(2023b)]%
        {sam}
\bibfield{author}{\bibinfo{person}{Alexander Kirillov}, \bibinfo{person}{Eric Mintun}, \bibinfo{person}{Nikhila Ravi}, \bibinfo{person}{Hanzi Mao}, \bibinfo{person}{Chloe Rolland}, \bibinfo{person}{Laura Gustafson}, \bibinfo{person}{Tete Xiao}, \bibinfo{person}{Spencer Whitehead}, \bibinfo{person}{Alexander~C Berg}, \bibinfo{person}{Wan-Yen Lo}, {et~al\mbox{.}}} \bibinfo{year}{2023}\natexlab{b}.
\newblock \showarticletitle{Segment anything}. In \bibinfo{booktitle}{\emph{Proceedings of the IEEE/CVF international conference on computer vision}}. \bibinfo{pages}{4015--4026}.
\newblock


\bibitem[Kirkpatrick et~al\mbox{.}(2017)]%
        {kirkpatrick2017overcoming}
\bibfield{author}{\bibinfo{person}{James Kirkpatrick}, \bibinfo{person}{Razvan Pascanu}, \bibinfo{person}{Neil Rabinowitz}, \bibinfo{person}{Joel Veness}, \bibinfo{person}{Guillaume Desjardins}, \bibinfo{person}{Andrei~A Rusu}, \bibinfo{person}{Kieran Milan}, \bibinfo{person}{John Quan}, \bibinfo{person}{Tiago Ramalho}, \bibinfo{person}{Agnieszka Grabska-Barwinska}, {et~al\mbox{.}}} \bibinfo{year}{2017}\natexlab{}.
\newblock \showarticletitle{Overcoming catastrophic forgetting in neural networks}.
\newblock \bibinfo{journal}{\emph{Proceedings of the national academy of sciences}} \bibinfo{volume}{114}, \bibinfo{number}{13} (\bibinfo{year}{2017}), \bibinfo{pages}{3521--3526}.
\newblock


\bibitem[Krizhevsky et~al\mbox{.}(2012)]%
        {krizhevsky2012imagenet}
\bibfield{author}{\bibinfo{person}{Alex Krizhevsky}, \bibinfo{person}{Ilya Sutskever}, {and} \bibinfo{person}{Geoffrey~E Hinton}.} \bibinfo{year}{2012}\natexlab{}.
\newblock \showarticletitle{Imagenet classification with deep convolutional neural networks}.
\newblock \bibinfo{journal}{\emph{Advances in neural information processing systems}}  \bibinfo{volume}{25} (\bibinfo{year}{2012}).
\newblock


\bibitem[Kumar et~al\mbox{.}(2022)]%
        {kumar2022fine}
\bibfield{author}{\bibinfo{person}{Ananya Kumar}, \bibinfo{person}{Aditi Raghunathan}, \bibinfo{person}{Robbie Jones}, \bibinfo{person}{Tengyu Ma}, {and} \bibinfo{person}{Percy Liang}.} \bibinfo{year}{2022}\natexlab{}.
\newblock \showarticletitle{Fine-tuning can distort pretrained features and underperform out-of-distribution}.
\newblock \bibinfo{journal}{\emph{arXiv preprint arXiv:2202.10054}} (\bibinfo{year}{2022}).
\newblock


\bibitem[Lange et~al\mbox{.}(2012)]%
        {lange2012batch}
\bibfield{author}{\bibinfo{person}{Sascha Lange}, \bibinfo{person}{Thomas Gabel}, {and} \bibinfo{person}{Martin Riedmiller}.} \bibinfo{year}{2012}\natexlab{}.
\newblock \showarticletitle{Batch reinforcement learning}.
\newblock In \bibinfo{booktitle}{\emph{Reinforcement learning: State-of-the-art}}. \bibinfo{publisher}{Springer}, \bibinfo{pages}{45--73}.
\newblock


\bibitem[LeCun(2022)]%
        {JEPA}
\bibfield{author}{\bibinfo{person}{Yann LeCun}.} \bibinfo{year}{2022}\natexlab{}.
\newblock \showarticletitle{A path towards autonomous machine intelligence version 0.9. 2, 2022-06-27}.
\newblock \bibinfo{journal}{\emph{Open Review}} \bibinfo{volume}{62}, \bibinfo{number}{1} (\bibinfo{year}{2022}), \bibinfo{pages}{1--62}.
\newblock


\bibitem[Lewis et~al\mbox{.}(2020)]%
        {RAG}
\bibfield{author}{\bibinfo{person}{Patrick Lewis}, \bibinfo{person}{Ethan Perez}, \bibinfo{person}{Aleksandra Piktus}, \bibinfo{person}{Fabio Petroni}, \bibinfo{person}{Vladimir Karpukhin}, \bibinfo{person}{Naman Goyal}, \bibinfo{person}{Heinrich K{\"u}ttler}, \bibinfo{person}{Mike Lewis}, \bibinfo{person}{Wen-tau Yih}, \bibinfo{person}{Tim Rockt{\"a}schel}, {et~al\mbox{.}}} \bibinfo{year}{2020}\natexlab{}.
\newblock \showarticletitle{Retrieval-augmented generation for knowledge-intensive nlp tasks}.
\newblock \bibinfo{journal}{\emph{Advances in neural information processing systems}}  \bibinfo{volume}{33} (\bibinfo{year}{2020}), \bibinfo{pages}{9459--9474}.
\newblock


\bibitem[Li et~al\mbox{.}(2024a)]%
        {li2024multimodal}
\bibfield{author}{\bibinfo{person}{Chunyuan Li}, \bibinfo{person}{Zhe Gan}, \bibinfo{person}{Zhengyuan Yang}, \bibinfo{person}{Jianwei Yang}, \bibinfo{person}{Linjie Li}, \bibinfo{person}{Lijuan Wang}, \bibinfo{person}{Jianfeng Gao}, {et~al\mbox{.}}} \bibinfo{year}{2024}\natexlab{a}.
\newblock \showarticletitle{Multimodal foundation models: From specialists to general-purpose assistants}.
\newblock \bibinfo{journal}{\emph{Foundations and Trends{\textregistered} in Computer Graphics and Vision}} \bibinfo{volume}{16}, \bibinfo{number}{1-2} (\bibinfo{year}{2024}), \bibinfo{pages}{1--214}.
\newblock


\bibitem[Li et~al\mbox{.}(2025)]%
        {PointVLA}
\bibfield{author}{\bibinfo{person}{Chengmeng Li}, \bibinfo{person}{Junjie Wen}, \bibinfo{person}{Yan Peng}, \bibinfo{person}{Yaxin Peng}, \bibinfo{person}{Feifei Feng}, {and} \bibinfo{person}{Yichen Zhu}.} \bibinfo{year}{2025}\natexlab{}.
\newblock \showarticletitle{Pointvla: Injecting the 3d world into vision-language-action models}.
\newblock \bibinfo{journal}{\emph{arXiv preprint arXiv:2503.07511}} (\bibinfo{year}{2025}).
\newblock


\bibitem[Li et~al\mbox{.}(2024b)]%
        {automc-Reward}
\bibfield{author}{\bibinfo{person}{Hao Li}, \bibinfo{person}{Xue Yang}, \bibinfo{person}{Zhaokai Wang}, \bibinfo{person}{Xizhou Zhu}, \bibinfo{person}{Jie Zhou}, \bibinfo{person}{Yu Qiao}, \bibinfo{person}{Xiaogang Wang}, \bibinfo{person}{Hongsheng Li}, \bibinfo{person}{Lewei Lu}, {and} \bibinfo{person}{Jifeng Dai}.} \bibinfo{year}{2024}\natexlab{b}.
\newblock \showarticletitle{Auto mc-reward: Automated dense reward design with large language models for minecraft}. In \bibinfo{booktitle}{\emph{Proceedings of the IEEE/CVF Conference on Computer Vision and Pattern Recognition}}. \bibinfo{pages}{16426--16435}.
\newblock


\bibitem[Li et~al\mbox{.}(2023b)]%
        {Blip-2}
\bibfield{author}{\bibinfo{person}{Junnan Li}, \bibinfo{person}{Dongxu Li}, \bibinfo{person}{Silvio Savarese}, {and} \bibinfo{person}{Steven Hoi}.} \bibinfo{year}{2023}\natexlab{b}.
\newblock \showarticletitle{Blip-2: Bootstrapping language-image pre-training with frozen image encoders and large language models}. In \bibinfo{booktitle}{\emph{International conference on machine learning}}. PMLR, \bibinfo{pages}{19730--19742}.
\newblock


\bibitem[Li et~al\mbox{.}(2022a)]%
        {Blip}
\bibfield{author}{\bibinfo{person}{Junnan Li}, \bibinfo{person}{Dongxu Li}, \bibinfo{person}{Caiming Xiong}, {and} \bibinfo{person}{Steven Hoi}.} \bibinfo{year}{2022}\natexlab{a}.
\newblock \showarticletitle{Blip: Bootstrapping language-image pre-training for unified vision-language understanding and generation}. In \bibinfo{booktitle}{\emph{International conference on machine learning}}. PMLR, \bibinfo{pages}{12888--12900}.
\newblock


\bibitem[Li et~al\mbox{.}(2023a)]%
        {Video-Chat}
\bibfield{author}{\bibinfo{person}{KunChang Li}, \bibinfo{person}{Yinan He}, \bibinfo{person}{Yi Wang}, \bibinfo{person}{Yizhuo Li}, \bibinfo{person}{Wenhai Wang}, \bibinfo{person}{Ping Luo}, \bibinfo{person}{Yali Wang}, \bibinfo{person}{Limin Wang}, {and} \bibinfo{person}{Yu Qiao}.} \bibinfo{year}{2023}\natexlab{a}.
\newblock \showarticletitle{Videochat: Chat-centric video understanding}.
\newblock \bibinfo{journal}{\emph{arXiv preprint arXiv:2305.06355}} (\bibinfo{year}{2023}).
\newblock


\bibitem[Li et~al\mbox{.}(2022b)]%
        {li2022pre}
\bibfield{author}{\bibinfo{person}{Shuang Li}, \bibinfo{person}{Xavier Puig}, \bibinfo{person}{Chris Paxton}, \bibinfo{person}{Yilun Du}, \bibinfo{person}{Clinton Wang}, \bibinfo{person}{Linxi Fan}, \bibinfo{person}{Tao Chen}, \bibinfo{person}{De-An Huang}, \bibinfo{person}{Ekin Aky{\"u}rek}, \bibinfo{person}{Anima Anandkumar}, {et~al\mbox{.}}} \bibinfo{year}{2022}\natexlab{b}.
\newblock \showarticletitle{Pre-trained language models for interactive decision-making}.
\newblock \bibinfo{journal}{\emph{Advances in Neural Information Processing Systems}}  \bibinfo{volume}{35} (\bibinfo{year}{2022}), \bibinfo{pages}{31199--31212}.
\newblock


\bibitem[Li et~al\mbox{.}(2023c)]%
        {li2023vision}
\bibfield{author}{\bibinfo{person}{Xinghang Li}, \bibinfo{person}{Minghuan Liu}, \bibinfo{person}{Hanbo Zhang}, \bibinfo{person}{Cunjun Yu}, \bibinfo{person}{Jie Xu}, \bibinfo{person}{Hongtao Wu}, \bibinfo{person}{Chilam Cheang}, \bibinfo{person}{Ya Jing}, \bibinfo{person}{Weinan Zhang}, \bibinfo{person}{Huaping Liu}, {et~al\mbox{.}}} \bibinfo{year}{2023}\natexlab{c}.
\newblock \showarticletitle{Vision-language foundation models as effective robot imitators}.
\newblock \bibinfo{journal}{\emph{arXiv preprint arXiv:2311.01378}} (\bibinfo{year}{2023}).
\newblock


\bibitem[Liang et~al\mbox{.}(2022)]%
        {CaP}
\bibfield{author}{\bibinfo{person}{Jacky Liang}, \bibinfo{person}{Wenlong Huang}, \bibinfo{person}{Fei Xia}, \bibinfo{person}{Peng Xu}, \bibinfo{person}{Karol Hausman}, \bibinfo{person}{Brian Ichter}, \bibinfo{person}{Pete Florence}, {and} \bibinfo{person}{Andy Zeng}.} \bibinfo{year}{2022}\natexlab{}.
\newblock \showarticletitle{Code as policies: Language model programs for embodied control}.
\newblock \bibinfo{journal}{\emph{arXiv preprint arXiv:2209.07753}} (\bibinfo{year}{2022}).
\newblock


\bibitem[Liang et~al\mbox{.}(2024)]%
        {liang2024survey}
\bibfield{author}{\bibinfo{person}{Zijing Liang}, \bibinfo{person}{Yanjie Xu}, \bibinfo{person}{Yifan Hong}, \bibinfo{person}{Penghui Shang}, \bibinfo{person}{Qi Wang}, \bibinfo{person}{Qiang Fu}, {and} \bibinfo{person}{Ke Liu}.} \bibinfo{year}{2024}\natexlab{}.
\newblock \showarticletitle{A survey of multimodel large language models}. In \bibinfo{booktitle}{\emph{Proceedings of the 3rd International Conference on Computer, Artificial Intelligence and Control Engineering}}. \bibinfo{pages}{405--409}.
\newblock


\bibitem[Lin et~al\mbox{.}(2023)]%
        {Text2Motion}
\bibfield{author}{\bibinfo{person}{Kevin Lin}, \bibinfo{person}{Christopher Agia}, \bibinfo{person}{Toki Migimatsu}, \bibinfo{person}{Marco Pavone}, {and} \bibinfo{person}{Jeannette Bohg}.} \bibinfo{year}{2023}\natexlab{}.
\newblock \showarticletitle{Text2motion: From natural language instructions to feasible plans}.
\newblock \bibinfo{journal}{\emph{Autonomous Robots}} \bibinfo{volume}{47}, \bibinfo{number}{8} (\bibinfo{year}{2023}), \bibinfo{pages}{1345--1365}.
\newblock


\bibitem[Liu et~al\mbox{.}(2024)]%
        {DeepSeek-V3}
\bibfield{author}{\bibinfo{person}{Aixin Liu}, \bibinfo{person}{Bei Feng}, \bibinfo{person}{Bing Xue}, \bibinfo{person}{Bingxuan Wang}, \bibinfo{person}{Bochao Wu}, \bibinfo{person}{Chengda Lu}, \bibinfo{person}{Chenggang Zhao}, \bibinfo{person}{Chengqi Deng}, \bibinfo{person}{Chenyu Zhang}, \bibinfo{person}{Chong Ruan}, {et~al\mbox{.}}} \bibinfo{year}{2024}\natexlab{}.
\newblock \showarticletitle{Deepseek-v3 technical report}.
\newblock \bibinfo{journal}{\emph{arXiv preprint arXiv:2412.19437}} (\bibinfo{year}{2024}).
\newblock


\bibitem[Liu et~al\mbox{.}(2023)]%
        {LLM+P}
\bibfield{author}{\bibinfo{person}{Bo Liu}, \bibinfo{person}{Yuqian Jiang}, \bibinfo{person}{Xiaohan Zhang}, \bibinfo{person}{Qiang Liu}, \bibinfo{person}{Shiqi Zhang}, \bibinfo{person}{Joydeep Biswas}, {and} \bibinfo{person}{Peter Stone}.} \bibinfo{year}{2023}\natexlab{}.
\newblock \showarticletitle{Llm+ p: Empowering large language models with optimal planning proficiency}.
\newblock \bibinfo{journal}{\emph{arXiv preprint arXiv:2304.11477}} (\bibinfo{year}{2023}).
\newblock


\bibitem[Liu et~al\mbox{.}(2025)]%
        {18}
\bibfield{author}{\bibinfo{person}{Yang Liu}, \bibinfo{person}{Weixing Chen}, \bibinfo{person}{Yongjie Bai}, \bibinfo{person}{Xiaodan Liang}, \bibinfo{person}{Guanbin Li}, \bibinfo{person}{Wen Gao}, {and} \bibinfo{person}{Liang Lin}.} \bibinfo{year}{2025}\natexlab{}.
\newblock \showarticletitle{Aligning cyber space with physical world: A comprehensive survey on embodied ai}.
\newblock \bibinfo{journal}{\emph{IEEE/ASME Transactions on Mechatronics}} (\bibinfo{year}{2025}).
\newblock


\bibitem[Lu et~al\mbox{.}(2023)]%
        {SynthER}
\bibfield{author}{\bibinfo{person}{Cong Lu}, \bibinfo{person}{Philip Ball}, \bibinfo{person}{Yee~Whye Teh}, {and} \bibinfo{person}{Jack Parker-Holder}.} \bibinfo{year}{2023}\natexlab{}.
\newblock \showarticletitle{Synthetic experience replay}.
\newblock \bibinfo{journal}{\emph{Advances in Neural Information Processing Systems}}  \bibinfo{volume}{36} (\bibinfo{year}{2023}), \bibinfo{pages}{46323--46344}.
\newblock


\bibitem[Ma et~al\mbox{.}(2024)]%
        {15}
\bibfield{author}{\bibinfo{person}{Yueen Ma}, \bibinfo{person}{Zixing Song}, \bibinfo{person}{Yuzheng Zhuang}, \bibinfo{person}{Jianye Hao}, {and} \bibinfo{person}{Irwin King}.} \bibinfo{year}{2024}\natexlab{}.
\newblock \showarticletitle{A survey on vision-language-action models for embodied ai}.
\newblock \bibinfo{journal}{\emph{arXiv preprint arXiv:2405.14093}} (\bibinfo{year}{2024}).
\newblock


\bibitem[Ma et~al\mbox{.}(2023)]%
        {Eureka}
\bibfield{author}{\bibinfo{person}{Yecheng~Jason Ma}, \bibinfo{person}{William Liang}, \bibinfo{person}{Guanzhi Wang}, \bibinfo{person}{De-An Huang}, \bibinfo{person}{Osbert Bastani}, \bibinfo{person}{Dinesh Jayaraman}, \bibinfo{person}{Yuke Zhu}, \bibinfo{person}{Linxi Fan}, {and} \bibinfo{person}{Anima Anandkumar}.} \bibinfo{year}{2023}\natexlab{}.
\newblock \showarticletitle{Eureka: Human-level reward design via coding large language models}.
\newblock \bibinfo{journal}{\emph{arXiv preprint arXiv:2310.12931}} (\bibinfo{year}{2023}).
\newblock


\bibitem[Madaan et~al\mbox{.}(2023)]%
        {Self-Refine}
\bibfield{author}{\bibinfo{person}{Aman Madaan}, \bibinfo{person}{Niket Tandon}, \bibinfo{person}{Prakhar Gupta}, \bibinfo{person}{Skyler Hallinan}, \bibinfo{person}{Luyu Gao}, \bibinfo{person}{Sarah Wiegreffe}, \bibinfo{person}{Uri Alon}, \bibinfo{person}{Nouha Dziri}, \bibinfo{person}{Shrimai Prabhumoye}, \bibinfo{person}{Yiming Yang}, {et~al\mbox{.}}} \bibinfo{year}{2023}\natexlab{}.
\newblock \showarticletitle{Self-refine: Iterative refinement with self-feedback}.
\newblock \bibinfo{journal}{\emph{Advances in Neural Information Processing Systems}}  \bibinfo{volume}{36} (\bibinfo{year}{2023}), \bibinfo{pages}{46534--46594}.
\newblock


\bibitem[Mai et~al\mbox{.}(2024)]%
        {14}
\bibfield{author}{\bibinfo{person}{Xinji Mai}, \bibinfo{person}{Zeng Tao}, \bibinfo{person}{Junxiong Lin}, \bibinfo{person}{Haoran Wang}, \bibinfo{person}{Yang Chang}, \bibinfo{person}{Yanlan Kang}, \bibinfo{person}{Yan Wang}, {and} \bibinfo{person}{Wenqiang Zhang}.} \bibinfo{year}{2024}\natexlab{}.
\newblock \showarticletitle{From efficient multimodal models to world models: A survey}.
\newblock \bibinfo{journal}{\emph{arXiv preprint arXiv:2407.00118}} (\bibinfo{year}{2024}).
\newblock


\bibitem[Mandi et~al\mbox{.}(2024)]%
        {RoCo}
\bibfield{author}{\bibinfo{person}{Zhao Mandi}, \bibinfo{person}{Shreeya Jain}, {and} \bibinfo{person}{Shuran Song}.} \bibinfo{year}{2024}\natexlab{}.
\newblock \showarticletitle{Roco: Dialectic multi-robot collaboration with large language models}. In \bibinfo{booktitle}{\emph{2024 IEEE International Conference on Robotics and Automation (ICRA)}}. IEEE, \bibinfo{pages}{286--299}.
\newblock


\bibitem[Maniatopoulos et~al\mbox{.}(2016)]%
        {maniatopoulos2016reactive}
\bibfield{author}{\bibinfo{person}{Spyros Maniatopoulos}, \bibinfo{person}{Philipp Schillinger}, \bibinfo{person}{Vitchyr Pong}, \bibinfo{person}{David~C Conner}, {and} \bibinfo{person}{Hadas Kress-Gazit}.} \bibinfo{year}{2016}\natexlab{}.
\newblock \showarticletitle{Reactive high-level behavior synthesis for an atlas humanoid robot}. In \bibinfo{booktitle}{\emph{2016 IEEE international conference on robotics and automation (ICRA)}}. IEEE, \bibinfo{pages}{4192--4199}.
\newblock


\bibitem[Mayne et~al\mbox{.}(2000)]%
        {LQR}
\bibfield{author}{\bibinfo{person}{David~Q Mayne}, \bibinfo{person}{James~B Rawlings}, \bibinfo{person}{Christopher~V Rao}, {and} \bibinfo{person}{Pierre~OM Scokaert}.} \bibinfo{year}{2000}\natexlab{}.
\newblock \showarticletitle{Constrained model predictive control: Stability and optimality}.
\newblock \bibinfo{journal}{\emph{Automatica}} \bibinfo{volume}{36}, \bibinfo{number}{6} (\bibinfo{year}{2000}), \bibinfo{pages}{789--814}.
\newblock


\bibitem[McDermott et~al\mbox{.}(1998)]%
        {McDermott1998PDDLthePD}
\bibfield{author}{\bibinfo{person}{Drew McDermott}, \bibinfo{person}{Malik Ghallab}, \bibinfo{person}{Adele~E. Howe}, \bibinfo{person}{Craig~A. Knoblock}, \bibinfo{person}{Ashwin Ram}, \bibinfo{person}{Manuela~M. Veloso}, \bibinfo{person}{Daniel~S. Weld}, {and} \bibinfo{person}{David~E. Wilkins}.} \bibinfo{year}{1998}\natexlab{}.
\newblock \showarticletitle{PDDL-the planning domain definition language}.
\newblock
\urldef\tempurl%
\url{https://api.semanticscholar.org/CorpusID:59656859}
\showURL{%
\tempurl}


\bibitem[Mehta et~al\mbox{.}(2023)]%
        {mehta2023empirical}
\bibfield{author}{\bibinfo{person}{Sanket~Vaibhav Mehta}, \bibinfo{person}{Darshan Patil}, \bibinfo{person}{Sarath Chandar}, {and} \bibinfo{person}{Emma Strubell}.} \bibinfo{year}{2023}\natexlab{}.
\newblock \showarticletitle{An empirical investigation of the role of pre-training in lifelong learning}.
\newblock \bibinfo{journal}{\emph{Journal of Machine Learning Research}} \bibinfo{volume}{24}, \bibinfo{number}{214} (\bibinfo{year}{2023}), \bibinfo{pages}{1--50}.
\newblock


\bibitem[Mendonca et~al\mbox{.}(2023)]%
        {SWIM}
\bibfield{author}{\bibinfo{person}{Russell Mendonca}, \bibinfo{person}{Shikhar Bahl}, {and} \bibinfo{person}{Deepak Pathak}.} \bibinfo{year}{2023}\natexlab{}.
\newblock \showarticletitle{Structured world models from human videos}.
\newblock \bibinfo{journal}{\emph{arXiv preprint arXiv:2308.10901}} (\bibinfo{year}{2023}).
\newblock


\bibitem[Micheli et~al\mbox{.}(2022)]%
        {IRIS}
\bibfield{author}{\bibinfo{person}{Vincent Micheli}, \bibinfo{person}{Eloi Alonso}, {and} \bibinfo{person}{Fran{\c{c}}ois Fleuret}.} \bibinfo{year}{2022}\natexlab{}.
\newblock \showarticletitle{Transformers are sample-efficient world models}.
\newblock \bibinfo{journal}{\emph{arXiv preprint arXiv:2209.00588}} (\bibinfo{year}{2022}).
\newblock


\bibitem[Mnih et~al\mbox{.}(2015)]%
        {DQN}
\bibfield{author}{\bibinfo{person}{Volodymyr Mnih}, \bibinfo{person}{Koray Kavukcuoglu}, \bibinfo{person}{David Silver}, \bibinfo{person}{Andrei~A Rusu}, \bibinfo{person}{Joel Veness}, \bibinfo{person}{Marc~G Bellemare}, \bibinfo{person}{Alex Graves}, \bibinfo{person}{Martin Riedmiller}, \bibinfo{person}{Andreas~K Fidjeland}, \bibinfo{person}{Georg Ostrovski}, {et~al\mbox{.}}} \bibinfo{year}{2015}\natexlab{}.
\newblock \showarticletitle{Human-level control through deep reinforcement learning}.
\newblock \bibinfo{journal}{\emph{nature}} \bibinfo{volume}{518}, \bibinfo{number}{7540} (\bibinfo{year}{2015}), \bibinfo{pages}{529--533}.
\newblock


\bibitem[Mu et~al\mbox{.}(2023a)]%
        {Embodied-GPT}
\bibfield{author}{\bibinfo{person}{Yao Mu}, \bibinfo{person}{Qinglong Zhang}, \bibinfo{person}{Mengkang Hu}, \bibinfo{person}{Wenhai Wang}, \bibinfo{person}{Mingyu Ding}, \bibinfo{person}{Jun Jin}, \bibinfo{person}{Bin Wang}, \bibinfo{person}{Jifeng Dai}, \bibinfo{person}{Yu Qiao}, {and} \bibinfo{person}{Ping Luo}.} \bibinfo{year}{2023}\natexlab{a}.
\newblock \showarticletitle{Embodiedgpt: Vision-language pre-training via embodied chain of thought}.
\newblock \bibinfo{journal}{\emph{Advances in Neural Information Processing Systems}}  \bibinfo{volume}{36} (\bibinfo{year}{2023}), \bibinfo{pages}{25081--25094}.
\newblock


\bibitem[Mu et~al\mbox{.}(2023b)]%
        {EmbodiedGPT}
\bibfield{author}{\bibinfo{person}{Yao Mu}, \bibinfo{person}{Qinglong Zhang}, \bibinfo{person}{Mengkang Hu}, \bibinfo{person}{Wenhai Wang}, \bibinfo{person}{Mingyu Ding}, \bibinfo{person}{Jun Jin}, \bibinfo{person}{Bin Wang}, \bibinfo{person}{Jifeng Dai}, \bibinfo{person}{Yu Qiao}, {and} \bibinfo{person}{Ping Luo}.} \bibinfo{year}{2023}\natexlab{b}.
\newblock \showarticletitle{Embodiedgpt: Vision-language pre-training via embodied chain of thought}.
\newblock \bibinfo{journal}{\emph{Advances in Neural Information Processing Systems}}  \bibinfo{volume}{36} (\bibinfo{year}{2023}), \bibinfo{pages}{25081--25094}.
\newblock


\bibitem[Murphy(2012)]%
        {murphy2012machine}
\bibfield{author}{\bibinfo{person}{Kevin~P Murphy}.} \bibinfo{year}{2012}\natexlab{}.
\newblock \bibinfo{booktitle}{\emph{Machine learning: a probabilistic perspective}}.
\newblock \bibinfo{publisher}{MIT press}.
\newblock


\bibitem[Newbury et~al\mbox{.}(2024)]%
        {newbury2024review}
\bibfield{author}{\bibinfo{person}{Rhys Newbury}, \bibinfo{person}{Jack Collins}, \bibinfo{person}{Kerry He}, \bibinfo{person}{Jiahe Pan}, \bibinfo{person}{Ingmar Posner}, \bibinfo{person}{David Howard}, {and} \bibinfo{person}{Akansel Cosgun}.} \bibinfo{year}{2024}\natexlab{}.
\newblock \showarticletitle{A review of differentiable simulators}.
\newblock \bibinfo{journal}{\emph{IEEE Access}} (\bibinfo{year}{2024}).
\newblock


\bibitem[Ng et~al\mbox{.}(2000)]%
        {ng2000algorithms}
\bibfield{author}{\bibinfo{person}{Andrew~Y Ng}, \bibinfo{person}{Stuart Russell}, {et~al\mbox{.}}} \bibinfo{year}{2000}\natexlab{}.
\newblock \showarticletitle{Algorithms for inverse reinforcement learning.}. In \bibinfo{booktitle}{\emph{Icml}}, Vol.~\bibinfo{volume}{1}. \bibinfo{pages}{2}.
\newblock


\bibitem[Oquab et~al\mbox{.}(2023)]%
        {DINOv2}
\bibfield{author}{\bibinfo{person}{Maxime Oquab}, \bibinfo{person}{Timoth{\'e}e Darcet}, \bibinfo{person}{Th{\'e}o Moutakanni}, \bibinfo{person}{Huy Vo}, \bibinfo{person}{Marc Szafraniec}, \bibinfo{person}{Vasil Khalidov}, \bibinfo{person}{Pierre Fernandez}, \bibinfo{person}{Daniel Haziza}, \bibinfo{person}{Francisco Massa}, \bibinfo{person}{Alaaeldin El-Nouby}, {et~al\mbox{.}}} \bibinfo{year}{2023}\natexlab{}.
\newblock \showarticletitle{Dinov2: Learning robust visual features without supervision}.
\newblock \bibinfo{journal}{\emph{arXiv preprint arXiv:2304.07193}} (\bibinfo{year}{2023}).
\newblock


\bibitem[Ouyang et~al\mbox{.}(2022a)]%
        {ouyang2022training}
\bibfield{author}{\bibinfo{person}{Long Ouyang}, \bibinfo{person}{Jeffrey Wu}, \bibinfo{person}{Xu Jiang}, \bibinfo{person}{Diogo Almeida}, \bibinfo{person}{Carroll Wainwright}, \bibinfo{person}{Pamela Mishkin}, \bibinfo{person}{Chong Zhang}, \bibinfo{person}{Sandhini Agarwal}, \bibinfo{person}{Katarina Slama}, \bibinfo{person}{Alex Ray}, {et~al\mbox{.}}} \bibinfo{year}{2022}\natexlab{a}.
\newblock \showarticletitle{Training language models to follow instructions with human feedback}.
\newblock \bibinfo{journal}{\emph{Advances in neural information processing systems}}  \bibinfo{volume}{35} (\bibinfo{year}{2022}), \bibinfo{pages}{27730--27744}.
\newblock


\bibitem[Ouyang et~al\mbox{.}(2022b)]%
        {InstructGPT}
\bibfield{author}{\bibinfo{person}{Long Ouyang}, \bibinfo{person}{Jeffrey Wu}, \bibinfo{person}{Xu Jiang}, \bibinfo{person}{Diogo Almeida}, \bibinfo{person}{Carroll Wainwright}, \bibinfo{person}{Pamela Mishkin}, \bibinfo{person}{Chong Zhang}, \bibinfo{person}{Sandhini Agarwal}, \bibinfo{person}{Katarina Slama}, \bibinfo{person}{Alex Ray}, {et~al\mbox{.}}} \bibinfo{year}{2022}\natexlab{b}.
\newblock \showarticletitle{Training language models to follow instructions with human feedback}.
\newblock \bibinfo{journal}{\emph{Advances in neural information processing systems}}  \bibinfo{volume}{35} (\bibinfo{year}{2022}), \bibinfo{pages}{27730--27744}.
\newblock


\bibitem[Pan et~al\mbox{.}(2025)]%
        {CRL}
\bibfield{author}{\bibinfo{person}{Chaofan Pan}, \bibinfo{person}{Xin Yang}, \bibinfo{person}{Yanhua Li}, \bibinfo{person}{Wei Wei}, \bibinfo{person}{Tianrui Li}, \bibinfo{person}{Bo An}, {and} \bibinfo{person}{Jiye Liang}.} \bibinfo{year}{2025}\natexlab{}.
\newblock \showarticletitle{A Survey of Continual Reinforcement Learning}.
\newblock \bibinfo{journal}{\emph{arXiv preprint arXiv:2506.21872}} (\bibinfo{year}{2025}).
\newblock


\bibitem[Pan et~al\mbox{.}(2024)]%
        {Hi-Core}
\bibfield{author}{\bibinfo{person}{Cheng Pan}, \bibinfo{person}{Xi Yang}, \bibinfo{person}{Haoran Wang}, \bibinfo{person}{Jiaming Zhang}, \bibinfo{person}{Jushun Li}, {and} \bibinfo{person}{Jie Xu}.} \bibinfo{year}{2024}\natexlab{}.
\newblock \showarticletitle{Hierarchical Continual Reinforcement Learning via Large Language Model}.
\newblock \bibinfo{journal}{\emph{arXiv preprint arXiv:2401.15098}} (\bibinfo{year}{2024}).
\newblock


\bibitem[Parisi et~al\mbox{.}(2019)]%
        {parisi2019continual}
\bibfield{author}{\bibinfo{person}{German~I Parisi}, \bibinfo{person}{Ronald Kemker}, \bibinfo{person}{Jose~L Part}, \bibinfo{person}{Christopher Kanan}, {and} \bibinfo{person}{Stefan Wermter}.} \bibinfo{year}{2019}\natexlab{}.
\newblock \showarticletitle{Continual lifelong learning with neural networks: A review}.
\newblock \bibinfo{journal}{\emph{Neural networks}}  \bibinfo{volume}{113} (\bibinfo{year}{2019}), \bibinfo{pages}{54--71}.
\newblock


\bibitem[Pearce et~al\mbox{.}(2023)]%
        {Pearce}
\bibfield{author}{\bibinfo{person}{Tim Pearce}, \bibinfo{person}{Tabish Rashid}, \bibinfo{person}{Anssi Kanervisto}, \bibinfo{person}{Dave Bignell}, \bibinfo{person}{Mingfei Sun}, \bibinfo{person}{Raluca Georgescu}, \bibinfo{person}{Sergio~Valcarcel Macua}, \bibinfo{person}{Shan~Zheng Tan}, \bibinfo{person}{Ida Momennejad}, \bibinfo{person}{Katja Hofmann}, {et~al\mbox{.}}} \bibinfo{year}{2023}\natexlab{}.
\newblock \showarticletitle{Imitating human behaviour with diffusion models}.
\newblock \bibinfo{journal}{\emph{arXiv preprint arXiv:2301.10677}} (\bibinfo{year}{2023}).
\newblock


\bibitem[Pertsch et~al\mbox{.}(2025)]%
        {fast}
\bibfield{author}{\bibinfo{person}{Karl Pertsch}, \bibinfo{person}{Kyle Stachowicz}, \bibinfo{person}{Brian Ichter}, \bibinfo{person}{Danny Driess}, \bibinfo{person}{Suraj Nair}, \bibinfo{person}{Quan Vuong}, \bibinfo{person}{Oier Mees}, \bibinfo{person}{Chelsea Finn}, {and} \bibinfo{person}{Sergey Levine}.} \bibinfo{year}{2025}\natexlab{}.
\newblock \showarticletitle{Fast: Efficient action tokenization for vision-language-action models}.
\newblock \bibinfo{journal}{\emph{arXiv preprint arXiv:2501.09747}} (\bibinfo{year}{2025}).
\newblock


\bibitem[Piatti et~al\mbox{.}(2024)]%
        {GovSim}
\bibfield{author}{\bibinfo{person}{Giorgio Piatti}, \bibinfo{person}{Zhijing Jin}, \bibinfo{person}{Max Kleiman-Weiner}, \bibinfo{person}{Bernhard Sch{\"o}lkopf}, \bibinfo{person}{Mrinmaya Sachan}, {and} \bibinfo{person}{Rada Mihalcea}.} \bibinfo{year}{2024}\natexlab{}.
\newblock \showarticletitle{Cooperate or collapse: Emergence of sustainable cooperation in a society of llm agents}.
\newblock \bibinfo{journal}{\emph{Advances in Neural Information Processing Systems}}  \bibinfo{volume}{37} (\bibinfo{year}{2024}), \bibinfo{pages}{111715--111759}.
\newblock


\bibitem[Piergiovanni et~al\mbox{.}(2019)]%
        {RobotDreamPolicy}
\bibfield{author}{\bibinfo{person}{AJ Piergiovanni}, \bibinfo{person}{Alan Wu}, {and} \bibinfo{person}{Michael~S Ryoo}.} \bibinfo{year}{2019}\natexlab{}.
\newblock \showarticletitle{Learning real-world robot policies by dreaming}. In \bibinfo{booktitle}{\emph{2019 IEEE/RSJ International Conference on Intelligent Robots and Systems (IROS)}}. IEEE, \bibinfo{pages}{7680--7687}.
\newblock


\bibitem[Qiao et~al\mbox{.}(2024)]%
        {WKM}
\bibfield{author}{\bibinfo{person}{Shuofei Qiao}, \bibinfo{person}{Runnan Fang}, \bibinfo{person}{Ningyu Zhang}, \bibinfo{person}{Yuqi Zhu}, \bibinfo{person}{Xiang Chen}, \bibinfo{person}{Shumin Deng}, \bibinfo{person}{Yong Jiang}, \bibinfo{person}{Pengjun Xie}, \bibinfo{person}{Fei Huang}, {and} \bibinfo{person}{Huajun Chen}.} \bibinfo{year}{2024}\natexlab{}.
\newblock \showarticletitle{Agent planning with world knowledge model}.
\newblock \bibinfo{journal}{\emph{Advances in Neural Information Processing Systems}}  \bibinfo{volume}{37} (\bibinfo{year}{2024}), \bibinfo{pages}{114843--114871}.
\newblock


\bibitem[Qu et~al\mbox{.}(2025)]%
        {spatialvla}
\bibfield{author}{\bibinfo{person}{Delin Qu}, \bibinfo{person}{Haoming Song}, \bibinfo{person}{Qizhi Chen}, \bibinfo{person}{Yuanqi Yao}, \bibinfo{person}{Xinyi Ye}, \bibinfo{person}{Yan Ding}, \bibinfo{person}{Zhigang Wang}, \bibinfo{person}{JiaYuan Gu}, \bibinfo{person}{Bin Zhao}, \bibinfo{person}{Dong Wang}, {et~al\mbox{.}}} \bibinfo{year}{2025}\natexlab{}.
\newblock \showarticletitle{Spatialvla: Exploring spatial representations for visual-language-action model}.
\newblock \bibinfo{journal}{\emph{arXiv preprint arXiv:2501.15830}} (\bibinfo{year}{2025}).
\newblock


\bibitem[Radford et~al\mbox{.}(2021)]%
        {CLIP}
\bibfield{author}{\bibinfo{person}{Alec Radford}, \bibinfo{person}{Jong~Wook Kim}, \bibinfo{person}{Chris Hallacy}, \bibinfo{person}{Aditya Ramesh}, \bibinfo{person}{Gabriel Goh}, \bibinfo{person}{Sandhini Agarwal}, \bibinfo{person}{Girish Sastry}, \bibinfo{person}{Amanda Askell}, \bibinfo{person}{Pamela Mishkin}, \bibinfo{person}{Jack Clark}, {et~al\mbox{.}}} \bibinfo{year}{2021}\natexlab{}.
\newblock \showarticletitle{Learning transferable visual models from natural language supervision}. In \bibinfo{booktitle}{\emph{International conference on machine learning}}. PmLR, \bibinfo{pages}{8748--8763}.
\newblock


\bibitem[Radford et~al\mbox{.}(2018)]%
        {GPT}
\bibfield{author}{\bibinfo{person}{Alec Radford}, \bibinfo{person}{Karthik Narasimhan}, \bibinfo{person}{Tim Salimans}, \bibinfo{person}{Ilya Sutskever}, {et~al\mbox{.}}} \bibinfo{year}{2018}\natexlab{}.
\newblock \showarticletitle{Improving language understanding by generative pre-training}.
\newblock  (\bibinfo{year}{2018}).
\newblock


\bibitem[Radford et~al\mbox{.}(2019)]%
        {GPT-2}
\bibfield{author}{\bibinfo{person}{Alec Radford}, \bibinfo{person}{Jeffrey Wu}, \bibinfo{person}{Rewon Child}, \bibinfo{person}{David Luan}, \bibinfo{person}{Dario Amodei}, \bibinfo{person}{Ilya Sutskever}, {et~al\mbox{.}}} \bibinfo{year}{2019}\natexlab{}.
\newblock \showarticletitle{Language models are unsupervised multitask learners}.
\newblock \bibinfo{journal}{\emph{OpenAI blog}} \bibinfo{volume}{1}, \bibinfo{number}{8} (\bibinfo{year}{2019}), \bibinfo{pages}{9}.
\newblock


\bibitem[Raiaan et~al\mbox{.}(2024)]%
        {raiaan2024review}
\bibfield{author}{\bibinfo{person}{Mohaimenul Azam~Khan Raiaan}, \bibinfo{person}{Md~Saddam~Hossain Mukta}, \bibinfo{person}{Kaniz Fatema}, \bibinfo{person}{Nur~Mohammad Fahad}, \bibinfo{person}{Sadman Sakib}, \bibinfo{person}{Most Marufatul~Jannat Mim}, \bibinfo{person}{Jubaer Ahmad}, \bibinfo{person}{Mohammed~Eunus Ali}, {and} \bibinfo{person}{Sami Azam}.} \bibinfo{year}{2024}\natexlab{}.
\newblock \showarticletitle{A review on large language models: Architectures, applications, taxonomies, open issues and challenges}.
\newblock \bibinfo{journal}{\emph{IEEE access}}  \bibinfo{volume}{12} (\bibinfo{year}{2024}), \bibinfo{pages}{26839--26874}.
\newblock


\bibitem[Rakelly et~al\mbox{.}(2019)]%
        {transferlearning}
\bibfield{author}{\bibinfo{person}{Kate Rakelly}, \bibinfo{person}{Aurick Zhou}, \bibinfo{person}{Chelsea Finn}, \bibinfo{person}{Sergey Levine}, {and} \bibinfo{person}{Deirdre Quillen}.} \bibinfo{year}{2019}\natexlab{}.
\newblock \showarticletitle{Efficient off-policy meta-reinforcement learning via probabilistic context variables}. In \bibinfo{booktitle}{\emph{International conference on machine learning}}. PMLR, \bibinfo{pages}{5331--5340}.
\newblock


\bibitem[Raman et~al\mbox{.}(2022)]%
        {raman2022planning}
\bibfield{author}{\bibinfo{person}{Sarthak~S Raman}, \bibinfo{person}{Vashisht Cohen}, \bibinfo{person}{Erez Rosen}, \bibinfo{person}{Shreyas Mirchandani}, \bibinfo{person}{Benjamin Arkin}, \bibinfo{person}{David Hedges}, \bibinfo{person}{Alex Sorokin}, \bibinfo{person}{Brent Gold}, \bibinfo{person}{Tsung-Yen Fu}, \bibinfo{person}{David Salac}, {et~al\mbox{.}}} \bibinfo{year}{2022}\natexlab{}.
\newblock \showarticletitle{Planning with large language models via corrective re-prompting}.
\newblock \bibinfo{journal}{\emph{arXiv preprint arXiv:2210.03952}} (\bibinfo{year}{2022}).
\newblock


\bibitem[Ramesh et~al\mbox{.}(2022)]%
        {DALL·E2}
\bibfield{author}{\bibinfo{person}{Aditya Ramesh}, \bibinfo{person}{Prafulla Dhariwal}, \bibinfo{person}{Alex Nichol}, \bibinfo{person}{Casey Chu}, {and} \bibinfo{person}{Mark Chen}.} \bibinfo{year}{2022}\natexlab{}.
\newblock \showarticletitle{Hierarchical text-conditional image generation with clip latents}.
\newblock \bibinfo{journal}{\emph{arXiv preprint arXiv:2204.06125}} \bibinfo{volume}{1}, \bibinfo{number}{2} (\bibinfo{year}{2022}), \bibinfo{pages}{3}.
\newblock


\bibitem[Ramesh et~al\mbox{.}(2021)]%
        {DALL·E}
\bibfield{author}{\bibinfo{person}{Aditya Ramesh}, \bibinfo{person}{Mikhail Pavlov}, \bibinfo{person}{Gabriel Goh}, \bibinfo{person}{Scott Gray}, \bibinfo{person}{Chelsea Voss}, \bibinfo{person}{Alec Radford}, \bibinfo{person}{Mark Chen}, {and} \bibinfo{person}{Ilya Sutskever}.} \bibinfo{year}{2021}\natexlab{}.
\newblock \showarticletitle{Zero-shot text-to-image generation}. In \bibinfo{booktitle}{\emph{International conference on machine learning}}. Pmlr, \bibinfo{pages}{8821--8831}.
\newblock


\bibitem[Ravi et~al\mbox{.}(2024)]%
        {sam2}
\bibfield{author}{\bibinfo{person}{Nikhila Ravi}, \bibinfo{person}{Valentin Gabeur}, \bibinfo{person}{Yuan-Ting Hu}, \bibinfo{person}{Ronghang Hu}, \bibinfo{person}{Chaitanya Ryali}, \bibinfo{person}{Tengyu Ma}, \bibinfo{person}{Haitham Khedr}, \bibinfo{person}{Roman R{\"a}dle}, \bibinfo{person}{Chloe Rolland}, \bibinfo{person}{Laura Gustafson}, {et~al\mbox{.}}} \bibinfo{year}{2024}\natexlab{}.
\newblock \showarticletitle{Sam 2: Segment anything in images and videos}.
\newblock \bibinfo{journal}{\emph{arXiv preprint arXiv:2408.00714}} (\bibinfo{year}{2024}).
\newblock


\bibitem[Raychaudhuri and Chang(2025)]%
        {raychaudhuri2025semantic}
\bibfield{author}{\bibinfo{person}{Sonia Raychaudhuri} {and} \bibinfo{person}{Angel~X Chang}.} \bibinfo{year}{2025}\natexlab{}.
\newblock \showarticletitle{Semantic Mapping in Indoor Embodied AI--A Comprehensive Survey and Future Directions}.
\newblock \bibinfo{journal}{\emph{arXiv preprint arXiv:2501.05750}} (\bibinfo{year}{2025}).
\newblock


\bibitem[Reed et~al\mbox{.}(2022)]%
        {Gato}
\bibfield{author}{\bibinfo{person}{Scott Reed}, \bibinfo{person}{Konrad Zolna}, \bibinfo{person}{Emilio Parisotto}, \bibinfo{person}{Sergio~Gomez Colmenarejo}, \bibinfo{person}{Alexander Novikov}, \bibinfo{person}{Gabriel Barth-Maron}, \bibinfo{person}{Mai Gimenez}, \bibinfo{person}{Yury Sulsky}, \bibinfo{person}{Jackie Kay}, \bibinfo{person}{Jost~Tobias Springenberg}, {et~al\mbox{.}}} \bibinfo{year}{2022}\natexlab{}.
\newblock \showarticletitle{A generalist agent}.
\newblock \bibinfo{journal}{\emph{arXiv preprint arXiv:2205.06175}} (\bibinfo{year}{2022}).
\newblock


\bibitem[Reid et~al\mbox{.}(2022)]%
        {Reid}
\bibfield{author}{\bibinfo{person}{Machel Reid}, \bibinfo{person}{Yutaro Yamada}, {and} \bibinfo{person}{Shixiang~Shane Gu}.} \bibinfo{year}{2022}\natexlab{}.
\newblock \showarticletitle{Can wikipedia help offline reinforcement learning?}
\newblock \bibinfo{journal}{\emph{arXiv preprint arXiv:2201.12122}} (\bibinfo{year}{2022}).
\newblock


\bibitem[Reily et~al\mbox{.}(2022)]%
        {reily2022real}
\bibfield{author}{\bibinfo{person}{Brian Reily}, \bibinfo{person}{Peng Gao}, \bibinfo{person}{Fei Han}, \bibinfo{person}{Hua Wang}, {and} \bibinfo{person}{Hao Zhang}.} \bibinfo{year}{2022}\natexlab{}.
\newblock \showarticletitle{Real-time recognition of team behaviors by multisensory graph-embedded robot learning}.
\newblock \bibinfo{journal}{\emph{The International Journal of Robotics Research}} \bibinfo{volume}{41}, \bibinfo{number}{8} (\bibinfo{year}{2022}), \bibinfo{pages}{798--811}.
\newblock


\bibitem[Ren et~al\mbox{.}(2023)]%
        {KNOWNO}
\bibfield{author}{\bibinfo{person}{Allen~Z Ren}, \bibinfo{person}{Anushri Dixit}, \bibinfo{person}{Alexandra Bodrova}, \bibinfo{person}{Sumeet Singh}, \bibinfo{person}{Stephen Tu}, \bibinfo{person}{Noah Brown}, \bibinfo{person}{Peng Xu}, \bibinfo{person}{Leila Takayama}, \bibinfo{person}{Fei Xia}, \bibinfo{person}{Jake Varley}, {et~al\mbox{.}}} \bibinfo{year}{2023}\natexlab{}.
\newblock \showarticletitle{Robots that ask for help: Uncertainty alignment for large language model planners}.
\newblock \bibinfo{journal}{\emph{arXiv preprint arXiv:2307.01928}} (\bibinfo{year}{2023}).
\newblock


\bibitem[Robine et~al\mbox{.}(2023)]%
        {TWM}
\bibfield{author}{\bibinfo{person}{Jan Robine}, \bibinfo{person}{Marc H{\"o}ftmann}, \bibinfo{person}{Tobias Uelwer}, {and} \bibinfo{person}{Stefan Harmeling}.} \bibinfo{year}{2023}\natexlab{}.
\newblock \showarticletitle{Transformer-based world models are happy with 100k interactions}.
\newblock \bibinfo{journal}{\emph{arXiv preprint arXiv:2303.07109}} (\bibinfo{year}{2023}).
\newblock


\bibitem[Rudin et~al\mbox{.}(2022)]%
        {rudin2022learning}
\bibfield{author}{\bibinfo{person}{Nikita Rudin}, \bibinfo{person}{David Hoeller}, \bibinfo{person}{Philipp Reist}, {and} \bibinfo{person}{Marco Hutter}.} \bibinfo{year}{2022}\natexlab{}.
\newblock \showarticletitle{Learning to walk in minutes using massively parallel deep reinforcement learning}. In \bibinfo{booktitle}{\emph{Conference on robot learning}}. PMLR, \bibinfo{pages}{91--100}.
\newblock


\bibitem[Rummery and Niranjan(1994)]%
        {SARSA}
\bibfield{author}{\bibinfo{person}{Gavin~A Rummery} {and} \bibinfo{person}{Mahesan Niranjan}.} \bibinfo{year}{1994}\natexlab{}.
\newblock \bibinfo{booktitle}{\emph{On-line Q-learning using connectionist systems}}. Vol.~\bibinfo{volume}{37}.
\newblock \bibinfo{publisher}{University of Cambridge, Department of Engineering Cambridge, UK}.
\newblock


\bibitem[Sapkota et~al\mbox{.}(2025)]%
        {12}
\bibfield{author}{\bibinfo{person}{Ranjan Sapkota}, \bibinfo{person}{Yang Cao}, \bibinfo{person}{Konstantinos~I Roumeliotis}, {and} \bibinfo{person}{Manoj Karkee}.} \bibinfo{year}{2025}\natexlab{}.
\newblock \showarticletitle{Vision-language-action models: Concepts, progress, applications and challenges}.
\newblock \bibinfo{journal}{\emph{arXiv preprint arXiv:2505.04769}} (\bibinfo{year}{2025}).
\newblock


\bibitem[Schulman et~al\mbox{.}(2017)]%
        {PPO}
\bibfield{author}{\bibinfo{person}{John Schulman}, \bibinfo{person}{Filip Wolski}, \bibinfo{person}{Prafulla Dhariwal}, \bibinfo{person}{Alec Radford}, {and} \bibinfo{person}{Oleg Klimov}.} \bibinfo{year}{2017}\natexlab{}.
\newblock \showarticletitle{Proximal policy optimization algorithms}.
\newblock \bibinfo{journal}{\emph{arXiv preprint arXiv:1707.06347}} (\bibinfo{year}{2017}).
\newblock


\bibitem[Sharma et~al\mbox{.}(2021)]%
        {sharma2021skill}
\bibfield{author}{\bibinfo{person}{Pratyusha Sharma}, \bibinfo{person}{Antonio Torralba}, {and} \bibinfo{person}{Jacob Andreas}.} \bibinfo{year}{2021}\natexlab{}.
\newblock \showarticletitle{Skill induction and planning with latent language}.
\newblock \bibinfo{journal}{\emph{arXiv preprint arXiv:2110.01517}} (\bibinfo{year}{2021}).
\newblock


\bibitem[Shi et~al\mbox{.}(2024)]%
        {YAY}
\bibfield{author}{\bibinfo{person}{Lucy~Xiaoyang Shi}, \bibinfo{person}{Zheyuan Hu}, \bibinfo{person}{Tony~Z Zhao}, \bibinfo{person}{Archit Sharma}, \bibinfo{person}{Karl Pertsch}, \bibinfo{person}{Jianlan Luo}, \bibinfo{person}{Sergey Levine}, {and} \bibinfo{person}{Chelsea Finn}.} \bibinfo{year}{2024}\natexlab{}.
\newblock \showarticletitle{Yell at your robot: Improving on-the-fly from language corrections}.
\newblock \bibinfo{journal}{\emph{arXiv preprint arXiv:2403.12910}} (\bibinfo{year}{2024}).
\newblock


\bibitem[Shi et~al\mbox{.}(2023)]%
        {LaMo}
\bibfield{author}{\bibinfo{person}{Ruizhe Shi}, \bibinfo{person}{Yuyao Liu}, \bibinfo{person}{Yanjie Ze}, \bibinfo{person}{Simon~S Du}, {and} \bibinfo{person}{Huazhe Xu}.} \bibinfo{year}{2023}\natexlab{}.
\newblock \showarticletitle{Unleashing the power of pre-trained language models for offline reinforcement learning}.
\newblock \bibinfo{journal}{\emph{arXiv preprint arXiv:2310.20587}} (\bibinfo{year}{2023}).
\newblock


\bibitem[Shinn et~al\mbox{.}(2023)]%
        {reflexion}
\bibfield{author}{\bibinfo{person}{Noah Shinn}, \bibinfo{person}{Federico Cassano}, \bibinfo{person}{Ashwin Gopinath}, \bibinfo{person}{Karthik Narasimhan}, {and} \bibinfo{person}{Shunyu Yao}.} \bibinfo{year}{2023}\natexlab{}.
\newblock \showarticletitle{Reflexion: Language agents with verbal reinforcement learning}.
\newblock \bibinfo{journal}{\emph{Advances in Neural Information Processing Systems}}  \bibinfo{volume}{36} (\bibinfo{year}{2023}), \bibinfo{pages}{8634--8652}.
\newblock


\bibitem[Shirai et~al\mbox{.}(2024)]%
        {ViLaIn}
\bibfield{author}{\bibinfo{person}{Keisuke Shirai}, \bibinfo{person}{Cristian~C Beltran-Hernandez}, \bibinfo{person}{Masashi Hamaya}, \bibinfo{person}{Atsushi Hashimoto}, \bibinfo{person}{Shohei Tanaka}, \bibinfo{person}{Kento Kawaharazuka}, \bibinfo{person}{Kazutoshi Tanaka}, \bibinfo{person}{Yoshitaka Ushiku}, {and} \bibinfo{person}{Shinsuke Mori}.} \bibinfo{year}{2024}\natexlab{}.
\newblock \showarticletitle{Vision-language interpreter for robot task planning}. In \bibinfo{booktitle}{\emph{2024 IEEE International Conference on Robotics and Automation (ICRA)}}. IEEE, \bibinfo{pages}{2051--2058}.
\newblock


\bibitem[Shridhar et~al\mbox{.}(2022)]%
        {CLIPort}
\bibfield{author}{\bibinfo{person}{Mohit Shridhar}, \bibinfo{person}{Lucas Manuelli}, {and} \bibinfo{person}{Dieter Fox}.} \bibinfo{year}{2022}\natexlab{}.
\newblock \showarticletitle{Cliport: What and where pathways for robotic manipulation}. In \bibinfo{booktitle}{\emph{Conference on Robot Learning}}. PMLR, \bibinfo{pages}{894--906}.
\newblock


\bibitem[Silver et~al\mbox{.}(2016a)]%
        {silver2016mastering}
\bibfield{author}{\bibinfo{person}{David Silver}, \bibinfo{person}{Aja Huang}, \bibinfo{person}{Chris~J Maddison}, \bibinfo{person}{Arthur Guez}, \bibinfo{person}{Laurent Sifre}, \bibinfo{person}{George Van Den~Driessche}, \bibinfo{person}{Julian Schrittwieser}, \bibinfo{person}{Ioannis Antonoglou}, \bibinfo{person}{Veda Panneershelvam}, \bibinfo{person}{Marc Lanctot}, {et~al\mbox{.}}} \bibinfo{year}{2016}\natexlab{a}.
\newblock \showarticletitle{Mastering the game of Go with deep neural networks and tree search}.
\newblock \bibinfo{journal}{\emph{nature}} \bibinfo{volume}{529}, \bibinfo{number}{7587} (\bibinfo{year}{2016}), \bibinfo{pages}{484--489}.
\newblock


\bibitem[Silver et~al\mbox{.}(2016b)]%
        {AlphaGo}
\bibfield{author}{\bibinfo{person}{David Silver}, \bibinfo{person}{Aja Huang}, \bibinfo{person}{Chris~J Maddison}, \bibinfo{person}{Arthur Guez}, \bibinfo{person}{Laurent Sifre}, \bibinfo{person}{George Van Den~Driessche}, \bibinfo{person}{Julian Schrittwieser}, \bibinfo{person}{Ioannis Antonoglou}, \bibinfo{person}{Veda Panneershelvam}, \bibinfo{person}{Marc Lanctot}, {et~al\mbox{.}}} \bibinfo{year}{2016}\natexlab{b}.
\newblock \showarticletitle{Mastering the game of Go with deep neural networks and tree search}.
\newblock \bibinfo{journal}{\emph{nature}} \bibinfo{volume}{529}, \bibinfo{number}{7587} (\bibinfo{year}{2016}), \bibinfo{pages}{484--489}.
\newblock


\bibitem[Silver et~al\mbox{.}(2022)]%
        {FSP-LLM}
\bibfield{author}{\bibinfo{person}{Tom Silver}, \bibinfo{person}{Varun Hariprasad}, \bibinfo{person}{Reece~S Shuttleworth}, \bibinfo{person}{Nishanth Kumar}, \bibinfo{person}{Tom{\'a}s Lozano-P{\'e}rez}, {and} \bibinfo{person}{Leslie~Pack Kaelbling}.} \bibinfo{year}{2022}\natexlab{}.
\newblock \showarticletitle{PDDL planning with pretrained large language models}. In \bibinfo{booktitle}{\emph{NeurIPS 2022 foundation models for decision making workshop}}.
\newblock


\bibitem[Singh et~al\mbox{.}(2022)]%
        {ProgPrompt}
\bibfield{author}{\bibinfo{person}{Ishika Singh}, \bibinfo{person}{Valts Blukis}, \bibinfo{person}{Arsalan Mousavian}, \bibinfo{person}{Ankit Goyal}, \bibinfo{person}{Danfei Xu}, \bibinfo{person}{Jonathan Tremblay}, \bibinfo{person}{Dieter Fox}, \bibinfo{person}{Jesse Thomason}, {and} \bibinfo{person}{Animesh Garg}.} \bibinfo{year}{2022}\natexlab{}.
\newblock \showarticletitle{Progprompt: Generating situated robot task plans using large language models}.
\newblock \bibinfo{journal}{\emph{arXiv preprint arXiv:2209.11302}} (\bibinfo{year}{2022}).
\newblock


\bibitem[Sun et~al\mbox{.}(2024)]%
        {sun2024comprehensive}
\bibfield{author}{\bibinfo{person}{Fuchun Sun}, \bibinfo{person}{Runfa Chen}, \bibinfo{person}{Tianying Ji}, \bibinfo{person}{Yu Luo}, \bibinfo{person}{Huaidong Zhou}, {and} \bibinfo{person}{Huaping Liu}.} \bibinfo{year}{2024}\natexlab{}.
\newblock \showarticletitle{A comprehensive survey on embodied intelligence: Advancements, challenges, and future perspectives}.
\newblock \bibinfo{journal}{\emph{CAAI Artificial Intelligence Research}}  \bibinfo{volume}{3} (\bibinfo{year}{2024}).
\newblock


\bibitem[Tate et~al\mbox{.}(2009)]%
        {sfc}
\bibfield{author}{\bibinfo{person}{Ed~D Tate}, \bibinfo{person}{Jessy~W Grizzle}, {and} \bibinfo{person}{Huei Peng}.} \bibinfo{year}{2009}\natexlab{}.
\newblock \showarticletitle{SP-SDP for fuel consumption and tailpipe emissions minimization in an EVT hybrid}.
\newblock \bibinfo{journal}{\emph{IEEE Transactions on Control Systems Technology}} \bibinfo{volume}{18}, \bibinfo{number}{3} (\bibinfo{year}{2009}), \bibinfo{pages}{673--687}.
\newblock


\bibitem[Team et~al\mbox{.}(2023)]%
        {Gemini}
\bibfield{author}{\bibinfo{person}{Gemini Team}, \bibinfo{person}{Rohan Anil}, \bibinfo{person}{Sebastian Borgeaud}, \bibinfo{person}{Jean-Baptiste Alayrac}, \bibinfo{person}{Jiahui Yu}, \bibinfo{person}{Radu Soricut}, \bibinfo{person}{Johan Schalkwyk}, \bibinfo{person}{Andrew~M Dai}, \bibinfo{person}{Anja Hauth}, \bibinfo{person}{Katie Millican}, {et~al\mbox{.}}} \bibinfo{year}{2023}\natexlab{}.
\newblock \showarticletitle{Gemini: a family of highly capable multimodal models}.
\newblock \bibinfo{journal}{\emph{arXiv preprint arXiv:2312.11805}} (\bibinfo{year}{2023}).
\newblock


\bibitem[Team et~al\mbox{.}(2024)]%
        {Octo}
\bibfield{author}{\bibinfo{person}{Octo~Model Team}, \bibinfo{person}{Dibya Ghosh}, \bibinfo{person}{Homer Walke}, \bibinfo{person}{Karl Pertsch}, \bibinfo{person}{Kevin Black}, \bibinfo{person}{Oier Mees}, \bibinfo{person}{Sudeep Dasari}, \bibinfo{person}{Joey Hejna}, \bibinfo{person}{Tobias Kreiman}, \bibinfo{person}{Charles Xu}, {et~al\mbox{.}}} \bibinfo{year}{2024}\natexlab{}.
\newblock \showarticletitle{Octo: An open-source generalist robot policy}.
\newblock \bibinfo{journal}{\emph{arXiv preprint arXiv:2405.12213}} (\bibinfo{year}{2024}).
\newblock


\bibitem[Tirinzoni et~al\mbox{.}(2018)]%
        {tirinzoni2018importance}
\bibfield{author}{\bibinfo{person}{Andrea Tirinzoni}, \bibinfo{person}{Andrea Sessa}, \bibinfo{person}{Matteo Pirotta}, {and} \bibinfo{person}{Marcello Restelli}.} \bibinfo{year}{2018}\natexlab{}.
\newblock \showarticletitle{Importance weighted transfer of samples in reinforcement learning}. In \bibinfo{booktitle}{\emph{International Conference on Machine Learning}}. PMLR, \bibinfo{pages}{4936--4945}.
\newblock


\bibitem[Touvron et~al\mbox{.}(2023a)]%
        {Llama1}
\bibfield{author}{\bibinfo{person}{Hugo Touvron}, \bibinfo{person}{Thibaut Lavril}, \bibinfo{person}{Gautier Izacard}, \bibinfo{person}{Xavier Martinet}, \bibinfo{person}{Marie-Anne Lachaux}, \bibinfo{person}{Timoth{\'e}e Lacroix}, \bibinfo{person}{Baptiste Rozi{\`e}re}, \bibinfo{person}{Naman Goyal}, \bibinfo{person}{Eric Hambro}, \bibinfo{person}{Faisal Azhar}, {et~al\mbox{.}}} \bibinfo{year}{2023}\natexlab{a}.
\newblock \showarticletitle{Llama: Open and efficient foundation language models}.
\newblock \bibinfo{journal}{\emph{arXiv preprint arXiv:2302.13971}} (\bibinfo{year}{2023}).
\newblock


\bibitem[Touvron et~al\mbox{.}(2023b)]%
        {Llama2}
\bibfield{author}{\bibinfo{person}{Hugo Touvron}, \bibinfo{person}{Louis Martin}, \bibinfo{person}{Kevin Stone}, \bibinfo{person}{Peter Albert}, \bibinfo{person}{Amjad Almahairi}, \bibinfo{person}{Yasmine Babaei}, \bibinfo{person}{Nikolay Bashlykov}, \bibinfo{person}{Soumya Batra}, \bibinfo{person}{Prajjwal Bhargava}, \bibinfo{person}{Shruti Bhosale}, {et~al\mbox{.}}} \bibinfo{year}{2023}\natexlab{b}.
\newblock \showarticletitle{Llama 2: Open foundation and fine-tuned chat models}.
\newblock \bibinfo{journal}{\emph{arXiv preprint arXiv:2307.09288}} (\bibinfo{year}{2023}).
\newblock


\bibitem[Turing(1950)]%
        {AGI}
\bibfield{author}{\bibinfo{person}{A.~M. Turing}.} \bibinfo{year}{1950}\natexlab{}.
\newblock \showarticletitle{Computing machinery and intelligence}.
\newblock \bibinfo{journal}{\emph{Mind}} \bibinfo{volume}{LIX}, \bibinfo{number}{236} (\bibinfo{year}{1950}), \bibinfo{pages}{433--460}.
\newblock


\bibitem[Valmeekam et~al\mbox{.}(2022)]%
        {valmeekam2022large}
\bibfield{author}{\bibinfo{person}{Karthik Valmeekam}, \bibinfo{person}{Alberto Olmo}, \bibinfo{person}{Sarath Sreedharan}, {and} \bibinfo{person}{Subbarao Kambhampati}.} \bibinfo{year}{2022}\natexlab{}.
\newblock \showarticletitle{Large language models still can't plan (a benchmark for LLMs on planning and reasoning about change)}. In \bibinfo{booktitle}{\emph{NeurIPS 2022 Foundation Models for Decision Making Workshop}}.
\newblock


\bibitem[Vuong et~al\mbox{.}(2023)]%
        {openx}
\bibfield{author}{\bibinfo{person}{Quan Vuong}, \bibinfo{person}{Sergey Levine}, \bibinfo{person}{Homer~Rich Walke}, \bibinfo{person}{Karl Pertsch}, \bibinfo{person}{Anikait Singh}, \bibinfo{person}{Ria Doshi}, \bibinfo{person}{Charles Xu}, \bibinfo{person}{Jianlan Luo}, \bibinfo{person}{Liam Tan}, \bibinfo{person}{Dhruv Shah}, {et~al\mbox{.}}} \bibinfo{year}{2023}\natexlab{}.
\newblock \showarticletitle{Open x-embodiment: Robotic learning datasets and rt-x models}. In \bibinfo{booktitle}{\emph{Towards Generalist Robots: Learning Paradigms for Scalable Skill Acquisition@ CoRL2023}}.
\newblock


\bibitem[Wagenmaker et~al\mbox{.}(2024)]%
        {wagenmaker2024overcoming}
\bibfield{author}{\bibinfo{person}{Andrew Wagenmaker}, \bibinfo{person}{Kevin Huang}, \bibinfo{person}{Liyiming Ke}, \bibinfo{person}{Kevin Jamieson}, {and} \bibinfo{person}{Abhishek Gupta}.} \bibinfo{year}{2024}\natexlab{}.
\newblock \showarticletitle{Overcoming the Sim-to-Real Gap: Leveraging Simulation to Learn to Explore for Real-World RL}.
\newblock \bibinfo{journal}{\emph{Advances in Neural Information Processing Systems}}  \bibinfo{volume}{37} (\bibinfo{year}{2024}), \bibinfo{pages}{78715--78765}.
\newblock


\bibitem[Wan et~al\mbox{.}(2024)]%
        {wan2024thinking}
\bibfield{author}{\bibinfo{person}{Zishen Wan}, \bibinfo{person}{Yuhang Du}, \bibinfo{person}{Mohamed Ibrahim}, \bibinfo{person}{Yang Zhao}, \bibinfo{person}{Tushar Krishna}, {and} \bibinfo{person}{Arijit Raychowdhury}.} \bibinfo{year}{2024}\natexlab{}.
\newblock \showarticletitle{Thinking and moving: An efficient computing approach for integrated task and motion planning in cooperative embodied ai systems}. In \bibinfo{booktitle}{\emph{Proceedings of the 43rd IEEE/ACM International Conference on Computer-Aided Design}}. \bibinfo{pages}{1--7}.
\newblock


\bibitem[Wang et~al\mbox{.}(2023b)]%
        {Voyager}
\bibfield{author}{\bibinfo{person}{Guanzhi Wang}, \bibinfo{person}{Yuqi Xie}, \bibinfo{person}{Yunfan Jiang}, \bibinfo{person}{Ajay Mandlekar}, \bibinfo{person}{Chaowei Xiao}, \bibinfo{person}{Yuke Zhu}, \bibinfo{person}{Linxi Fan}, {and} \bibinfo{person}{Anima Anandkumar}.} \bibinfo{year}{2023}\natexlab{b}.
\newblock \showarticletitle{Voyager: An open-ended embodied agent with large language models}.
\newblock \bibinfo{journal}{\emph{arXiv preprint arXiv:2305.16291}} (\bibinfo{year}{2023}).
\newblock


\bibitem[Wang et~al\mbox{.}(2025)]%
        {16}
\bibfield{author}{\bibinfo{person}{Jiaqi Wang}, \bibinfo{person}{Enze Shi}, \bibinfo{person}{Huawen Hu}, \bibinfo{person}{Chong Ma}, \bibinfo{person}{Yiheng Liu}, \bibinfo{person}{Xuhui Wang}, \bibinfo{person}{Yincheng Yao}, \bibinfo{person}{Xuan Liu}, \bibinfo{person}{Bao Ge}, {and} \bibinfo{person}{Shu Zhang}.} \bibinfo{year}{2025}\natexlab{}.
\newblock \showarticletitle{Large language models for robotics: Opportunities, challenges, and perspectives}.
\newblock \bibinfo{journal}{\emph{Journal of Automation and Intelligence}} \bibinfo{volume}{4}, \bibinfo{number}{1} (\bibinfo{year}{2025}), \bibinfo{pages}{52--64}.
\newblock


\bibitem[Wang et~al\mbox{.}(2024)]%
        {wang2024exploring}
\bibfield{author}{\bibinfo{person}{Yiqi Wang}, \bibinfo{person}{Wentao Chen}, \bibinfo{person}{Xiaotian Han}, \bibinfo{person}{Xudong Lin}, \bibinfo{person}{Haiteng Zhao}, \bibinfo{person}{Yongfei Liu}, \bibinfo{person}{Bohan Zhai}, \bibinfo{person}{Jianbo Yuan}, \bibinfo{person}{Quanzeng You}, {and} \bibinfo{person}{Hongxia Yang}.} \bibinfo{year}{2024}\natexlab{}.
\newblock \showarticletitle{Exploring the reasoning abilities of multimodal large language models (mllms): A comprehensive survey on emerging trends in multimodal reasoning}.
\newblock \bibinfo{journal}{\emph{arXiv preprint arXiv:2401.06805}} (\bibinfo{year}{2024}).
\newblock


\bibitem[Wang et~al\mbox{.}(2023a)]%
        {DEPS}
\bibfield{author}{\bibinfo{person}{Ziyi Wang}, \bibinfo{person}{Sheng Cai}, \bibinfo{person}{Guandao Chen}, \bibinfo{person}{Si-Yuan Chen}, \bibinfo{person}{Hong-Xin Yang}, \bibinfo{person}{Shuyuan Liu}, \bibinfo{person}{Zihao Zhao}, \bibinfo{person}{Yuxiang Wang}, \bibinfo{person}{Chen Chen}, \bibinfo{person}{Leo~J Li}, {et~al\mbox{.}}} \bibinfo{year}{2023}\natexlab{a}.
\newblock \showarticletitle{Describe, explain, plan and select: Interactive planning with large language models enables open-world multi-task agents}.
\newblock \bibinfo{journal}{\emph{arXiv preprint arXiv:2302.01560}} (\bibinfo{year}{2023}).
\newblock


\bibitem[Wang et~al\mbox{.}(2022)]%
        {Diffusion-QL}
\bibfield{author}{\bibinfo{person}{Zhendong Wang}, \bibinfo{person}{Jonathan~J Hunt}, {and} \bibinfo{person}{Mingyuan Zhou}.} \bibinfo{year}{2022}\natexlab{}.
\newblock \showarticletitle{Diffusion policies as an expressive policy class for offline reinforcement learning}.
\newblock \bibinfo{journal}{\emph{arXiv preprint arXiv:2208.06193}} (\bibinfo{year}{2022}).
\newblock


\bibitem[Watkins et~al\mbox{.}(1989)]%
        {Q-Learning}
\bibfield{author}{\bibinfo{person}{Christopher John Cornish~Hellaby Watkins} {et~al\mbox{.}}} \bibinfo{year}{1989}\natexlab{}.
\newblock \showarticletitle{Learning from delayed rewards}.
\newblock  (\bibinfo{year}{1989}).
\newblock


\bibitem[Wei et~al\mbox{.}(2022)]%
        {CoT}
\bibfield{author}{\bibinfo{person}{Jason Wei}, \bibinfo{person}{Xuezhi Wang}, \bibinfo{person}{Dale Schuurmans}, \bibinfo{person}{Maarten Bosma}, \bibinfo{person}{Fei Xia}, \bibinfo{person}{Ed Chi}, \bibinfo{person}{Quoc~V Le}, \bibinfo{person}{Denny Zhou}, {et~al\mbox{.}}} \bibinfo{year}{2022}\natexlab{}.
\newblock \showarticletitle{Chain-of-thought prompting elicits reasoning in large language models}.
\newblock \bibinfo{journal}{\emph{Advances in neural information processing systems}}  \bibinfo{volume}{35} (\bibinfo{year}{2022}), \bibinfo{pages}{24824--24837}.
\newblock


\bibitem[Wen et~al\mbox{.}(2024)]%
        {Diffusion-VLA}
\bibfield{author}{\bibinfo{person}{Junjie Wen}, \bibinfo{person}{Minjie Zhu}, \bibinfo{person}{Yichen Zhu}, \bibinfo{person}{Zhibin Tang}, \bibinfo{person}{Jinming Li}, \bibinfo{person}{Zhongyi Zhou}, \bibinfo{person}{Chengmeng Li}, \bibinfo{person}{Xiaoyu Liu}, \bibinfo{person}{Yaxin Peng}, \bibinfo{person}{Chaomin Shen}, {et~al\mbox{.}}} \bibinfo{year}{2024}\natexlab{}.
\newblock \showarticletitle{Diffusion-VLA: Generalizable and Interpretable Robot Foundation Model via Self-Generated Reasoning}.
\newblock \bibinfo{journal}{\emph{arXiv preprint arXiv:2412.03293}} (\bibinfo{year}{2024}).
\newblock


\bibitem[Wen et~al\mbox{.}(2025a)]%
        {dexvla}
\bibfield{author}{\bibinfo{person}{Junjie Wen}, \bibinfo{person}{Yichen Zhu}, \bibinfo{person}{Jinming Li}, \bibinfo{person}{Zhibin Tang}, \bibinfo{person}{Chaomin Shen}, {and} \bibinfo{person}{Feifei Feng}.} \bibinfo{year}{2025}\natexlab{a}.
\newblock \showarticletitle{Dexvla: Vision-language model with plug-in diffusion expert for general robot control}.
\newblock \bibinfo{journal}{\emph{arXiv preprint arXiv:2502.05855}} (\bibinfo{year}{2025}).
\newblock


\bibitem[Wen et~al\mbox{.}(2025b)]%
        {TinyVLA}
\bibfield{author}{\bibinfo{person}{Junjie Wen}, \bibinfo{person}{Yichen Zhu}, \bibinfo{person}{Jinming Li}, \bibinfo{person}{Minjie Zhu}, \bibinfo{person}{Zhibin Tang}, \bibinfo{person}{Kun Wu}, \bibinfo{person}{Zhiyuan Xu}, \bibinfo{person}{Ning Liu}, \bibinfo{person}{Ran Cheng}, \bibinfo{person}{Chaomin Shen}, {et~al\mbox{.}}} \bibinfo{year}{2025}\natexlab{b}.
\newblock \showarticletitle{Tinyvla: Towards fast, data-efficient vision-language-action models for robotic manipulation}.
\newblock \bibinfo{journal}{\emph{IEEE Robotics and Automation Letters}} (\bibinfo{year}{2025}).
\newblock


\bibitem[Weng et~al\mbox{.}(2024)]%
        {CycleResearcher}
\bibfield{author}{\bibinfo{person}{Yixuan Weng}, \bibinfo{person}{Minjun Zhu}, \bibinfo{person}{Guangsheng Bao}, \bibinfo{person}{Hongbo Zhang}, \bibinfo{person}{Jindong Wang}, \bibinfo{person}{Yue Zhang}, {and} \bibinfo{person}{Linyi Yang}.} \bibinfo{year}{2024}\natexlab{}.
\newblock \showarticletitle{Cycleresearcher: Improving automated research via automated review}.
\newblock \bibinfo{journal}{\emph{arXiv preprint arXiv:2411.00816}} (\bibinfo{year}{2024}).
\newblock


\bibitem[Wilkins(2014)]%
        {wilkins2014practical}
\bibfield{author}{\bibinfo{person}{David~E Wilkins}.} \bibinfo{year}{2014}\natexlab{}.
\newblock \bibinfo{booktitle}{\emph{Practical planning: extending the classical AI planning paradigm}}.
\newblock \bibinfo{publisher}{Elsevier}.
\newblock


\bibitem[Wong et~al\mbox{.}(2025)]%
        {wong2025survey}
\bibfield{author}{\bibinfo{person}{Lik Hang~Kenny Wong}, \bibinfo{person}{Xueyang Kang}, \bibinfo{person}{Kaixin Bai}, {and} \bibinfo{person}{Jianwei Zhang}.} \bibinfo{year}{2025}\natexlab{}.
\newblock \showarticletitle{A Survey of Robotic Navigation and Manipulation with Physics Simulators in the Era of Embodied AI}.
\newblock \bibinfo{journal}{\emph{arXiv preprint arXiv:2505.01458}} (\bibinfo{year}{2025}).
\newblock


\bibitem[Wu et~al\mbox{.}(2023a)]%
        {DayDreamer}
\bibfield{author}{\bibinfo{person}{Philipp Wu}, \bibinfo{person}{Alejandro Escontrela}, \bibinfo{person}{Danijar Hafner}, \bibinfo{person}{Pieter Abbeel}, {and} \bibinfo{person}{Ken Goldberg}.} \bibinfo{year}{2023}\natexlab{a}.
\newblock \showarticletitle{Daydreamer: World models for physical robot learning}. In \bibinfo{booktitle}{\emph{Conference on robot learning}}. PMLR, \bibinfo{pages}{2226--2240}.
\newblock


\bibitem[Wu et~al\mbox{.}(2023b)]%
        {TaPA}
\bibfield{author}{\bibinfo{person}{Zhenyu Wu}, \bibinfo{person}{Ziwei Wang}, \bibinfo{person}{Xiuwei Xu}, \bibinfo{person}{Jiwen Lu}, {and} \bibinfo{person}{Haibin Yan}.} \bibinfo{year}{2023}\natexlab{b}.
\newblock \showarticletitle{Embodied task planning with large language models}.
\newblock \bibinfo{journal}{\emph{arXiv preprint arXiv:2307.01848}} (\bibinfo{year}{2023}).
\newblock


\bibitem[Xiao et~al\mbox{.}(2025)]%
        {robotlearning}
\bibfield{author}{\bibinfo{person}{Xuan Xiao}, \bibinfo{person}{Jiahang Liu}, \bibinfo{person}{Zhipeng Wang}, \bibinfo{person}{Yanmin Zhou}, \bibinfo{person}{Yong Qi}, \bibinfo{person}{Shuo Jiang}, \bibinfo{person}{Bin He}, {and} \bibinfo{person}{Qian Cheng}.} \bibinfo{year}{2025}\natexlab{}.
\newblock \showarticletitle{Robot learning in the era of foundation models: A survey}.
\newblock \bibinfo{journal}{\emph{Neurocomputing}} (\bibinfo{year}{2025}), \bibinfo{pages}{129963}.
\newblock


\bibitem[Xie et~al\mbox{.}(2024)]%
        {Text2Reward}
\bibfield{author}{\bibinfo{person}{Tianbao Xie}, \bibinfo{person}{Siheng Zhao}, \bibinfo{person}{Chen~Henry Wu}, \bibinfo{person}{Yitao Liu}, \bibinfo{person}{Qian Luo}, \bibinfo{person}{Victor Zhong}, \bibinfo{person}{Yanchao Yang}, {and} \bibinfo{person}{Tao Yu}.} \bibinfo{year}{2024}\natexlab{}.
\newblock \showarticletitle{Text2reward: Automated dense reward function generation for reinforcement learning}. In \bibinfo{booktitle}{\emph{International Conference on Learning Representations (ICLR), 2024 (07/05/2024-11/05/2024, Vienna, Austria)}}.
\newblock


\bibitem[Xu et~al\mbox{.}(2020)]%
        {ST-Transformer}
\bibfield{author}{\bibinfo{person}{Mingxing Xu}, \bibinfo{person}{Wenrui Dai}, \bibinfo{person}{Chunmiao Liu}, \bibinfo{person}{Xing Gao}, \bibinfo{person}{Weiyao Lin}, \bibinfo{person}{Guo-Jun Qi}, {and} \bibinfo{person}{Hongkai Xiong}.} \bibinfo{year}{2020}\natexlab{}.
\newblock \showarticletitle{Spatial-temporal transformer networks for traffic flow forecasting}.
\newblock \bibinfo{journal}{\emph{arXiv preprint arXiv:2001.02908}} (\bibinfo{year}{2020}).
\newblock


\bibitem[Xu et~al\mbox{.}(2022)]%
        {Prompt-DT}
\bibfield{author}{\bibinfo{person}{Mengdi Xu}, \bibinfo{person}{Yikang Shen}, \bibinfo{person}{Shun Zhang}, \bibinfo{person}{Yuchen Lu}, \bibinfo{person}{Ding Zhao}, \bibinfo{person}{Joshua Tenenbaum}, {and} \bibinfo{person}{Chuang Gan}.} \bibinfo{year}{2022}\natexlab{}.
\newblock \showarticletitle{Prompting decision transformer for few-shot policy generalization}. In \bibinfo{booktitle}{\emph{international conference on machine learning}}. PMLR, \bibinfo{pages}{24631--24645}.
\newblock


\bibitem[Xu et~al\mbox{.}(2025)]%
        {casheVLA}
\bibfield{author}{\bibinfo{person}{Siyu Xu}, \bibinfo{person}{Yunke Wang}, \bibinfo{person}{Chenghao Xia}, \bibinfo{person}{Dihao Zhu}, \bibinfo{person}{Tao Huang}, {and} \bibinfo{person}{Chang Xu}.} \bibinfo{year}{2025}\natexlab{}.
\newblock \showarticletitle{Vla-cache: Towards efficient vision-language-action model via adaptive token caching in robotic manipulation}.
\newblock \bibinfo{journal}{\emph{arXiv preprint arXiv:2502.02175}} (\bibinfo{year}{2025}).
\newblock


\bibitem[Xu et~al\mbox{.}(2024a)]%
        {xu2024survey}
\bibfield{author}{\bibinfo{person}{Zhiyuan Xu}, \bibinfo{person}{Kun Wu}, \bibinfo{person}{Junjie Wen}, \bibinfo{person}{Jinming Li}, \bibinfo{person}{Ning Liu}, \bibinfo{person}{Zhengping Che}, {and} \bibinfo{person}{Jian Tang}.} \bibinfo{year}{2024}\natexlab{a}.
\newblock \showarticletitle{A survey on robotics with foundation models: toward embodied ai}.
\newblock \bibinfo{journal}{\emph{arXiv preprint arXiv:2402.02385}} (\bibinfo{year}{2024}).
\newblock


\bibitem[Xu et~al\mbox{.}(2024b)]%
        {7}
\bibfield{author}{\bibinfo{person}{Zhiyuan Xu}, \bibinfo{person}{Kun Wu}, \bibinfo{person}{Junjie Wen}, \bibinfo{person}{Jinming Li}, \bibinfo{person}{Ning Liu}, \bibinfo{person}{Zhengping Che}, {and} \bibinfo{person}{Jian Tang}.} \bibinfo{year}{2024}\natexlab{b}.
\newblock \showarticletitle{A survey on robotics with foundation models: toward embodied ai}.
\newblock \bibinfo{journal}{\emph{arXiv preprint arXiv:2402.02385}} (\bibinfo{year}{2024}).
\newblock


\bibitem[Yang et~al\mbox{.}(2024)]%
        {Octopus}
\bibfield{author}{\bibinfo{person}{Jingkang Yang}, \bibinfo{person}{Yuhao Dong}, \bibinfo{person}{Shuai Liu}, \bibinfo{person}{Bo Li}, \bibinfo{person}{Ziyue Wang}, \bibinfo{person}{Haoran Tan}, \bibinfo{person}{Chencheng Jiang}, \bibinfo{person}{Jiamu Kang}, \bibinfo{person}{Yuanhan Zhang}, \bibinfo{person}{Kaiyang Zhou}, {et~al\mbox{.}}} \bibinfo{year}{2024}\natexlab{}.
\newblock \showarticletitle{Octopus: Embodied vision-language programmer from environmental feedback}. In \bibinfo{booktitle}{\emph{European conference on computer vision}}. Springer, \bibinfo{pages}{20--38}.
\newblock


\bibitem[Yang et~al\mbox{.}(2023)]%
        {UniSim}
\bibfield{author}{\bibinfo{person}{Mengjiao Yang}, \bibinfo{person}{Yilun Du}, \bibinfo{person}{Kamyar Ghasemipour}, \bibinfo{person}{Jonathan Tompson}, \bibinfo{person}{Dale Schuurmans}, {and} \bibinfo{person}{Pieter Abbeel}.} \bibinfo{year}{2023}\natexlab{}.
\newblock \showarticletitle{Learning interactive real-world simulators}.
\newblock \bibinfo{journal}{\emph{arXiv preprint arXiv:2310.06114}} \bibinfo{volume}{1}, \bibinfo{number}{2} (\bibinfo{year}{2023}), \bibinfo{pages}{6}.
\newblock


\bibitem[Yao et~al\mbox{.}(2023a)]%
        {ToT}
\bibfield{author}{\bibinfo{person}{Shunyu Yao}, \bibinfo{person}{Dian Yu}, \bibinfo{person}{Jeffrey Zhao}, \bibinfo{person}{Izhak Shafran}, \bibinfo{person}{Tom Griffiths}, \bibinfo{person}{Yuan Cao}, {and} \bibinfo{person}{Karthik Narasimhan}.} \bibinfo{year}{2023}\natexlab{a}.
\newblock \showarticletitle{Tree of thoughts: Deliberate problem solving with large language models}.
\newblock \bibinfo{journal}{\emph{Advances in neural information processing systems}}  \bibinfo{volume}{36} (\bibinfo{year}{2023}), \bibinfo{pages}{11809--11822}.
\newblock


\bibitem[Yao et~al\mbox{.}(2023b)]%
        {ReAct}
\bibfield{author}{\bibinfo{person}{Shunyu Yao}, \bibinfo{person}{Jeffrey Zhao}, \bibinfo{person}{Dian Yu}, \bibinfo{person}{Nan Du}, \bibinfo{person}{Izhak Shafran}, \bibinfo{person}{Karthik Narasimhan}, {and} \bibinfo{person}{Yuan Cao}.} \bibinfo{year}{2023}\natexlab{b}.
\newblock \showarticletitle{React: Synergizing reasoning and acting in language models}. In \bibinfo{booktitle}{\emph{International Conference on Learning Representations (ICLR)}}.
\newblock


\bibitem[Yu et~al\mbox{.}(2023)]%
        {L2R}
\bibfield{author}{\bibinfo{person}{Wenhao Yu}, \bibinfo{person}{Nimrod Gileadi}, \bibinfo{person}{Chuyuan Fu}, \bibinfo{person}{Sean Kirmani}, \bibinfo{person}{Kuang-Huei Lee}, \bibinfo{person}{Montse~Gonzalez Arenas}, \bibinfo{person}{Hao-Tien~Lewis Chiang}, \bibinfo{person}{Tom Erez}, \bibinfo{person}{Leonard Hasenclever}, \bibinfo{person}{Jan Humplik}, {et~al\mbox{.}}} \bibinfo{year}{2023}\natexlab{}.
\newblock \showarticletitle{Language to rewards for robotic skill synthesis}.
\newblock \bibinfo{journal}{\emph{arXiv preprint arXiv:2306.08647}} (\bibinfo{year}{2023}).
\newblock


\bibitem[Yurtsever et~al\mbox{.}(2020)]%
        {Autonomousdriving}
\bibfield{author}{\bibinfo{person}{Ekim Yurtsever}, \bibinfo{person}{Jacob Lambert}, \bibinfo{person}{Alexander Carballo}, {and} \bibinfo{person}{Kazuya Takeda}.} \bibinfo{year}{2020}\natexlab{}.
\newblock \showarticletitle{A survey of autonomous driving: Common practices and emerging technologies}.
\newblock \bibinfo{journal}{\emph{IEEE access}}  \bibinfo{volume}{8} (\bibinfo{year}{2020}), \bibinfo{pages}{58443--58469}.
\newblock


\bibitem[Ze et~al\mbox{.}(2024)]%
        {3D-Diffusion}
\bibfield{author}{\bibinfo{person}{Yanjie Ze}, \bibinfo{person}{Gu Zhang}, \bibinfo{person}{Kangning Zhang}, \bibinfo{person}{Chenyuan Hu}, \bibinfo{person}{Muhan Wang}, {and} \bibinfo{person}{Huazhe Xu}.} \bibinfo{year}{2024}\natexlab{}.
\newblock \showarticletitle{3d diffusion policy: Generalizable visuomotor policy learning via simple 3d representations}.
\newblock \bibinfo{journal}{\emph{arXiv preprint arXiv:2403.03954}} (\bibinfo{year}{2024}).
\newblock


\bibitem[Zhang et~al\mbox{.}(2023a)]%
        {VideoLLaMA}
\bibfield{author}{\bibinfo{person}{Hang Zhang}, \bibinfo{person}{Xin Li}, {and} \bibinfo{person}{Lidong Bing}.} \bibinfo{year}{2023}\natexlab{a}.
\newblock \showarticletitle{Video-llama: An instruction-tuned audio-visual language model for video understanding}.
\newblock \bibinfo{journal}{\emph{arXiv preprint arXiv:2306.02858}} (\bibinfo{year}{2023}).
\newblock


\bibitem[Zhang et~al\mbox{.}(2025)]%
        {MoleVLA}
\bibfield{author}{\bibinfo{person}{Rongyu Zhang}, \bibinfo{person}{Menghang Dong}, \bibinfo{person}{Yuan Zhang}, \bibinfo{person}{Liang Heng}, \bibinfo{person}{Xiaowei Chi}, \bibinfo{person}{Gaole Dai}, \bibinfo{person}{Li Du}, \bibinfo{person}{Yuan Du}, {and} \bibinfo{person}{Shanghang Zhang}.} \bibinfo{year}{2025}\natexlab{}.
\newblock \showarticletitle{Mole-vla: Dynamic layer-skipping vision language action model via mixture-of-layers for efficient robot manipulation}.
\newblock \bibinfo{journal}{\emph{arXiv preprint arXiv:2503.20384}} (\bibinfo{year}{2025}).
\newblock


\bibitem[Zhang et~al\mbox{.}(2022)]%
        {zhang2022survey}
\bibfield{author}{\bibinfo{person}{Tianyao Zhang}, \bibinfo{person}{Xiaoguang Hu}, \bibinfo{person}{Jin Xiao}, {and} \bibinfo{person}{Guofeng Zhang}.} \bibinfo{year}{2022}\natexlab{}.
\newblock \showarticletitle{A survey of visual navigation: From geometry to embodied AI}.
\newblock \bibinfo{journal}{\emph{Engineering Applications of Artificial Intelligence}}  \bibinfo{volume}{114} (\bibinfo{year}{2022}), \bibinfo{pages}{105036}.
\newblock


\bibitem[Zhang et~al\mbox{.}(2024)]%
        {Agent-Pro}
\bibfield{author}{\bibinfo{person}{Wenqi Zhang}, \bibinfo{person}{Ke Tang}, \bibinfo{person}{Hai Wu}, \bibinfo{person}{Mengna Wang}, \bibinfo{person}{Yongliang Shen}, \bibinfo{person}{Guiyang Hou}, \bibinfo{person}{Zeqi Tan}, \bibinfo{person}{Peng Li}, \bibinfo{person}{Yueting Zhuang}, {and} \bibinfo{person}{Weiming Lu}.} \bibinfo{year}{2024}\natexlab{}.
\newblock \showarticletitle{Agent-pro: Learning to evolve via policy-level reflection and optimization}.
\newblock \bibinfo{journal}{\emph{arXiv preprint arXiv:2402.17574}} (\bibinfo{year}{2024}).
\newblock


\bibitem[Zhang et~al\mbox{.}(2023c)]%
        {STORM}
\bibfield{author}{\bibinfo{person}{Weipu Zhang}, \bibinfo{person}{Gang Wang}, \bibinfo{person}{Jian Sun}, \bibinfo{person}{Yetian Yuan}, {and} \bibinfo{person}{Gao Huang}.} \bibinfo{year}{2023}\natexlab{c}.
\newblock \showarticletitle{Storm: Efficient stochastic transformer based world models for reinforcement learning}.
\newblock \bibinfo{journal}{\emph{Advances in Neural Information Processing Systems}}  \bibinfo{volume}{36} (\bibinfo{year}{2023}), \bibinfo{pages}{27147--27166}.
\newblock


\bibitem[Zhang et~al\mbox{.}(2023b)]%
        {gpt-4v}
\bibfield{author}{\bibinfo{person}{Xinlu Zhang}, \bibinfo{person}{Yujie Lu}, \bibinfo{person}{Weizhi Wang}, \bibinfo{person}{An Yan}, \bibinfo{person}{Jun Yan}, \bibinfo{person}{Lianke Qin}, \bibinfo{person}{Heng Wang}, \bibinfo{person}{Xifeng Yan}, \bibinfo{person}{William~Yang Wang}, {and} \bibinfo{person}{Linda~Ruth Petzold}.} \bibinfo{year}{2023}\natexlab{b}.
\newblock \showarticletitle{Gpt-4v (ision) as a generalist evaluator for vision-language tasks}.
\newblock \bibinfo{journal}{\emph{arXiv preprint arXiv:2311.01361}} (\bibinfo{year}{2023}).
\newblock


\bibitem[Zhao et~al\mbox{.}(2023a)]%
        {ALOHA}
\bibfield{author}{\bibinfo{person}{Tony~Z Zhao}, \bibinfo{person}{Vikash Kumar}, \bibinfo{person}{Sergey Levine}, {and} \bibinfo{person}{Chelsea Finn}.} \bibinfo{year}{2023}\natexlab{a}.
\newblock \showarticletitle{Learning fine-grained bimanual manipulation with low-cost hardware}.
\newblock \bibinfo{journal}{\emph{arXiv preprint arXiv:2304.13705}} (\bibinfo{year}{2023}).
\newblock


\bibitem[Zhao et~al\mbox{.}(2023b)]%
        {zhao2023survey}
\bibfield{author}{\bibinfo{person}{Wayne~Xin Zhao}, \bibinfo{person}{Kun Zhou}, \bibinfo{person}{Junyi Li}, \bibinfo{person}{Tianyi Tang}, \bibinfo{person}{Xiaolei Wang}, \bibinfo{person}{Yupeng Hou}, \bibinfo{person}{Yingqian Min}, \bibinfo{person}{Beichen Zhang}, \bibinfo{person}{Junjie Zhang}, \bibinfo{person}{Zican Dong}, {et~al\mbox{.}}} \bibinfo{year}{2023}\natexlab{b}.
\newblock \showarticletitle{A survey of large language models}.
\newblock \bibinfo{journal}{\emph{arXiv preprint arXiv:2303.18223}} \bibinfo{volume}{1}, \bibinfo{number}{2} (\bibinfo{year}{2023}).
\newblock


\bibitem[Zhen et~al\mbox{.}(2024)]%
        {3D-VLA}
\bibfield{author}{\bibinfo{person}{Haoyu Zhen}, \bibinfo{person}{Xiaowen Qiu}, \bibinfo{person}{Peihao Chen}, \bibinfo{person}{Jincheng Yang}, \bibinfo{person}{Xin Yan}, \bibinfo{person}{Yilun Du}, \bibinfo{person}{Yining Hong}, {and} \bibinfo{person}{Chuang Gan}.} \bibinfo{year}{2024}\natexlab{}.
\newblock \showarticletitle{3d-vla: A 3d vision-language-action generative world model}.
\newblock \bibinfo{journal}{\emph{arXiv preprint arXiv:2403.09631}} (\bibinfo{year}{2024}).
\newblock


\bibitem[Zheng et~al\mbox{.}(2025)]%
        {roadmap}
\bibfield{author}{\bibinfo{person}{Junhao Zheng}, \bibinfo{person}{Chengming Shi}, \bibinfo{person}{Xidi Cai}, \bibinfo{person}{Qiuke Li}, \bibinfo{person}{Duzhen Zhang}, \bibinfo{person}{Chenxing Li}, \bibinfo{person}{Dong Yu}, {and} \bibinfo{person}{Qianli Ma}.} \bibinfo{year}{2025}\natexlab{}.
\newblock \showarticletitle{Lifelong learning of large language model based agents: A roadmap}.
\newblock \bibinfo{journal}{\emph{arXiv preprint arXiv:2501.07278}} (\bibinfo{year}{2025}).
\newblock


\bibitem[Zheng et~al\mbox{.}(2022)]%
        {ODT}
\bibfield{author}{\bibinfo{person}{Qinqing Zheng}, \bibinfo{person}{Amy Zhang}, {and} \bibinfo{person}{Aditya Grover}.} \bibinfo{year}{2022}\natexlab{}.
\newblock \showarticletitle{Online decision transformer}. In \bibinfo{booktitle}{\emph{international conference on machine learning}}. PMLR, \bibinfo{pages}{27042--27059}.
\newblock


\bibitem[Zheng et~al\mbox{.}(2024)]%
        {TraceVLA}
\bibfield{author}{\bibinfo{person}{Ruijie Zheng}, \bibinfo{person}{Yongyuan Liang}, \bibinfo{person}{Shuaiyi Huang}, \bibinfo{person}{Jianfeng Gao}, \bibinfo{person}{Hal Daum{\'e}~III}, \bibinfo{person}{Andrey Kolobov}, \bibinfo{person}{Furong Huang}, {and} \bibinfo{person}{Jianwei Yang}.} \bibinfo{year}{2024}\natexlab{}.
\newblock \showarticletitle{Tracevla: Visual trace prompting enhances spatial-temporal awareness for generalist robotic policies}.
\newblock \bibinfo{journal}{\emph{arXiv preprint arXiv:2412.10345}} (\bibinfo{year}{2024}).
\newblock


\bibitem[Zhong et~al\mbox{.}(2025)]%
        {dexgraspvla}
\bibfield{author}{\bibinfo{person}{Yifan Zhong}, \bibinfo{person}{Xuchuan Huang}, \bibinfo{person}{Ruochong Li}, \bibinfo{person}{Ceyao Zhang}, \bibinfo{person}{Yitao Liang}, \bibinfo{person}{Yaodong Yang}, {and} \bibinfo{person}{Yuanpei Chen}.} \bibinfo{year}{2025}\natexlab{}.
\newblock \showarticletitle{Dexgraspvla: A vision-language-action framework towards general dexterous grasping}.
\newblock \bibinfo{journal}{\emph{arXiv preprint arXiv:2502.20900}} (\bibinfo{year}{2025}).
\newblock


\bibitem[Zhou et~al\mbox{.}(2024)]%
        {ISR-LLM}
\bibfield{author}{\bibinfo{person}{Zhehua Zhou}, \bibinfo{person}{Jiayang Song}, \bibinfo{person}{Kunpeng Yao}, \bibinfo{person}{Zhan Shu}, {and} \bibinfo{person}{Lei Ma}.} \bibinfo{year}{2024}\natexlab{}.
\newblock \showarticletitle{Isr-llm: Iterative self-refined large language model for long-horizon sequential task planning}. In \bibinfo{booktitle}{\emph{2024 IEEE International Conference on Robotics and Automation (ICRA)}}. IEEE, \bibinfo{pages}{2081--2088}.
\newblock


\bibitem[Zhu et~al\mbox{.}(2023)]%
        {MiniGPT-4}
\bibfield{author}{\bibinfo{person}{Deyao Zhu}, \bibinfo{person}{Jun Chen}, \bibinfo{person}{Xiaoqian Shen}, \bibinfo{person}{Xiang Li}, {and} \bibinfo{person}{Mohamed Elhoseiny}.} \bibinfo{year}{2023}\natexlab{}.
\newblock \showarticletitle{Minigpt-4: Enhancing vision-language understanding with advanced large language models}.
\newblock \bibinfo{journal}{\emph{arXiv preprint arXiv:2304.10592}} (\bibinfo{year}{2023}).
\newblock


\bibitem[Zhu et~al\mbox{.}(2024)]%
        {worldmodel}
\bibfield{author}{\bibinfo{person}{Zheng Zhu}, \bibinfo{person}{Xiaofeng Wang}, \bibinfo{person}{Wangbo Zhao}, \bibinfo{person}{Chen Min}, \bibinfo{person}{Nianchen Deng}, \bibinfo{person}{Min Dou}, \bibinfo{person}{Yuqi Wang}, \bibinfo{person}{Botian Shi}, \bibinfo{person}{Kai Wang}, \bibinfo{person}{Chi Zhang}, {et~al\mbox{.}}} \bibinfo{year}{2024}\natexlab{}.
\newblock \showarticletitle{Is sora a world simulator? a comprehensive survey on general world models and beyond}.
\newblock \bibinfo{journal}{\emph{arXiv preprint arXiv:2405.03520}} (\bibinfo{year}{2024}).
\newblock


\bibitem[Zitkovich et~al\mbox{.}(2023)]%
        {RT-2}
\bibfield{author}{\bibinfo{person}{Brianna Zitkovich}, \bibinfo{person}{Tianhe Yu}, \bibinfo{person}{Sichun Xu}, \bibinfo{person}{Peng Xu}, \bibinfo{person}{Ted Xiao}, \bibinfo{person}{Fei Xia}, \bibinfo{person}{Jialin Wu}, \bibinfo{person}{Paul Wohlhart}, \bibinfo{person}{Stefan Welker}, \bibinfo{person}{Ayzaan Wahid}, {et~al\mbox{.}}} \bibinfo{year}{2023}\natexlab{}.
\newblock \showarticletitle{Rt-2: Vision-language-action models transfer web knowledge to robotic control}. In \bibinfo{booktitle}{\emph{Conference on Robot Learning}}. PMLR, \bibinfo{pages}{2165--2183}.
\newblock


\end{thebibliography}


\end{document}